\documentclass[a4paper,fleqn]{cas-sc}

\usepackage[authoryear,longnamesfirst]{natbib}
\usepackage{amsmath,amsfonts}
\usepackage{amssymb} 

\usepackage{adjustbox}

\usepackage{multirow}
\usepackage{multicol}
\usepackage{color}

\def\tsc#1{\csdef{#1}{\textsc{\lowercase{#1}}\xspace}}
\tsc{WGM}
\tsc{QE}

\begin{document}
\let\WriteBookmarks\relax
\def\floatpagepagefraction{1}
\def\textpagefraction{.001}

\shorttitle{Thorough GNN-based RUL prediction Survey}    

\shortauthors{Wang et al.}  

\title [mode = title]{A Survey on Graph Neural Networks for Remaining Useful Life Prediction: Methodologies, Evaluation and Future Trends}  



%

\author[1,2]{Yucheng Wang}



\ead{yucheng003@e.ntu.edu.sg}



\author[1]{Min Wu}



\ead{wumin@i2r.a-star.edu.sg}

\author[1,3]{Xiaoli Li}

\ead{xlli@i2r.a-star.edu.sg}

\author[2]{Lihua Xie}

\ead{elhxie@ntu.edu.s}

\author[1,3]{Zhenghua Chen}

\ead{chen0832@e.ntu.edu.sg}
\cormark[1]

\affiliation[1]{organization={Institute for Infocomm Research, A$^*$STAR},
            city={Singapore},
            postcode={138632}, 
            country={Singapore}}

\affiliation[2]{organization={School of Electrical and Electronic Engineering, Nanyang Technological University},
            city={Singapore},
            postcode={639798}, 
            country={Singapore}}

\affiliation[3]{organization={Centre for Frontier AI Research, A$^*$STAR},
            city={Singapore},
            postcode={138632}, 
            country={Singapore}}

\cortext[1]{Corresponding author}



\begin{abstract}
Remaining Useful Life (RUL) prediction is a critical aspect of Prognostics and Health Management (PHM), aimed at predicting the future state of a system to enable timely maintenance and prevent unexpected failures. While existing deep learning methods have shown promise, they often struggle to fully leverage the spatial information inherent in complex systems, limiting their effectiveness in RUL prediction. To address this challenge, recent research has explored the use of Graph Neural Networks (GNNs) to model spatial information for more accurate RUL prediction. This paper presents a comprehensive review of GNN techniques applied to RUL prediction, summarizing existing methods and offering guidance for future research. We first propose a novel taxonomy based on the stages of adapting GNNs to RUL prediction, systematically categorizing approaches into four key stages: graph construction, graph modeling, graph information processing, and graph readout. By organizing the field in this way, we highlight the unique challenges and considerations at each stage of the GNN pipeline. Additionally, we conduct a thorough evaluation of various state-of-the-art (SOTA) GNN methods, ensuring consistent experimental settings for fair comparisons. This rigorous analysis yields valuable insights into the strengths and weaknesses of different approaches, serving as an experimental guide for researchers and practitioners working in this area. Finally, we identify and discuss several promising research directions that could further advance the field, emphasizing the potential for GNNs to revolutionize RUL prediction and enhance the effectiveness of PHM strategies. The benchmarking codes are available in GitHub: https://github.com/Frank-Wang-oss/GNN\_RUL\_Benchmarking.

\end{abstract}


\begin{highlights}
\item We provide a survey for Graph Neural Network-based Remaining Useful Life prediction
\item The survey organized by a novel taxonomy based on four stages of adapting GNNs
\item We provide a thorough evaluation of SOTA RUL GNN methods for fair comparisons
\item Promising research directions are provided to further advance the field
\item Codes provided for SOTA methods' re-implementation
\end{highlights}


\maketitle

\section{Introduction}



The prediction of Remaining Useful Life (RUL) is a critical component in the field of Prognostics and Health Management (PHM), which aims to predict the future state of a system to ensure timely maintenance and prevent unexpected failures \citep{wang2024remaining,karatzinis2024aircraft,zhang2024novel}. Accurate RUL prediction enable predictive maintenance, which can significantly reduce downtime, improve safety, and optimize the lifecycle management of machinery and equipment. Additionally, effective RUL prediction can enhance decision-making processes, improve resource allocation, and reduce maintenance costs. In recent years, deep learning has become increasingly important in RUL prediction due to its ability to model complex patterns and dependencies, providing more accurate and reliable predictions compared to traditional methods, such as statistical approaches \citep{si2011remaining} and physics-based models \citep{lei2016model,sikorska2011prognostic,li2024review}.

Existing studies in RUL prediction have primarily focused on utilizing temporal encoders such as Temporal Convolutional Networks (TCN) \citep{qiu2023piecewise}, Gated Recurrent Units (GRU), Convolutional Neural Networks (CNN) \citep{shang2024novel}, and Long Short-Term Memory (LSTM) networks. These methods have achieved strong performance due to their ability to capture temporal information, which refers to the time-based patterns and sequences within the data, such as trends and periodic behaviors. However, they are not effective at capturing spatial information, which limits their performance in RUL prediction. Spatial information pertains to the structural and relational patterns within the data, such as the connections and interactions between different components of a system. For example, multiple sensors are typically deployed to detect various physical parameters to fully assess the health status of a component, and there is important spatial information among the sensors. Existing temporal methods struggle to model the multi-channel relationships and dependencies inherent in the spatial domain, restricting their effectiveness for RUL prediction.

Graph Neural Networks (GNNs) offer a promising solution to the above limitation by providing a framework to capture both temporal and spatial information. GNNs are specifically designed to operate on graph-structured data, which can naturally represent the spatial relationships between different components of a system. By leveraging the capabilities of GNNs, it is possible to model the intricate dependencies and interactions within the data, leading to more comprehensive and accurate RUL prediction. This makes GNNs a powerful tool for enhancing the predictive performance for RUL where both temporal and spatial information are crucial. In recent years, GNNs have been widely used in this area. Thus, it is necessary to present a survey to summarize the contributions of recent works.

\begin{figure}[!htbp]
    \centering
    \includegraphics[width = .4\linewidth]{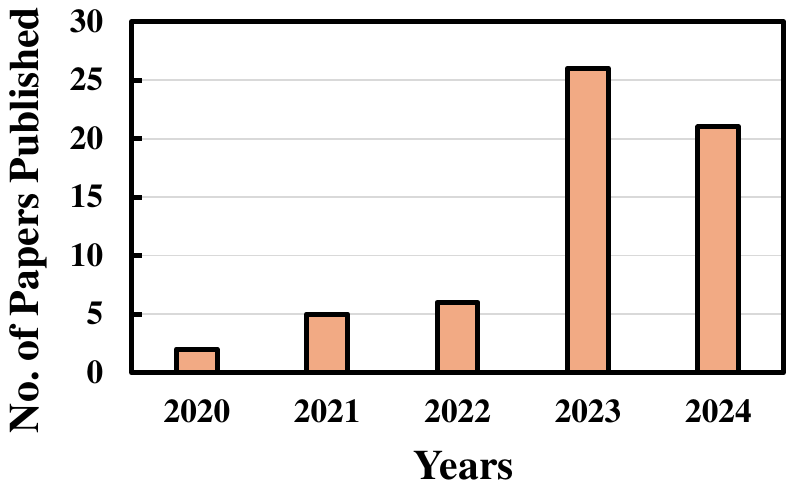}
    \caption{The number of papers working on GNNs for RUL prediction published in recent years.}
    \label{fig:paperincreasing}
\end{figure}

Numerous survey papers have been published to review the working mechanisms of GNNs and their applications across various domains \citep{wu2020comprehensive,liang2022survey,yuan2022explainability}. Additionally, some surveys focus on specific time-series applications of GNNs, such as EEG classification \citep{klepl2024graph}, traffic forecasting \citep{jiang2022graph}, fault diagonis \citep{chen2021graph} and so on \citep{cao2020spectral}. Meanwhile, there are also review papers dedicated to RUL prediction  \citep{wang2020remaining,ferreira2022remaining,chen2023transfer}. However, while survey papers exist for both GNNs and RUL prediction, there is a scarcity of surveys that specifically address the intersection of these two areas, failing to discuss the GNN models specifically designed for RUL prediction. Currently, there is one survey paper \citep{li2022emerging} that discusses the use of GNNs in PHM, which includes a section on RUL prediction. However, this paper has several limitations. First, it primarily focuses on fault diagnosis, lacking a detailed analysis of how GNNs should be applied to RUL prediction. Second, it was published in late 2021, and since then, the number of GNN-based studies has increased exponentially, particularly after 2022, as shown in \ref{fig:paperincreasing}. Consequently, the survey fails to cover these recent advancements. Third, while it evaluates general GNN models, it does not provide comprehensive evaluations of recent GNN models specifically designed for RUL prediction. Notably, existing works fail to maintain consistent experimental settings, making fair comparisons challenging. This inconsistency hampers newcomers' ability to compare their research with existing works, thus restricting the development of new research in this area.
Given these limitations, there is a need for a new survey paper that summarizes the recent advancements in applying GNNs to RUL prediction and categorizes the works based on the properties of this area. This survey should not only analyze the latest research but also provide comprehensive benchmarks of existing GNN models tailored for RUL prediction, offering valuable evaluation and guidance for researchers.

This paper offers a focused review of GNNs for RUL prediction, covering recent advancements, comprehensive evaluations, and future opportunities for enhancing GNNs in this domain. Our analysis of the literature aims to clarify and evaluate the current state of GNNs in RUL prediction while identifying potential points for further improvement. To categorize existing works, we introduce a novel taxonomy tailored to the unique properties of RUL prediction, providing a focused and comprehensive analysis of GNN methods applied to this field. Additionally, we contribute by benchmarking selected GNN approaches specifically designed for RUL prediction, facilitating easy understanding of this area and practical assessments for newcomers. 
The key contributions of the proposed survey are summarized as follows:
\begin{itemize}
    \item \textbf{Novel Taxonomy:} Introducing a new taxonomy tailored to the properties of RUL prediction, including graph construction, graph models, graph information, and graph readout, to effectively summarize existing works.
    \item \textbf{Comprehensive Evaluation:} Conducting thorough evaluations of existing GNN approaches for RUL prediction. These evaluations provide valuable benchmarks that researchers can use to compare their proposed methods, facilitating research engagement and practical assessments.
    \item \textbf{Research Opportunities:} Identifying future research directions based on the properties of RUL to better improve the adaptation of GNN models in this area.
\end{itemize}

\section{Problem Formulation}

Given a sample \(\mathbf{X} \in \mathbb{R}^{S \times L}\) collected from machines, where \(S\) represents the number of channels and \(L\) represents the time length, we denote \(x_s\) as the signals with \(L\) timestamps, i.e., \(\{x_{s,1}, x_{s,2}, \ldots, x_{s,L}\}\), collected from the \(s\)-th channel. Here, \(S\) could be one or more than one, depending on specific downstream tasks.
The data \(\mathbf{X}\) is used to construct the graph \(\mathcal{G} = \{V, E\}\), where:
\begin{itemize}
    \item \(V = \{v_n\}_n^N\) represents the set of nodes, with node features \(\mathbf{Z} = \{\mathbf{z}_n\}_n^N\), derived from the signals \(X\), where $N$ is the number of nodes. These features could be the raw signals themselves or features extracted through some preprocessing or feature engineering methods.
    \item \(E\) represents the set of edges, indicating the connections between nodes. To represent the connections mathematically, the adjacency matrix \(\mathbf{A} \in \mathbb{R}^{N \times N}\) is defined, encoding the graph structure \(\mathcal{G}\). Each element \(A_{nm}\) in the matrix represents the presence (and possibly the strength) of an edge between node \(n\) and node \(m\):
\end{itemize}
\[
A_{nm} = \begin{cases}
1 & \text{if } (v_n, v_m) \in E, \\
0 & \text{otherwise}.
\end{cases}
\]
If the edges have weights to indicate the strength of the relationships, $A$ can be defined as a weighted adjacency matrix:
\[
A_{nm} = \begin{cases}
w_{nm} & \text{if } (v_n, v_m) \in E, \\
0 & \text{otherwise}.
\end{cases}
\]
Then, the dependency information within the graph can be captured by graph models, learning effective final representations for RUL prediction.

\section{Graph Neural Network for RUL prediction}

\subsection{Workflow}
\begin{figure*}[b]
    \centering
    \includegraphics[width = 1.\linewidth]{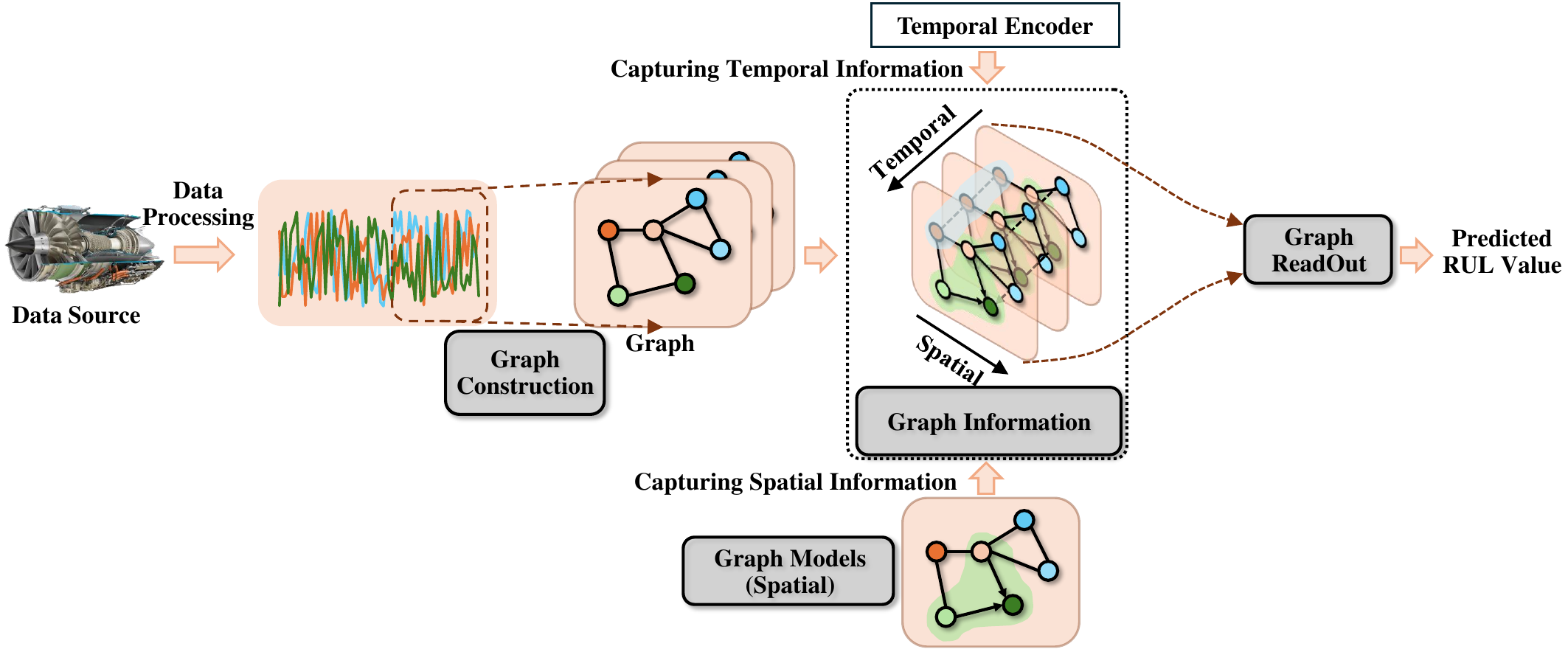}
    \caption{Workflow of GNN for RUL prediction. }
    \label{fig:overallflow}
\end{figure*}

The overall workflow of GNN for RUL prediction is illustrated in Fig. \ref{fig:overallflow}. Time-series data is collected from machines, such as turbofan engines or bearings. Since graphs are implicit within the data, it is necessary to construct graphs to model the spatial-temporal dependencies. Thus, after processing the data, a graph construction module is typically required. With the constructed graphs, graph models and temporal encoders are utilized to capture the spatial and temporal information within the graphs, respectively. Finally, a graph readout function is used to aggregate and extract high-level information from these graphs to predict RUL values.

\subsection{Taxonomy}
\begin{figure*}[b]
    \centering
    \includegraphics[width = 1.\linewidth]{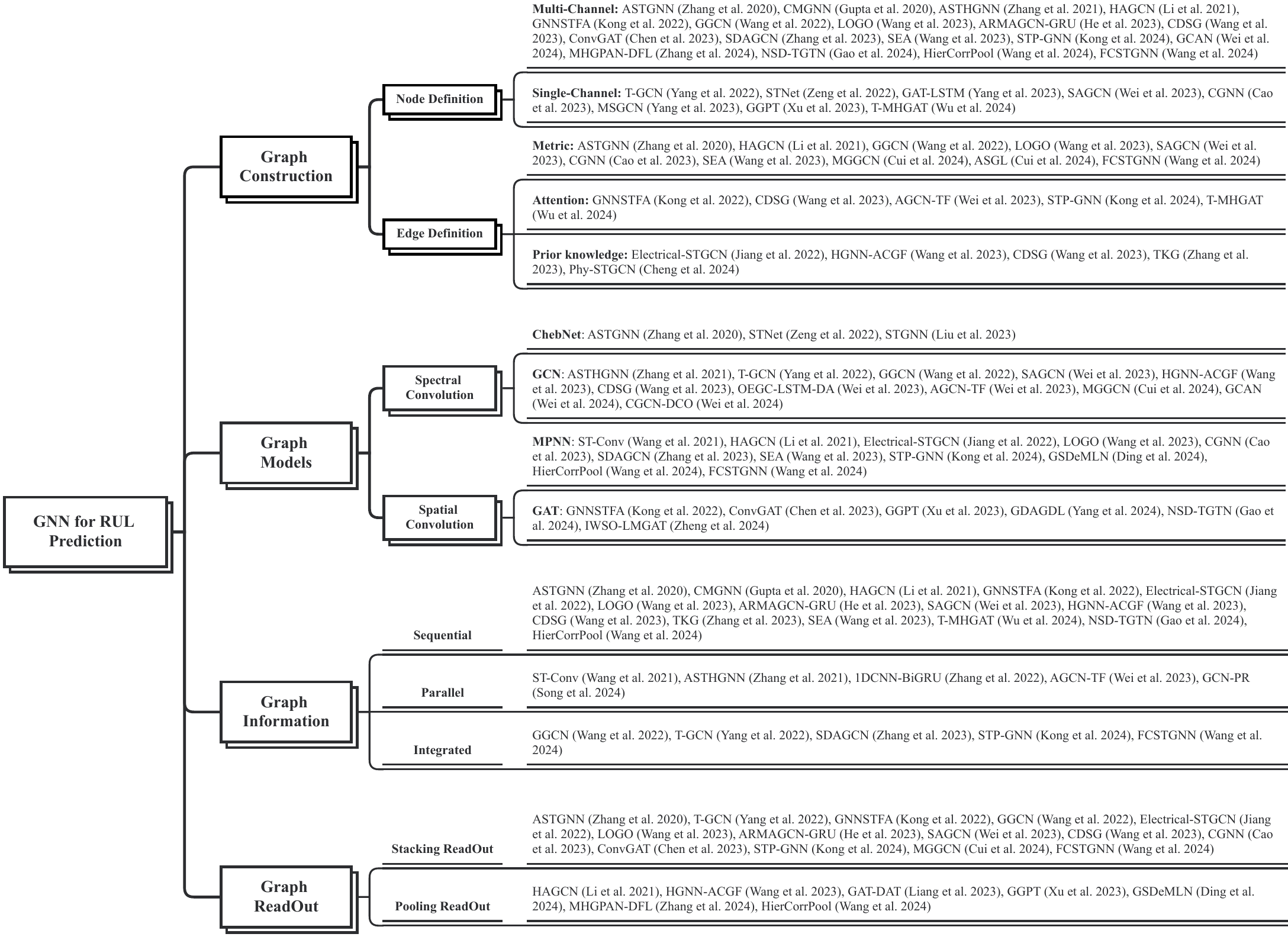}
    \caption{The taxonomy of existing GNN approaches for RUL prediction.}
    \label{fig:overall_literature}
\end{figure*}

To better adapt GNNs for RUL prediction, existing works have made significant contributions at different stages of the workflow. \textcolor{black}{Based on this workflow, we introduce a taxonomy, as shown in Fig. \ref{fig:overall_literature}, to highlight these contributions.} This taxonomy includes graph construction, graph models, graph information, and graph readout functions, all of which are critical components to impact GNNs' performance for RUL prediction. \textcolor{black}{Each of these components is further subdivided into fine-grained categories, with a critical analysis provided for each, evaluating the advantages and disadvantages of the various approaches within each component. Notably,} temporal encoders are not included in this taxonomy, as we focus on summarizing the contributions of GNNs for RUL prediction.

\begin{itemize}
  \item \textbf{Graph Construction:} This is the foundational step to explicitly represent the spatial-temporal information within the data. It involves how existing works convert time-series data into graphs by defining nodes and edges.
  \item \textbf{Graph Models:} This stage focuses on capturing the spatial information within the graphs. It explores the details of GNN models commonly used in RUL prediction, including both spectral and spatial convolutional techniques, and examines how existing works apply these models for effective RUL prediction.
  \item \textbf{Graph Information:} This refers to the spatial-temporal dependency information within the graphs. Effectively utilizing this information through graph models and temporal encoders is crucial for improving RUL prediction. This part summarizes existing approaches into three categories: sequential, parallel, and integrated methods for capturing this information.
  \item \textbf{Graph ReadOut:} After capturing the information within the graphs, this part aims to aggregate all node features to extract graph-level information for RUL prediction. This section reviews the graph readout functions used in this area, including stacking and pooling readout functions.
  \end{itemize}

\subsection{Graph Construction}
Given the absence of explicit graphs to represent spatial-temporal information in RUL prediction, graphs are typically constructed before applying GNNs. With a sample $\mathbf{X}$, nodes and edges are defined to establish the graph structure. Notably, creating effective graphs for GNN-based RUL prediction necessitates considering the characteristics of the data to accurately define nodes and edges. In the subsequent sections, we delve into the process of graph construction in existing works by detailing the definitions of nodes and edges tailored specifically for this domain.

\subsubsection{Nodes}
Node definition involves specifying the properties of nodes $V = \{v_n\}_n^N$, where $N$ represents the number of nodes in the graph, and the features of these nodes \textcolor{black}{are represented as} $\{\mathbf{z}_n\}$. \textcolor{black}{Based on the number of available channels, current research to define nodes can be categorized as multi-channel-based and single-channel-based approaches.}

\paragraph{Multi-Channel}
\begin{figure}[b]
    \centering
    \includegraphics[width = .5\linewidth]{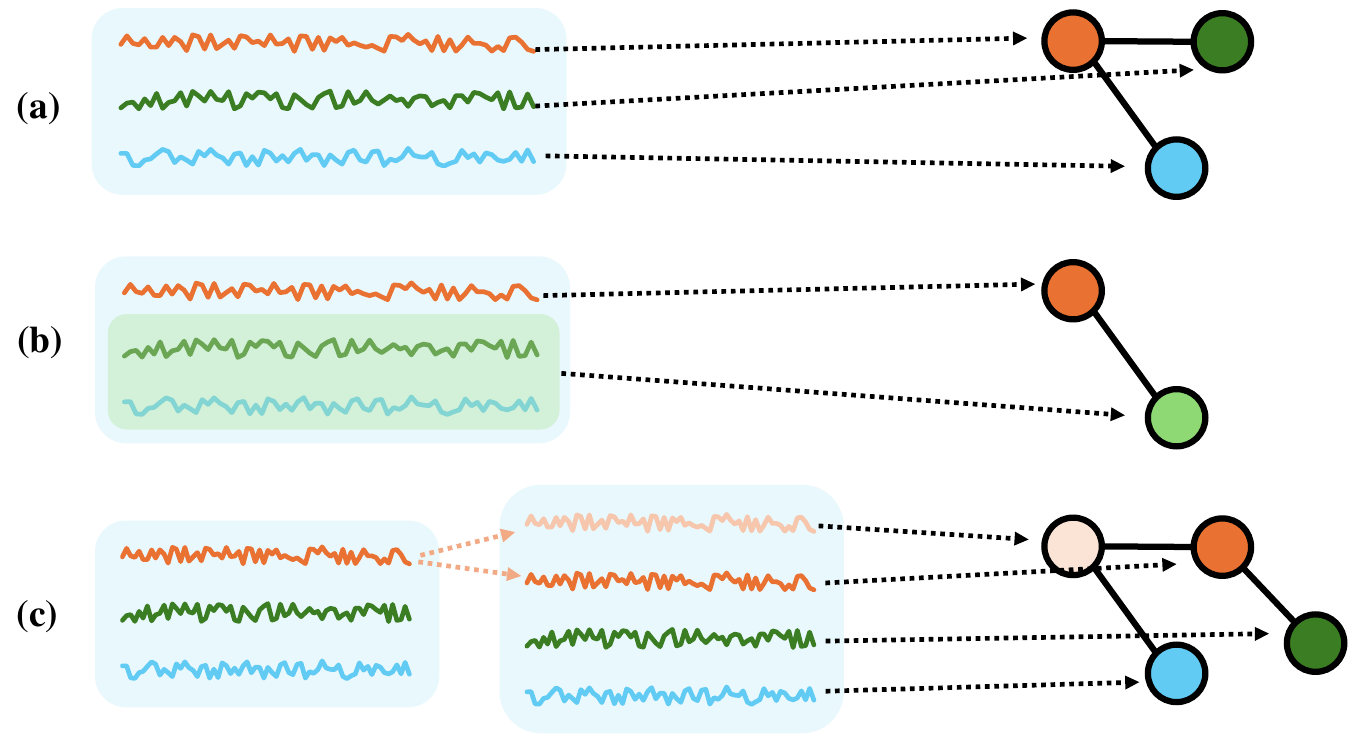}
    \caption{(a) Each channel as a node; (b) Channel clusters as a node; (c) Channel expansion.}
    \label{fig:multi_channel}
\end{figure}
When monitoring complex machines like turbofan engines in aircraft, multiple sensors are typically deployed to measure various parameters such as fan speed, temperature, and pressure. These sensors can be considered as different channels, and nodes in graphs are defined based on these channels to represent spatial dependencies among these physical parameters. In the case of multiple channels, three main approaches are commonly adopted to define nodes, as shown in Fig. \ref{fig:multi_channel}, \textcolor{black}{including each channel as a node, channel clusters as a node, and channel expansion.}

First, channels can directly correspond to nodes, where $S=N$, indicating that the number of nodes equals the number of channels. Zhang et al. \citep{zhang2020adaptive_1} applied GNN to turbofan engines with 14 sensors. The signals of each sensor were processed by CNN to extract temporal features, which were then defined as nodes in the graph. This pioneering work has inspired numerous subsequent studies in GNN-based RUL prediction \citep{zhang2021adaptive_4,wang2023local_15,wang2023multivariate_59,wang2024fully_60}. \textcolor{black}{Defining each sensor channel as a node in graph construction is intuitive and effective, particularly in cases where each channel contains important information. However, too many nodes can lead to high-dimensional graphs that might be computationally expensive to process. Meanwhile, they also struggle to capture the hierarchical structures among channels, leading to underperformance in scenarios with such complexities.}

\textcolor{black}{By capturing the hierarchical structures via clustering certain channels to define a node, the second approach can address the above problems.} For example, Wang et al. \citep{wang2023comprehensive_19} studied turbofan engines equipped with 14 sensors distributed across components. They observed that the sensors within the same component present similar properties. Thus, they proposed fusing sensors within each component and defining components rather than individual sensors as nodes for graph construction. Similarly, Jiang et al. \citep{jiang2022electrical_13} encountered a scenario where data from 15 channels were collected, each providing three values related to five physical parameters at each timestamp. They aggregated values from the same parameters and used each parameter as a node for constructing the graph. \textcolor{black}{These approaches tend to perform better in cases where certain sensors exhibit similar properties. However, they may not be applicable in situations with only a limited number of sensors, as insufficient sensor data can hinder the ability to capture hierarchical structures effectively.}

\textcolor{black}{The third approach was proposed to address the challenges posed by limited sensor sources. In scenarios where only a few sensors are available, such as monitoring battery status with measurements like voltage, current, and temperature \citep{wei2024remaining_44}, constructing a graph with just a few channels may not fully capture the spatial information within the system. This limitation can hinder the performance of GNNs for RUL prediction.} To overcome this issue, researchers proposed to expand these channels for node definitions. For example, Wei et al. \citep{wei2023prediction_25} proposed to expand the number of channels by extracting additional temporal features. For instance, features like `Time to the maximum voltage in charging' and `Time charged under constant voltage mode' were derived, effectively expanding the initial three channels into eight. These additional channels provided richer node definitions for more robust graph construction. \textcolor{black}{While these approaches offer solutions to address limited sensor sources, they are sensitive to the quality of the available channels. Therefore, it would be practical to develop methods that better handle scenarios with limited sensor data.}

\begin{figure}[b]
    \centering
    \includegraphics[width = .5\linewidth]{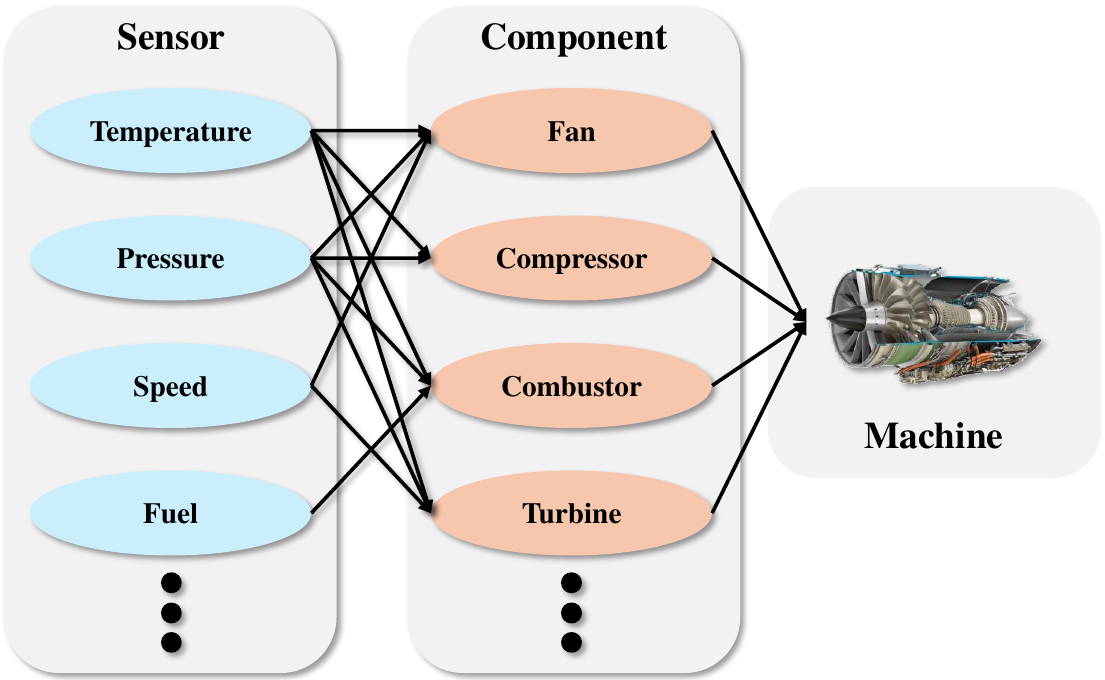}
    \caption{Hierarchical structures within multi-channel cases \citep{wang2023comprehensive_19}.}
    \label{fig:componentasnodes}
\end{figure}
The aforementioned approaches primarily define nodes based on features extracted from each channel, which may not fully capture nonlinear relationships between channels. To tackle this limitation, Chen et al. \citep{chen2023convolution_24} introduced learnable embedding vectors for each node. These embeddings are utilized in graph construction to enhance the nonlinear capabilities of nodes, thereby improving the effectiveness of GNNs in learning complex relationships. Similarly, Zhang et al. \citep{zhang2023spatial_28} also introduced learnable node embeddings for graph construction. In contrast to the prior work \citep{chen2023convolution_24}, they focused on using these embeddings primarily for computing the adjacency matrix, while still learning node features from signal data. Notably, these works assumed all channels are available, which may not always be the case in the real-world system. To address the challenge of missing channels, Gupta et al. \citep{gupta2020handling_2} proposed to use only available channels for node definition in graph construction. This approach harnesses the flexibility of GNNs to effectively handle graphs with varying node configurations, accommodating scenarios with missing sensor data.

\paragraph{Single-Channel}
In applications where space constraints limit the installation of multiple sensors, such as in compact machinery like bearings, leveraging GNNs for improved RUL prediction from single-channel data has been explored through various approaches. One perspective involves using graphs to model long-range temporal dependencies between timestamps in time-series data $X = \{x_1, x_2, ..., x_L\}$. Traditional temporal encoders like TCNs and CNNs struggle with capturing these dependencies over long periods. By folding the time-series signals to form a graph, GNNs provide a solution to model the long-range dependencies. Wei et al. \citep{wei2023bearing_17} proposed segmenting time-series signals into patches, where each patch is defined as a node for graph construction. Similarly, Cao et al. \citep{cao2023picture_20} adopted a similar approach, segmenting time-series into local temporal periods and using fully connected layers to define nodes. However, predefined segment sizes in these methods may impair capturing periodic properties across timestamps. To overcome these limitations, Yang et al. \citep{yang2022bearing_8} introduced learning shapelets directly from time-series signals and using them as nodes, thereby capturing temporal patterns more flexibly. Alternatively, Yang et al. \citep{Yang2023path_14} proposed a pathgraph approach where each timestamp serves as a node, accommodating dynamic temporal dependencies. \textcolor{black}{While these approaches effectively leverage GNNs to capture temporal dependencies, they fall short in capturing the crucial spatial information within the single-channel data.}

\textcolor{black}{To capture the spatial information within the data,} another strategy is to transform single-channel data into multi-channel representations \citep{song2024remaining_45}. Zeng et al. \citep{zeng2022remaining_11} utilized Short-Time Fourier Transform (STFT) to extract frequency components from time-series signals. Each frequency component was then defined as a node, with node features representing temporal amplitudes. Li et al. \citep{li2021remaining_6} and Shen et al. \citep{shen2023graph_34} extracted temporal statistical features from time-series signals, defining nodes with these statistics to model dependencies within statistical characteristics. Zhang et al. \citep{zhang2024multi_51} combined the benefits of frequency components and temporal statistics, generating more channels compared to the above methods. Furthermore, recognizing the limitations of these static feature extraction methods, Yang et al. \citep{yang2023bearing_22} introduced Spectral Energy Difference (SED). By transforming signals into the frequency domain and calculating differences in frequency spectrum over time, they defined nodes based on temporal periods, with features representing SED to capture state change information during degradation processes. \textcolor{black}{Although these approaches have demonstrated effectiveness in capturing spatial information within single-channel data for RUL prediction, they still face the challenge of channel expansion, particularly due to their sensitivity to channel quality. Therefore, enhancing channel quality and subsequently expanding the single channel could be a promising solution.}

\subsubsection{Edges}
Once nodes are defined, the next step involves establishing connections between them through edges to model their structural information. This is achieved by learning the adjacency matrix $\mathbf{A}$ which defines the relationships between nodes $V = \{v_n\}_n^N$. Specifically, this part requires designing a function $\mathbf{A} = \mathcal{F}(Z)$, where ${A}_{nm} = \mathcal{F}(\mathbf{z}_n, \mathbf{z}_m)$, aiming to learn edges by considering the properties between nodes. Currently, most methods for learning the adjacency matrix can be categorized into three main types: metric-based, attention-based, and prior-knowledge-based.

\paragraph{Metric-based}
Metric-based methods aim to establish connections between nodes based on the similarity or distance metrics computed from their feature representations. Here are six commonly used metric-based methods for learning the adjacency matrix in graph construction: dot-product distance, cosine similarity distance, Pearson Correlation Coefficient (PCC), Euclidean distance, Gaussian kernel weight function, and generalized Mahalanobis distance:

\textbf{Dot-Product Distance:}
\begin{equation}
    \label{eq:dotproduct}
    A_{nm}^{DP} = \mathbf{z}_n \cdot \mathbf{z}_m = \sum_{i=1}^{d} z_{n,i} \cdot z_{m,i}
\end{equation}
where $z_{n,i}$ and $z_{m,i}$ represents the $i$-th feature of channels $n$ and $m$ respectively.

\textbf{Cosine Similarity Distance:}
\begin{equation}
    \label{eq:cosine}
    A_{nm}^{CS} = \frac{\mathbf{z}_n \cdot \mathbf{z}_m}{\|\mathbf{z}_n\| \|\mathbf{z}_m\|}
\end{equation}

\textbf{Euclidean Distance:}
\begin{equation}
    \label{eq:euclidean}
    A_{nm}^{ED} = \frac{1}{\sqrt{\sum_{i=1}^{d} (z_{n,i} - z_{m,i})^2}}
\end{equation}

\textbf{Pearson Correlation Coefficient:}
\begin{equation}
    \label{eq:pcc}
    A_{nm}^{PCC} = \frac{\sum_{i=1}^{d} (z_{n,i} - \overline{z}_m)(z_{m,i} - \overline{z}_n)}{\sqrt{\sum_{i=1}^{d} (z_{n,i} - \overline{z}_n)^2} \sqrt{\sum_{i=1}^{d} (z_{m,i} - \overline{z}_m)^2}}
\end{equation}
where $\overline{z}_n$ and $\overline{z}_m$ are the mean values of node features $\mathbf{z}_n$ and $\mathbf{z}_m$ respectively.

\textbf{Gaussian Kernel Weight Function:}
\begin{equation}
    \label{eq:gaussian}
    A_{nm}^{GK} = \exp\left(-\frac{\|\mathbf{z}_n - \mathbf{z}_m\|^2}{2\sigma^2}\right)
\end{equation}
where $\|\mathbf{z}_n - \mathbf{z}_m\|$ is the Euclidean distance between $\mathbf{z}_n$ and $\mathbf{z}_m$, and $\sigma$ is a parameter that determines the width of the Gaussian kernel.

\textbf{Generalized Mahalanobis Distance:}
\begin{equation}
    \label{eq:GMD}
    A_{nm}^{MD} = \sqrt{ (\mathbf{z}_n - \mathbf{z}_m)^T \mathbf{S}^{-1} (\mathbf{z}_n - \mathbf{z}_m) }
\end{equation}
where $\mathbf{S}^{-1}$ is the inverse of the covariance matrix of the node features.

Most studies rely on these metric-based approaches to construct the adjacency matrix. Commonly used distance metrics include dot-product, cosine similarity, and Euclidean distance, which are favored for their simplicity and applicability across various datasets \citep{yang2023bearing_22,zhang2023spatial_28,xing2023stcgcn_29,wang2022gated_10,wang2023local_15,wang2023sensor_39,wang2024fully_60,chen2023convolution_24,cui2024adaptive_55}. However, Wang et al. \citep{wang2021spatio_3} highlighted limitations in using Euclidean distance alone, arguing it may not adequately represent close relationships between sensors. Thus, they proposed using PCC to compute the adjacency matrix, which captures more nuanced sensor relationships, and this approach has influenced subsequent research \citep{li2021remaining_6,he2023systematic_16,song2024remaining_45}. Beyond these, Gaussian kernel weight function \citep{liu2023condition_31,ding2024graph_48,xiao2024multi} and generalized Mahalanobis distance \citep{long2023gnnrotating_33} have been utilized for their ability to model complex, nonlinear relationships. However, these methods are less explored compared to other distance metrics.

With these distances, most works simply constructed fully-connected graphs, where every node is connected to every other node, resulting in dense adjacency matrices. However, researchers have recognized that not all edges in these graphs may provide meaningful information for GNN-based RUL prediction. To address this concern, several filtering-based methods have been developed, including top-\underline{k} \citep{lv2023new_27,liu2023condition_31}, K-Nearest Neighbor (KNN) \citep{cui2024digital_43,wu2024temporal_46}, and threshold-based methods \citep{wang2022gated_10,liang2023remaining_23,wei2023prediction_25,wang2024dvgtformer_54,xiao2024multi}. Chen et al. \citep{chen2023convolution_24}, using cosine similarity to compute connections , filtered out weak connections by selecting only the top-\underline{k} edges per node that were deemed most relevant. Xu et al. \citep{xu2023novel_35} employed KNN to select neighbors that are then connected to form the graph, focusing on local relationships. Wei et al. \citep{wei2023bearing_17} predefined a threshold to eliminate weak connections in their graph construction process.

In addition to single metric approaches, some methods leveraged multiple metrics to obtain the better adjacency matrix. Xing et al. \citep{xing2023stcgcn_29} proposed combining cosine similarity and Euclidean distances using an adjustable hyperparameter to achieve an improved adjacency matrix. Wang et al. \citep{wang2023local_15} introduced a method involving both global and local graphs to capture different aspects of sensor correlations. The global graph was constructed using PCC, while local graphs were built based on dot-product distances. These graphs were then fused together using an adaptive fusion mechanism to integrate global and local information effectively. Jiang et al. \citep{jiang2022electrical_13} constructed two graphs with the same node features but employing different graph construction methods. One graph was constructed with Euclidean distance, while another graph was built by introducing a symmetric-normalized Laplacian matrix to mitigate oversmoothing issues during graph learning. Xiao et al. \citep{xiao2024multi} constructed two graphs, a RadiusGraph and a PathGraph with cosine similarity and Gaussian Kernel Weight function respectively, to fully capture the local feature similarity and long-term trends within signals.

Most of above methods directly applying metrics to compute the adjacency matrix, which may not fully capture the complex relationships among nodes. Thus, researchers improved the metric-based graph construction. Zhang et al. \citep{zhang2020adaptive_1} observed limitations in using Euclidean distance within vanilla Gaussian kernel weight functions for graph-structured data. To address this, they introduced the Generalized Mahalanobis distance as an alternative metric, aiming to improve the Gaussian kernel weight function for constructing an improved adjacency matrix. Wei et al. \citep{wei2023bearing_17} proposed a dynamic graph construction method. An initial graph was constructed with cosine similarity distances, and then an randomly initialized adjacency matrix was introduced to act as a transition matrix to transform the initial graph into a dynamic structure. These authors \citep{wei2023prediction_25} further developed a two-stage graph construction method. A first-stage graph was constructed with the covariance matrix of node features. These node features are meanwhile used to derive the mutual information of temporal features. The mutual information and the first-stage graph are then fed into a graph entropy optimization model to construct the second-stage graph by maximizing entropy and minimizing density simultaneously. 

\textcolor{black}{Metric-based approaches offer several advantages. First, they are straightforward to implement and interpret, as they rely on well-known mathematical distances or similarities. Second, they are generally computationally efficient, making them scalable for large datasets. However, these approaches may struggle to capture complex relationships between nodes that are not adequately reflected by simple distance metrics. For instance, non-linear dependencies or hidden patterns might be overlooked. Additionally, the performance of the graph heavily depends on the choice of metric, which may not always be well-suited to the data at hand.}

\paragraph{Attention-based}
\textcolor{black}{By allowing models to learn node relationships, attention-based methods can well address the mentioned problems.} These methods involve three steps: a linear transformation to increase the non-linear expressiveness of each node, calculation attention coefficients to compute weights between nodes, and normalization to restrict connection weights within [0,1]. Here, $\mathcal{N}(n)$ represents the neighbor nodes of the node $n$, and $f_{att}(\mathbf{W} \mathbf{z}_n, \mathbf{W} \mathbf{z}_m)$ is normally achieved by $\mathbf{W}_a^T[\mathbf{W} \mathbf{z}_n||\mathbf{W} \mathbf{z}_m]$, where $[a||b]$ represents the concatenation operation. It is noted that the attention coefficient part is the core part to obtain edges for node connections.
\begin{align}
\label{eq:atten}
\text{Linear Transformation:} & \quad \mathbf{z}'_n = \mathbf{W} \mathbf{z}_n \\
\text{Attention Coefficient:} & \quad a_{nm} = f_{att}(\mathbf{W} \mathbf{z}_n, \mathbf{W} \mathbf{z}_m) \\
\text{Normalization:} & \quad A_{nm} = \frac{\exp(a_{nm})}{\sum_{k \in \mathcal{N}(n)} \exp(a_{nk})}
\end{align}

Currently, attention-based methods have shown effective for graph construction. For example, Wei et al. \citep{wei2023remaining_26} employed the self-attention mechanism to compute the adjacency matrix from node features, aiming to adaptively generate graph edges. Furthermore, some researchers have leveraged attention as an auxiliary task. Zeng et al. \citep{zeng2022remaining_11} calculated the average and maximum values of each node and concatenated them to determine the attention weight for each node. Nodes with attention weights above a predefined threshold were then connected to form the adjacency matrix. Wang et al. \citep{wang2023comprehensive_19} introduced an attention-based framework to learn masks for edge filtering. Initially, they computed the adjacency matrix using node features via a fully connected layer and then derived a masking matrix through node-to-node attentions. The masking matrix was integrated with the graph edges using the Hadamard product to filter edges. Notably, attention-based methods typically involve more trainable weights, which can increase the demand for training data. To address this, Gupta et al. \citep{gupta2020handling_2} simplified the attention mechanism by eliminating linear transformation weights, thereby reducing the number of trainable parameters and the dependency on extensive training data.
\textcolor{black}{Although attention-based methods show promise in fully capturing complex and non-linear relationships between nodes, they still suffer when dealing with limited data. Unlike metric-based approaches that rely on predefined metrics, attention-based methods learn node relationships solely from data. When data is scarce, these methods struggle to learn effective graph structures, making them more prone to overfitting.}

\paragraph{Prior-Knowledge-based}
\textcolor{black}{The metric-based and attention-based approaches primarily rely on data for graph construction. However, data bias and noise can adversely impact graph construction, leading to inaccurate graphs that affect GNN-based RUL prediction. To mitigate this issue, some researchers have proposed incorporating prior-knowledge for enhanced graph construction.} Most existing prior-knowledge-based approaches utilize knowledge of physical parameters or physical locations. Kong et al. \citep{kong2022spatio_9} defined the connections of nodes with their physical relations. Jiang et al. \citep{jiang2022electrical_13} collected data from 15 channels of an electrical system. They constructed graphs by classifying these channels according to physical parameters, such as power factor, reactive power, current. Wang et al. \citep{wang2023comprehensive_19} introduced a component graph representing the physical connections between engine components, where component nodes are connected if the corresponding components are physically connected in the engine. Wang et al. \citep{wang2023hierarchical_18} proposed using both sensor and module graphs to capture multi-level spatial information, where module graph nodes are defined based on modules described in preexisting documents, and node features are derived by averaging the sensor features within each module.

Furthermore, Knowledge Graphs (KG) have proven effective in introducing prior-knowledge by defining triplet relations among entities, such as mechanical components, fault symptoms, and causes \citep{xia2023maintenance_38}. Zhang et al. \citep{zhang2023temporal_30} leveraged KGs to improve graph construction by defining a temporal knowledge graph, which was formed by triplet links from components to sensors according to their physical relations. For example, a triplet like [`Fan', `Cooling', `Combustion chamber'] was defined to represent the semantic relationship `Fan cools the combustion chamber'. Additionally, knowledge of physical models can provide valuable guidance for graph construction. Cheng et al. \citep{cheng2024research_57} designed a graph based on physical knowledge in gas turbine engines, specifically utilizing the Small Deviation (SD) theory and structural analysis. The SD theory helps organize the mechanism equations of the gas turbine, retaining only terms containing monitoring parameters while eliminating all others. These collated SD equations are then structurally analyzed to create a structural decomposition diagram, clarifying the correlations between parameters.

\textcolor{black}{Prior-knowledge can provide valuable domain-specific insights for accurately modeling graph structures. Additionally, as the prior-knowledge is based on known relationships or physical rules, graphs built on prior-knowledge are often more interpretable, making it easier to understand the model's decisions. However, relying on prior-knowledge can also limit the generalizability of the model, as this knowledge is typically domain-specific. Additionally, the need for expert input to define such prior-knowledge can be a drawback, as it might not always be readily available.}

\paragraph{Others}
Some works employ unique methods for graph construction that differ from previously discussed approaches. Yang et al. \citep{yang2022bearing_8} defined nodes using shapelets, where the relationship between two nodes is determined by the probability of one shapelet appearing after another. If one shapelet is more likely to follow another, their correlation is considered high, and the graph edge is represented by these probabilities. Wu et al. \citep{wu2024temporal_46} designed a hypergraph by computing hyperedges for nodes, connecting multiple nodes to model high-order relationships. Specifically, KNN is used to filter close samples, and a hypergraph with \( m \) hyperedges is produced by grouping these samples with \( K \) neighboring samples in Euclidean space to form hyperedges. Multi-resolution hypergraphs are also generated to enhance the effectiveness of the hypergraph.

Graphs have been widely utilized in RUL prediction through various approaches \citep{zhuang2024graph_56}. Li et al. \citep{li2023remaining_36} designed a knowledge graph constructed with sensor relationships, serving as guidance for obtaining sensor clusters. These clusters are then processed by different deep learning models for effective sensor feature extraction. Cui et al. \citep{cui2024adaptive_55} developed an adaptive sparse graph learning method to automatically learn adjacency relationships. This method defines a graph learning optimization, optimized using the Lagrange multiplier method, and incorporates a regularization term to achieve sparse graphs.

\subsection{Graph Models}
Graph models aim to capture spatial information from the constructed graphs for improved RUL prediction. In this part, we introduce the most widely used GNN models, including spectral convolution and spatial convolution methods, and meanwhile, how they have been applied to RUL prediction.

\subsubsection{Spectral Convolution}
Spectral graph convolution pioneers the research to capture the information within graph-structured data by extending the concept of convolution to graphs, enabling powerful neural network architectures for non-Euclidean data. Spectral graph convolution is based on the graph Fourier transform. Consider the constructed graph \( \mathcal{G} = \{V, E\} \) with \( N \) nodes and an adjacency matrix \( \mathbf{A} \). The graph Laplacian \( \mathbf{L} \) is defined as \( \mathbf{L} = \mathbf{D} - \mathbf{A} \), where \( \mathbf{D} \) is the degree matrix. The normalized graph Laplacian is \( \mathcal{L} = \mathbf{I} - \mathbf{D}^{-1/2} \mathbf{A} \mathbf{D}^{-1/2} \).

The eigen decomposition of \( \mathcal{L} \) is \( \mathcal{L} = \mathbf{U} \mathbf{\Lambda} \mathbf{U}^\top \), where \( \mathbf{U} \) is the matrix of eigenvectors and \( \mathbf{\Lambda} \) is the diagonal matrix of eigenvalues. The graph Fourier transform of a signal \( \mathbf{Z} \) is defined as \( \hat{\mathbf{Z}} = \mathbf{U}^\top \mathbf{Z} \), and its inverse is \( \mathbf{Z} = \mathbf{U} \hat{\mathbf{Z}} \).

A spectral convolution operation on the graph is defined as:
\begin{equation}
    g_\theta * \mathbf{Z} = \mathbf{U} g_\theta(\mathbf{\Lambda}) \mathbf{U}^\top \mathbf{Z}
\end{equation}
where \( g_\theta(\mathbf{\Lambda}) \) is a filter applied in the spectral domain.

\paragraph{ChebNet (Chebyshev Spectral Graph Convolution)}

ChebNet, introduced by Defferrard et al. \citep{defferrard2016convolutional}, approximates spectral graph convolution using Chebyshev polynomials to avoid the computationally expensive eigen decomposition. The filter \( g_\theta(\mathbf{\Lambda}) \) is approximated by a truncated expansion of Chebyshev polynomials \( T_k(\mathcal{L}) \):
\begin{equation}
g_\theta(\mathcal{L}) \approx \sum_{k=0}^K \theta_k T_k(\tilde{\mathcal{L}})
\end{equation}
where \( \tilde{\mathcal{L}} = 2 \mathcal{L} / \lambda_{\max} - \mathbf{I} \), \( \lambda_{\max} \) is the largest eigenvalue of \( \mathcal{L} \), and \( T_k \) is the Chebyshev polynomial of degree \( k \). The convolution operation becomes:
\begin{equation}
g_\theta * \mathbf{Z} \approx \sum_{k=0}^K \theta_k T_k(\tilde{\mathcal{L}}) \mathbf{Z}
\end{equation}

This formulation allows for efficient computation of convolutions on large graphs by leveraging the recursive definition of Chebyshev polynomials.

\paragraph{Graph Convolutional Networks (GCNs)}

GCNs, proposed by Kipf and Welling \citep{kipf2016semi}, simplify the spectral graph convolution further by restricting the filters to be first-order polynomials. The convolution operation in a GCN is:
\begin{equation}
g_\theta * \mathbf{Z} \approx \theta_0 \mathbf{I} \mathbf{Z} + \theta_1 \mathcal{L} \mathbf{Z} = \theta (\mathbf{I} + \mathbf{D}^{-1/2} \mathbf{A} \mathbf{D}^{-1/2}) \mathbf{Z}
\end{equation}
For computational efficiency and to avoid numerical instabilities, they approximate \( \lambda_{\max} \) by 2, leading to the following simplified convolution:
\begin{equation}
\mathbf{Z}^{(l+1)} = \sigma (\tilde{\mathbf{D}}^{-1/2} \tilde{\mathbf{A}} \tilde{\mathbf{D}}^{-1/2} \mathbf{Z}^{(l)} \mathbf{W}^{(l)})
\end{equation}
where \( \tilde{\mathbf{A}} = \mathbf{A} + \mathbf{I} \) is the adjacency matrix with added self-loops, \( \tilde{\mathbf{D}} \) is the degree matrix of \( \tilde{\mathbf{A}} \), \( \mathbf{Z}^{(l)} \) is the feature matrix at layer \( l \), \( \mathbf{W}^{(l)} \) is the layer-specific trainable weight matrix, and \( \sigma \) is an activation function such as ReLU.

Spectral graph convolution has paved the way for powerful neural network architectures on graph-structured data. ChebNet introduces an efficient approximation using Chebyshev polynomials, while GCNs further simplify and improve computational efficiency, making them highly effective for various graph-based tasks. Due to their effectiveness, ChebNet and GCNs have been widely used to capture spatial information for RUL prediction. 

Initially, ChebNet was utilized by Zhang et al. \citep{zhang2020adaptive_1} for RUL prediction, where the adjacency matrix computed using generalized Mahalanobis distance was convolved by ChebNet. Following this, Zeng et al. \citep{zeng2022remaining_11} employed three layers of ChebNet to capture information within graphs, demonstrating the effectiveness of deep ChebNet for RUL prediction. Liu et al. \citep{liu2023condition_31} further validated the benefits of capturing wider spatial fields from high-order neighbor information by utilizing a three-order ChebNet. \textcolor{black}{However, choosing the appropriate polynomial degree and managing approximation errors can be complex. Thus, GCNs have become popular alternatives to ChebNet for spectral convolutions due to the simple implementation and improved computational efficiency} \citep{li2021remaining_6,yang2022bearing_8,wang2023hierarchical_18,zhu2023rgcnu_21,wei2023prediction_25, xing2023stcgcn_29,shen2023graph_34,cui2024digital_43,wei2024state_53,cheng2024research_57}. For instance, Zhang et al. \citep{zhang2021adaptive_4} extended their previous work \citep{zhang2020adaptive_1} by replacing ChebNet with GCNs for easier implementation. Wei et al. \citep{wei2023bearing_17,wei2023remaining_26} utilized GCNs and introduced bias weight vectors to enhance the nonlinear expressiveness of the model, leading to improved performance.

GCNs apply Laplacian smoothing during convolutions, which can lead to oversmoothing of node features across the graph, potentially losing important initial features over layers. To address this issue and enhance RUL prediction performance with GCNs, researchers have explored various strategies to preserve and integrate the information from initial layers. Zhang et al. \citep{zhang2021adaptive_4} introduced skip connections to inject initial information directly into the learning process. Similarly, Wang et al. \citep{wang2023comprehensive_19} implemented residual connections, with a hyperparameter controlling the influence of the initial input data. Lv et al. \citep{lv2023new_27} devised a message-selection mechanism to determine how much information from previous GNN layers should be retained. Additionally, Wei et al. \citep{wei2024remaining_44} designed a graph convolutional attention network, integrating an attention mechanism between every two spectral graph convolutional operations. This approach aims to mitigate potential adverse impacts from less significant features when integrating information from initial layers.

Furthermore, researchers \citep{yang2023bearing_22,long2023gnnrotating_33,liu2023aero_37,zhang2024multi_51} have explored leveraging information learned across multiple GNN layers instead of solely relying on information from initial layers to learn hidden features. For example, Yang et al. \citep{yang2023bearing_22} proposed a multi-scale GCN approach that concatenates GCN representations from different layers. This method mitigates the oversmoothing problem while incorporating spatial information across different scales for RUL prediction. Additionally, He et al. \citep{he2023systematic_16} introduced Auto-Regressive Moving Average (ARMA) filters to enhance GCNs. ARMA filters employ a weighted average mechanism to enhance the model's capability to capture and retain more relevant information across the entire learning period. This gradual weighting of graph information within the time series helps mitigate the risk of oversmoothing.

\textcolor{black}{Spectral convolution are good at capture global information due to their ability to operate in the frequency domain. Meanwhile, they were developed based on graph signal processing, providing a theoretical framework for understanding graph convolutions. However, transforming data into the frequency domain and back can be computationally intensive, especially for large graphs, which is impractical in real world systems.}

\subsubsection{Spatial Convolution}
By directly operating on the graphs and aggregating information from a node's neighbors, spatial graph convolution effectively addresses the limitations of spectral methods. This approach models spatial relationships among nodes without relying on spectral properties, facilitating efficient and effective learning on graph-structured data. Two prominent spatial graph convolution approaches are Message Passing Neural Networks and Graph Attention Networks.

\paragraph{Message Passing Neural Networks (MPNN)}
MPNNs, introduced by Gilmer et al. \citep{gilmer2017neural}, generalize various graph neural networks by defining a framework for message passing and node update.
During the message passing phase, each node \( v_n \in V \) aggregates messages from its neighbors. For \( T \) message passing steps, the messages \( \mathbf{h}_n^{(t)} \) and node states \( \mathbf{z}_n^{(t)} \) are updated as follows:
\begin{equation}
    \begin{aligned}
        \mathbf{h}_n^{(t+1)} &= \sum_{k \in \mathcal{N}(n)} M_t(\mathbf{z}_n^{(t)}, \mathbf{z}_k^{(t)}, {A}_{nk})\\
        \mathbf{z}_n^{(t+1)} &= U_t(\mathbf{z}_n^{(t)}, \mathbf{h}_n^{(t+1)})
    \end{aligned}
\end{equation}
where \( M_t \) is the message function, \( U_t \) is the node update function, \( {A}_{nk} \) represents the element of the adjacency matrix, and \( \mathcal{N}(n) \) denotes the neighbors of node \( n \).

\paragraph{Graph Attention Networks (GAT)}

GATs, introduced by Veličković et al. \citep{velivckovic2017graph}, leverage attention mechanisms to weigh the importance of neighboring nodes adaptively. This allows the model to focus on the most relevant neighbors during the aggregation process.

\textbf{Attention Mechanism:}
In GAT, the attention coefficients \( \alpha_{nm} \) between nodes \( n \) and \( m \) are computed as:
\begin{equation}
\alpha_{nm} = \frac{\exp(\mathrm{LeakyReLU}(\mathbf{a}^\top [\mathbf{W} \mathbf{z}_n || \mathbf{W} \mathbf{z}_m]))}{\sum_{k \in \mathcal{N}(n)} \exp(\mathrm{LeakyReLU}(\mathbf{a}^\top [\mathbf{W} \mathbf{z}_n || \mathbf{W} \mathbf{z}_k]))}
\end{equation}
where \( \mathbf{W} \) is a learnable weight matrix, \( \mathbf{a} \) is the attention vector, \( || \) denotes concatenation, and \( \mathcal{N}(n) \) represents the neighbors of node \( n \).

\textbf{Node Update:}
The node features are updated by computing a weighted sum of the neighbors' features using the attention coefficients:
\begin{equation}
\mathbf{z}_n' = \sigma \left( \sum_{k \in \mathcal{N}(n)} \alpha_{nk} \mathbf{W} \mathbf{z}_k \right)
\end{equation}
where \( \sigma \) is an activation function such as ReLU.

\textbf{Multi-Head Attention:}
To stabilize the learning process, GAT employs multi-head attention, where the above process is repeated \( P \) times with independent attention mechanisms, and the results are aggregated:
\begin{equation}
\mathbf{z}_n' = \parallel_{p=1}^P \sigma \left( \sum_{k \in \mathcal{N}(n)} \alpha_{nk}^p \mathbf{W}^p \mathbf{z}_k \right)
\end{equation}
where \( \parallel \) denotes concatenation.

Spatial graph convolution has enabled significant advancements in GNNs by focusing on the direct relationships between nodes. MPNN provides a flexible framework for message passing and node updates, while GAT introduces an adaptive attention mechanism to weigh the importance of neighboring nodes. These approaches have proven effective for various graph-based tasks, complementing spectral methods in the domain of GNNs.

MPNNs have been extensively applied to capture spatial graph information for RUL prediction, with significant research in this domain \citep{wang2023local_15,wang2023multivariate_59,wang2024fully_60,wang2023sensor_39,jiang2022electrical_13,zhang2023spatial_28,wen2024temporal_40,ma2024transformer_49}. Zhang et al. \citep{zhang2020adaptive_1} pioneered the use of MPNNs for RUL prediction, integrating spectral and spatial graph convolutions to enhance performance. Gupta et al. \citep{gupta2020handling_2} enhanced MPNNs by introducing learnable parameters into the message passing function, improving their nonlinear capabilities. Wang et al. \citep{wang2021spatio_3} simplified MPNNs by removing activation functions and introduced multi-order adjacency matrices for propagation and updates, capturing multi-scale spatial information effectively. Similarly, Cao et al. \citep{cao2023picture_20} and Ding et al. \citep{ding2024graph_48} also introduced multi-order adjacency matrices. While differently, they concatenated adjacency matrices at different orders to incorporate multi-hop nodes into message propagation and updating processes. Notably, MPNNs may suffer from information loss by mapping different neighborhoods to the same representation \citep{li2021hierarchical_5}. To address this, Li et al. \citep{li2021hierarchical_5} and Zhang et al. \citep{zhang2023temporal_30} adopted Graph Isomorphism Network (GIN), a variant of MPNN, employing injective MLPs to ensure distinct representations for different neighborhoods and thereby preserving their unique structures. 

\textcolor{black}{Although MPNNs have shown flexibility offer flexibility for different types of graph data and structures, they update node features based solely on their edges. This can be limiting in scenarios where edge weights are absent, as MPNNs cannot account for the strength of node connections.} To address this limitation, GATs have been developed to incorporate attention mechanisms that consider the weights between nodes, and extensive efforts have been devoted to leverage GATs for RUL prediction \citep{Yang2023path_14,chen2023convolution_24,liu2023mmoe_32,xu2023novel_35,huang2024spatio_42,yang2024dynamic_50,zheng2024improved_58,xiao2024multi}. To enhance the effectiveness of GAT in this domain, several approaches have improved upon the vanilla GAT. Kong et al. \citep{kong2022spatio_9} introduced dropout operations into the attention coefficients learned by GAT, masking attention weights randomly to mitigate overfitting issues. Liang et al. \citep{liang2023remaining_23} implemented multiple GAT layers and incorporated residual connections to pass the GAT output from hidden layers to the final output, aiming to prevent gradient vanishing during backpropagation. Zhou et al. \citep{zhou2024mst_47} addressed GAT's limitation in capturing local spatial information within graphs by proposing a multi-perspective GAT. This approach utilizes a multi-head ExpSparse self-attention mechanism to filter weak attention coefficients, allowing the model to focus more on local nodes relevant to central nodes. Additionally, Gao et al. \citep{gao2024nonlinear_52} enhanced the robustness of GAT for RUL prediction by replacing the LeakyReLU activation function used in vanilla GAT with the Mish activation function, which is known for its nonlinear activation capabilities and network robustness improvements \citep{xiang2023concise}.

\textcolor{black}{Compared to spectral convolution, spatial convolution tends to be more scalable and efficient for large graphs, as it operates locally and avoids complex transformations. Additionally, spatial convolution can easily adapt to changes in graph structure and dynamic data by focusing on local neighborhoods. However, without the grounding in graph signal processing theory, spatial convolution may lack interpretability. Furthermore, it risks overfitting to local noise if the graph is not properly regularized. Thus, it would be promising for GNN-based RUL prediction if the advantages of spectral and spatial convolution methods can be combined.}

\subsection{Graph Information}

\begin{figure}[b]
    \centering
    \includegraphics[width = .6\linewidth]{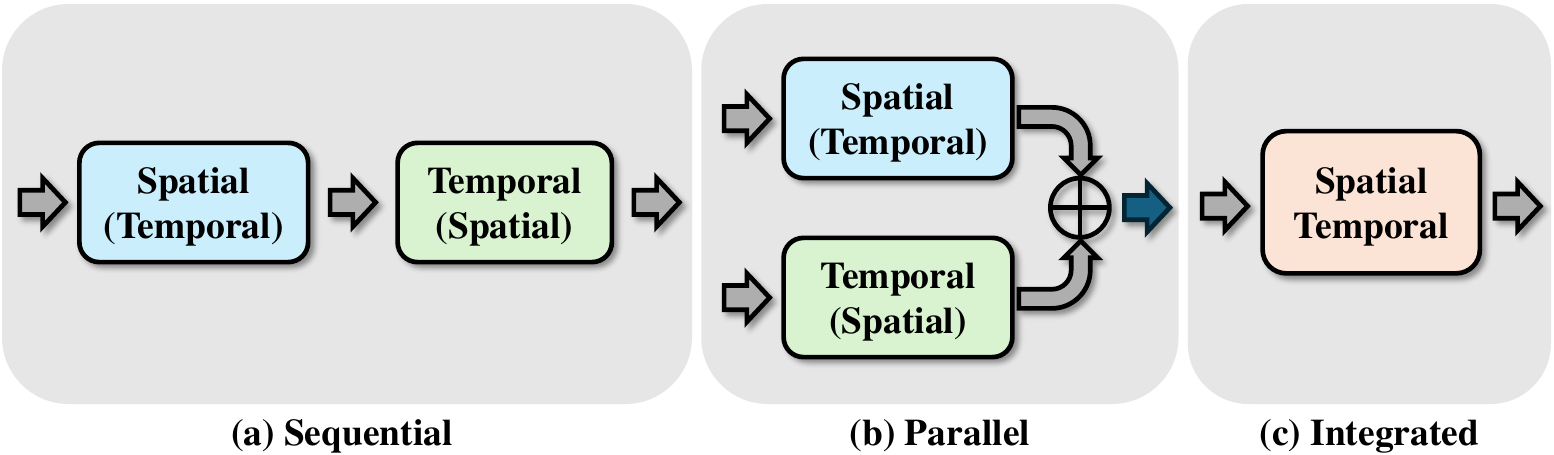}
    \caption{(a) Sequential capturing of spatial and temporal information; (b) Parallel capturing of spatial and temporal information; (c) Capturing integrated spatial-temporal information.}
    \label{fig:graph_info}
\end{figure}
Spatial and temporal information has been widely studied in RUL prediction \citep{xiang2021spatiotemporally_7,zhang2022parallel_12}. With graphs constructed with the aforementioned approaches to model the information, existing can be categorized into three main approaches to capture the information with GNN models, including sequential, parallel, and integrated. The differences of the categories are shown in Fig. \ref{fig:graph_info}. The definitions of these categories, \textcolor{black}{their pros and cons,} and how they are utilized to capture graph information are detailed in the subsequent sections.

\subsubsection{Sequential}

Most existing works capture spatial and temporal information in a sequential scheme, where spatial and temporal information is respectively captured by graph models and temporal encoders, such as CNN \citep{jiang2022electrical_13}, TCN \citep{xing2023stcgcn_29,zhou2024mst_47,li2021remaining_6}, LSTM \citep{Yang2023path_14,yang2024dynamic_50,wang2023local_15,wang2023sensor_39,wei2023prediction_25,kong2022spatio_9}, GRU \citep{wang2023local_15,liu2023condition_31,wu2024temporal_46}, Transformer \citep{ma2024transformer_49,zhang2023temporal_30,zhou2021informer}, etc., for RUL prediction. Based on the order of capturing spatial and temporal information, existing works can be mainly categorized into two approaches: spatial first and temporal first.

Most of the existing works adopt the spatial-first approach, where spatial information is initially captured to learn spatial features, followed by capturing temporal dependencies among these spatial features. Kong et al. \citep{kong2022spatio_9} and Wang et al. \citep{wang2023local_15} constructed sequential graphs where spatial information is first captured within each graph, and then temporal dependencies among these sequential features at the node level are captured using LSTM. Kong et al. \citep{kong2022spatio_9} further enhance this by designing an attention sequence embedding layer, learning attention scores to weight the importance of each sequential graph. Similarly, Wen et al. \citep{wen2024temporal_40} also leveraged MPNN to capture spatial information in each sequential graph. Differently, they designed a novel module to capture temporal information by gathering all historical neighbor information across past time steps in a time-decaying manner. Huang et al. \citep{huang2024spatio_42} capture spatial information using a multi-head spatial attention-based GCN. Subsequently, they introduce a multi-head temporal attention-based TCN to compute attentions for nodes in each sequential graph, effectively capturing temporal dependencies.

In contrast to the above methods mainly capturing temporal information at the node-level, some researchers focus on capturing temporal information using graph-level features after capturing spatial information. Zeng et al. \citep{zeng2022remaining_11} captured spatial information among frequency components in constructed frequency graphs. They employed a readout function to aggregate features for each graph, followed by capturing temporal dependencies among these graph features using BiLSTM. With the same idea, Liang et al. \citep{liang2023remaining_23} utilized a readout function to learn low-level features after capturing spatial information within each graph. These lower-level features were further processed by Transformer to capture temporal dependencies. Similarly, Zhang et al. \citep{zhang2023temporal_30} employed GIN to capture spatial information, followed by a designed pooling operation to learn low-level features. Informer \citep{zhou2021informer} was then adopted to capture temporal information. Additionally, Zhou et al. \citep{zhou2024mst_47} constructed three graphs to represent local and global spatial information. They leveraged GAT to capture the spatial information in each graph. Temporal dependencies among these concatenated graph features were then captured using TCN.

\textcolor{black}{In the spatial-first approach, capturing spatial information serves as a preprocessing stage, allowing for the learning of better features before applying a temporal encoder to capture temporal dynamics. This method is intuitive and straightforward for introducing GNNs into RUL prediction. However, constructing graphs directly from raw data may lead to inaccuracies in the graph structure.} To address this, several methods have been developed based on the temporal-first approach. In this approach, temporal information is first captured to extract meaningful temporal features, which are then used to construct graphs for more effective modeling of spatial relationships \citep{wang2023hierarchical_18,wei2024state_53,wang2023multivariate_59}. Zhang et al. \citep{zhang2020adaptive_1} introduced TCN to capture temporal information, which was then combined with GCN for spatial information extraction. They further introduced a gating mechanism to control the feature flow for better capturing temporal information. Li et al. \citep{li2021hierarchical_5} utilized BiLSTM to learn temporal features, followed by graph construction to model spatial relationships. By fixing the graphs, three GIN layers were adopted for capturing the spatial information. Wang et al. \citep{wang2023comprehensive_19} learned temporal features with GRU, which were used to construct graphs through a dynamic graph learning module. The spatial information in these graphs was then captured by GCN. Furthermore, some works stacked multiple temporal encoders and graph models to comprehensively capture temporal and spatial information. Lv et al. \citep{lv2023new_27} stacked multiple layers of TCN and GCN to iteratively capture the temporal and spatial information. Similarly, Cheng et al. \citep{cheng2024research_57} designed multiple layers of LSTM and GCN to capture the information for RUL prediction. Additionally, instead of using temporal features extracted by temporal encoders for graph construction, Chen \citep{chen2023convolution_24} initialized random node embeddings to learn graphs, while the temporal features extracted by CNN served as node features of the graphs. Then, the spatial information in the graphs was captured by a GAT layer. 

\textcolor{black}{Sequential-based approaches provide a straightforward framework that is easy to implement and understand. They also offer flexibility in model design, as spatial and temporal components can be developed, optimized, and updated independently. However, processing spatial and temporal information separately may result in missed interactions between them, potentially overlooking complex interdependencies. Additionally, the order of processing—whether spatial first or temporal first—can introduce biases and limit the model’s ability to capture comprehensive patterns.}

\subsubsection{Parallel}
\textcolor{black}{Parallel-based approaches can address the limitations of the sequential method by capturing spatial and temporal information simultaneously through two parallel branches. These approaches utilize dual processing streams: one for extracting spatial features using graph models and the other for capturing temporal features with temporal encoders.} The extracted spatial and temporal features are then concatenated for final representation learning, as illustrated in Fig. \ref{fig:graph_info} (b).
For example, Zhang et al. \citep{zhang2021adaptive_4} utilized two branches with GCN and TCN to capture spatial and temporal information respectively. Meanwhile, a gating mechanism was introduced for the temporal branch to capture temporal information. Wei et al. \citep{wei2023remaining_26} proposed a self-attention mechanism to generate two parallel temporal-correlated and feature-correlated graphs to represent temporal and spatial information. Both graphs are captured by GCN and then concatenated for final feature representation. Gao et al. \citep{gao2024nonlinear_52} learned spatial features with a GAT and designed masked temporal multi-head attention to capture temporal information. Then, a designed cross multi-head attention mechanism is used to fuse the spatial and temporal features for RUL prediction.

To effectively leverage the benefits of sequential and parallel schemes for capturing spatial and temporal information, some researchers proposed to combine these approaches. Wang et al. \citep{wang2021spatio_3} combined the sequential and parallel schemes by designing two branches. In one branch, GCN and CNN are used for spatial and temporal information respectively, while another branch captures temporal information using TCN. Song et al. \citep{song2024remaining_45} proposed a framework combining both schemes with two routes. Specifically, they designed a spatial route to capture spatial information with GCN and temporal information with LSTM. Simultaneously, a time route was designed to leverage multiple layers of LSTM to capture temporal information.

\textcolor{black}{Compared to sequential-based methods, parallel can simultaneously capture spatial and temporal information, allowing the model to learn more complex and interdependent features. However, running multiple models in parallel increases computational and memory demands. Most importantly, how to effectively combine features from separate branches can be difficult, potentially leading to suboptimal fusion and reduced performance.}

\subsubsection{Integrated}
\textcolor{black}{To more effectively integrate spatial and information, researchers propose integrated approaches, where} temporal and spatial information are captured in a more comprehensive and unified manner within a single module. This approach does not treat temporal and spatial information as separate phases but rather integrates them into a cohesive process to capture spatial and temporal information simultaneously. 

Existing solutions can be divided into two categories. In the first category, researchers proposed combining graph models with temporal encoders to generate temporal graph models. Yang et al. \citep{yang2022bearing_8} designed a temporal GCN, integrating temporal information into the GCN to achieve information extraction in both temporal and spatial domains. Leveraging the property of GRU to process hidden states from previous timestamps along with current states, the authors introduced spatial features learned by GCN as current states and features from previous timestamps as hidden states, enabling GRU to simultaneously capture spatial and temporal information. Kong et al. \citep{kong2024spatio_41} developed a similar method by incorporating node features from past timestamps into the message propagation of MPNN, introducing temporal information into the process of capturing spatial information. This approach allows for the simultaneous capture of spatial and temporal dependencies. Wang et al. \citep{wang2022gated_10} extended the vanilla GCN by integrating GRU, enabling graph models to simultaneously capture spatial and temporal information by replacing the vanilla linear connections in GRU with GCN connections.

In the second category, some researchers proposed capturing spatial and temporal information within a single model. Wang et al. \citep{wang2024dvgtformer_54} discovered that the node feature matrix contains two dimensions, one for temporal information and the other for spatial information. Leveraging this property, they designed a DVGTformer to simultaneously capture spatial and temporal information. Specifically, the model first captures temporal information. By conducting a transpose operation on the temporal features, the same model then captures spatial information. Wang et al. \citep{wang2024fully_60} designed a fully-connected graph to model the correlations of all nodes in sequential graphs, effectively representing spatial and temporal information within a single graph. This graph information is then captured by a specially designed fully-connected graph convolution block for RUL prediction. Zhang et al. \citep{zhang2023spatial_28} proposed an adaptive synchronous convolutional network to capture the heterogeneity among sensors at different times, thereby capturing spatial and temporal information simultaneously.

\textcolor{black}{Compared to sequential and parallel approaches, the integrated approach can jointly model spatial and temporal dependencies, leading to richer and more interconnected feature representations. However, the integrated approach faces scalability issues, especially with very large graphs or long time series. Addressing this limitation is crucial to make this approach practical for real-world applications.}

\subsection{Graph ReadOut}
\begin{figure}[b]
    \centering
    \includegraphics[width = .6\linewidth]{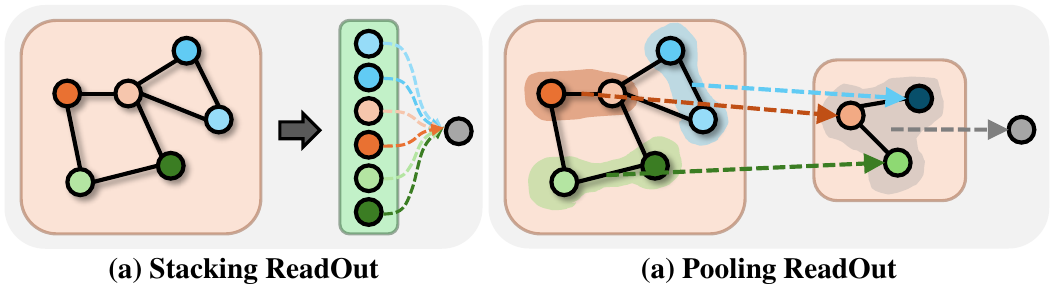}
    \caption{(a) Stacking readout, where the features from all nodes are simply stacked to learn graph representations; (b) Pooling readout, where node features are hierarhically learned by pooling operations.}
    \label{fig:readout}
\end{figure}

With graph models to learn node features by capturing the spatial-temporal information, the learned node-level features should be aggregated to produce graph-level features for RUL prediction. Two common readout methods, stacking readout and pooling readout, are employed by existing methods. In stacking readout, all node features are concatenated to learn graph-level features, while in pooling readout, nodes are selectively aggregated using techniques such as SAGPool and DiffPool.

\subsubsection{Stacking ReadOut}

In stacking readout, the features of all nodes are concatenated to form graph-level features. Given node features \( \mathbf{z}_n \) for \( n = 1, 2, \ldots, N \), the stacking readout can be represented as:
\begin{equation}
\mathbf{h}_\mathcal{G} = \parallel_{n=1}^N\mathbf{z}_n
\end{equation}
Here, \(\mathbf{h}_\mathcal{G}\) is the graph-level representation, and \(\parallel\) denotes the operation of concatenating node features. This method captures all node information but can be computationally expensive for large graphs.

Several studies focused on learning graph-level features by stacking node features in the last layer \citep{wei2023bearing_17,cao2023picture_20,zhu2023rgcnu_21,wei2023remaining_26,liu2023condition_31,kong2024spatio_41,ma2024transformer_49,wei2024state_53}. For instance, Chen et al. \citep{chen2023convolution_24} stacked all nodes' information, combining node features and generated node embedding vectors, to derive graph-level features. Xing et al. \citep{xing2023stcgcn_29} utilized two branches with distinct graph structures, stacking node features within each graph and then concatenating the graph-level features from both branches for final predictions. Wei et al. \citep{wei2024remaining_44} incorporate temporal features extracted by a TCN with stacked node features to enhance the comprehensiveness of graph-level representations for RUL prediction. Wang et al. \citep{wang2023sensor_39,wang2024sea++} similarly stacked information from all nodes and apply a fully connected layer to learn robust graph-level features, which were used not only for RUL prediction but also for domain adaptation to mitigate domain discrepancies effectively.

Instead of restricting node stacking to the final layer alone, some researchers have proposed combining node features across multiple layers for enhanced feature learning. Zhang et al. \citep{zhang2020adaptive_1} proposed stacking node features across multiple layers to facilitate comprehensive feature extraction. In their subsequent work \citep{zhang2021adaptive_4}, they further integrate node features from different layers by summing them up, enabling the learning of final representations while reducing feature dimensionality. Similarly, Wang et al. \citep{wang2023comprehensive_19} employed a summation operation across multiple layers to aggregate node features. They further introduced learnable parameters that dynamically adjust the importance of node features at each layer, enhancing the model's ability to leverage informative features for RUL prediction.

Some works have been proposed to learn graph-level features from sequential graphs with stacking readout. Following the approaches that consider features in different learning layers \citep{zhang2021adaptive_4,wang2023comprehensive_19}, Wang et al. \citep{wang2022gated_10} fused the node features of sequential graphs in multiple layers for RUL prediction. However, directly stacking the features from different layers may result in an excessively large feature dimension. To address this, Wang et al. \citep{wang2023local_15} conducted stacking readout in the last layer, stacking node features in the sequential graphs and mapping them with a mapping layer to learn graph-level features for RUL prediction. To effectively leverage the information in sequential graphs, He et al. \citep{he2023systematic_16} proposed a weighted average method to learn features for all sequential graphs, with the averaged node features then aggregated for RUL prediction. Kong et al. \citep{kong2022spatio_9} found that stacking node features in sequential graphs can still introduce extra computation costs, so they adopted the graph in the final timestamp, whose node information is stacked for final RUL prediction.

It is noted that stacking the node features from a graph or sequential graphs inevitably increases computation costs, so a few researchers have addressed this problem by extracting low-dimension features from each node \citep{li2021remaining_6}. For example, before stacking the node features, Gupta et al. \citep{gupta2020handling_2} leveraged a max operation to extract the maximum values of each node, thereby reducing redundant information within node features. Yang et al. \citep{yang2022bearing_8,yang2023bearing_22} adopted a sum operation to fuse the graphs in different learning layers, and then the features of all nodes were summed up to learn the graph-level features, which reduced computation costs. Similarly, Wen et al. \citep{wen2024temporal_40} employed a mean pooling operation to learn graph-level features by averaging all nodes' information.

\textcolor{black}{The stacking readout has several advantages. First, it preserves all node-level information in the final graph representation, which is beneficial when each node carries significant importance. Second, it is straightforward to implement. However, stacking faces scalability issues. Specifically, it can lead to very high-dimensional feature vectors, particularly in large graphs, making it computationally expensive and challenging to manage. Additionally, hierarchical structures cannot be effectively captured through simple stacking, which can limit the performance of GNN-based RUL prediction. Although some works have addressed computational complexity using sum or mean operations \citep{gupta2020handling_2,wen2024temporal_40}, these methods still struggle to capture hierarchical connections effectively.}

\subsubsection{Pooling ReadOut}
To address these limitations, pooling readout methods have been introduced to aggregate node features hierarchically, aiming to reduce computational complexity while capturing the hierarchical relationships. Two notable pooling methods, SAGPool and DiffPool, have been widely used.

\paragraph{SAGPool (Self-Attention Graph Pooling)}

SAGPool, introduced by Lee et al. \citep{lee2019self}, leverages self-attention mechanisms to determine the importance of nodes and perform pooling based on these attention scores. The steps involved in SAGPool are:

\textbf{Self-Attention:} Self-attention scores are computed for each node:
\begin{equation}
   \mathbf{s} = \text{GNN}_{\text{att}}(\mathbf{Z}, \mathbf{A})
\end{equation}
where \(\mathbf{Z} = [\mathbf{z}_1, \mathbf{z}_2, \ldots, \mathbf{z}_N]^\top\in\mathbb{R}^{N\times{f}}\) and \(\text{GNN}_{\text{att}}\) is a graph neural network layer used to compute attention scores \(\mathbf{s}\).

\textbf{Node Selection:} The neighbors with the $N'$ highest attention scores for each node are then selected for the pooling operation:
\begin{equation}
    \begin{aligned}
       \mathbf{Z}' &= \mathbf{Z}_{\text{top-k}}\\
        \mathbf{A}' &= \mathbf{A}_{\text{top-k}}
    \end{aligned}
\end{equation}
where \(\mathbf{Z}'\in\mathbb{R}^{N'\times{f}}\) and \(\mathbf{A}'\in\mathbb{R}^{N'\times{N'}}\) are node features and the adjacency matrix of the subgraph formed by the selected nodes, respectively.

\textbf{Pooled Graph Representation:}  The selected node features are then used to form the new pooled graph representation. This step involves applying another graph neural network layer to the selected nodes:
\begin{equation}
   \mathbf{h}_\mathcal{G} = \text{GNN}_{\text{pool}}(\mathbf{Z}', \mathbf{A}')
\end{equation}

\paragraph{DiffPool (Differentiable Pooling)}

DiffPool, introduced by Ying et al. \citep{ying2018hierarchical}, constructs a hierarchy of nodes by learning a differentiable soft clustering assignment matrix, which is used to pool node features hierarchically. The steps involved in DiffPool are:

\textbf{Assignments Matrix:} An assignment matrix is computed with noded features and their adjacency matrix.
\begin{equation}
   \mathbf{S} = \text{GNN}_{\text{cluster}}(\mathbf{Z}, \mathbf{A})
\end{equation}
   where \(\mathbf{S} \in \mathbb{R}^{N \times N'}\) is the soft assignment matrix.

\textbf{Cluster:} Nodes are clustered by computing hierarchical node features and adjacency matrix:
\begin{equation}
    \begin{aligned}
       \mathbf{Z}' &= \mathbf{S}^\top \mathbf{Z}\\
        \mathbf{A}' &= \mathbf{S}^\top \mathbf{A} \mathbf{S}
    \end{aligned}
\end{equation}
where \(\mathbf{Z}'\in\mathbb{R}^{N'\times{f}}\) and \(\mathbf{A}'\in\mathbb{R}^{N'\times{N'}}\) are node features and the adjacency matrix, respectively, after clustering with the assignment matrix.

\textbf{Pooled Graph Representation:} The clustered node features are then used for pooled graph representation with another graph model:
\begin{equation}
   \mathbf{h}_\mathcal{G} = \text{GNN}_{\text{pool}}(\mathbf{Z}', \mathbf{A}')
\end{equation}

Currently, several researchers have tried to adopt pooling readout methods to select the most relevant nodes to produce meaningful graph-level embeddings. Wang et al. \citep{wang2023hierarchical_18} introduced multiple layers of SAGPool-based GCN to iteratively learn graph-level features. As the layers deepen, graph-level representations are refined for final predictions. Similarly, Cheng et al. \citep{cheng2024research_57} stacked multiple layers of SAGPool-based GCN and LSTM for RUL prediction, where hierarchical spatial and temporal information is extracted in each layer. Liang et al. \citep{liang2023remaining_23} introduced residual connections to different graph pooling layers, effectively capturing global information from small-scale graph-structured data and fusing characteristics of low-order neighbor nodes. Liu et al. \citep{liu2023mmoe_32} utilized EdgePool \citep{diehl2019edge}, a graph pooling method that filters edges based on edge weights, to reduce the number of nodes, summarize information, and accelerate computation.

Instead of directly applying existing graph pooling methods, some works improved these methods by considering the characteristics of RUL prediction. Li et al. \citep{li2021hierarchical_5} found that the vanilla SAGPool struggles to learn proper attention for each node, so they regularized the pooling operation with a self-attention mechanism. This mechanism computes prior self-attention scores and then calculates the KL-divergence error between the prior scores and the learned attention scores as a regularization term. This regular term helps the model focus more on the nodes with larger initial scores during the learning process, thereby improving model stability. Wang et al. \citep{wang2023multivariate_59} noted that vanilla DiffPool fails to consider sensor correlations when generating the assignment matrix. To address this, they designed a novel assignment matrix that incorporates both sensor correlations and sensor features, and then learned hierarchical spatial information using DiffPool.

Additionally, some researchers have designed novel pooling methods to learn hierarchical information within graphs without relying on existing graph pooling methods. Zhang et al. \citep{zhang2023temporal_30} introduced a graph multiset pooling mechanism, including a graph pooling layer and a self-attention layer to capture hierarchical structural information. The graph pooling layer divides nodes into multiple sets based on their similarity, while the self-attention layer learns high-level relationships among these sets. Wang et al. \citep{wang2024fully_60} designed a fully-connected graph to represent spatial and temporal information within a single graph. To learn hierarchical information from this graph, they proposed a temporal pooling operation to extract high-level features from temporally close neighbors.

\textcolor{black}{Pooling-based readout approaches can effectively reduce the size of the feature vector, making the graph representation more manageable and computationally efficient. Additionally, they are capable of capturing hierarchical structure information, leading to better representation learning. However, this reduction process inevitably results in the loss of node information, potentially overlooking important node-level details. This can negatively impact the performance of GNN-based RUL prediction, particularly in cases where each sensor plays a crucial role. Therefore, in real-world applications, the choice between stacking and pooling-based readout should be guided by the specific properties of the data.}
\section{Comparative Analysis}
\begin{figure}[!hbpt]
    \centering
    \includegraphics[width = 0.35\textwidth]{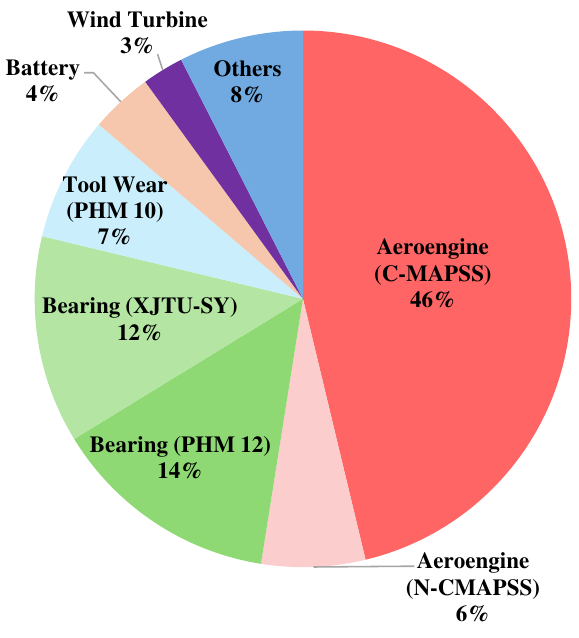}
    \caption{Dataset usage in existing works}
    \label{fig:data_usage}
\end{figure}
Although GNN-based RUL prediction has been widely studied in recent years, fair comparisons among these methods are challenging due to varying experimental settings. Key differences include:

\begin{itemize}
    \item \textbf{Different time length:} Existing works have utilized different time lengths of samples in their experiments. For example, Wang et al. \citep{wang2024fully_60} used a time length of 60 for experiments with C-MAPSS, while Kong et al. \citep{kong2022spatio_9} used a time length of 15.
    \item \textbf{Different number of channels:} Huang et al. \citep{huang2024spatio_42} utilized all 21 channels of C-MAPSS, whereas Zhang et al. \citep{zhang2020adaptive_1} used 18 channels, and Wang et al. \citep{wang2023hierarchical_18} used 14 channels. The varying number of channels contains different amounts of information, making graphs constructed with different number of nodes. Thus, it is difficult to achieve fair comparisons of GNN-based RUL prediction.
    \item \textbf{Different dataset sizes:} Zeng et al. \citep{zeng2022remaining_11} used the original length of 32,768 in XJTU-SY for their experiments, while Wei et al. \citep{wei2023bearing_17} used 10\% of each sample with sub-sampling to reduce computation costs. Deep learning models rely on extensive data for training, and different dataset sizes make different training.
    \item \textbf{Different methods for constructing sequential graphs to capture temporal information:} Li et al. \citep{li2021hierarchical_5} used multiple samples to construct sequential graphs, whereas Wang et al. \citep{wang2023local_15} constructed sequential graphs with a single sample using a multi-patch segmentation method to capture temporal dependencies within the sample.
\end{itemize}
These differences in experimental settings make fair comparisons difficult. Additionally, the lack of publicly available code for most existing works makes it difficult to implement these methods under the same settings, further complicating fair comparisons. This problem makes it hard for newcomers to compare their research with existing works, restricting the development of new research in this area. Therefore, establishing a benchmark for ensuring fair comparisons is crucial to advance the development of GNN-based methods.

To evaluate these methods for benchmarking, appropriate datasets need to be chosen. Currently, GNN-based RUL prediction has been explored in various applications. As shown in Fig. \ref{fig:data_usage}, existing works have utilized datasets for aeroengine \citep{wang2024fully_60,kong2022spatio_9,huang2024spatio_42,wang2023hierarchical_18}, bearing \citep{li2021remaining_6,yang2022bearing_8,zeng2022remaining_11,Yang2023path_14}, tool wear \citep{wang2022gated_10,liang2023remaining_23}, battery \citep{wei2023prediction_25,xu2023hybrid}, and more. From this figure, we observe that aeroengine and bearing datasets have been widely explored. Thus, we conducted a comprehensive analysis on four datasets from aeroengine and bearing applications, including C-MAPSS \citep{saxena2008damage}, N-CMAPSS \citep{arias2021aircraft}, PHM 12 \citep{nectoux2012pronostia}, and XJTU-SY \citep{wang2018hybrid}.

\subsection{Dataset Description}
\begin{table*}[htbp]
  \centering
  \caption{Statistics of C-MAPSS and N-CMAPSS}
    \begin{tabular}{c|cccc|ccc}
    \toprule
    \toprule
          & \multicolumn{4}{c|}{C-MAPSS}  & \multicolumn{3}{c}{N-CMAPSS} \\
          & FD001 & FD002 & FD003 & FD004 & Unit 11 & Unit 14 & Unit 15 \\
    \midrule
    Number of Training & 15,731  & 41,019  & 19,820  & 49,048  & \multicolumn{3}{c}{52,341 } \\
    \midrule
    Number of Testing & 100   & 259   & 100   & 248   & 6,586  & 1,519  & 4,285  \\
    \midrule
    Max RUL & \multicolumn{4}{c|}{125 }     & \multicolumn{3}{c}{88 } \\
    \midrule
    Time Length & \multicolumn{4}{c|}{50 }      & \multicolumn{3}{c}{50 } \\
    \midrule
    Number of Channels & \multicolumn{4}{c|}{14 }      & \multicolumn{3}{c}{20 } \\
    \bottomrule
    \bottomrule
    \end{tabular}%
  \label{tab:aero}%
\end{table*}%

\begin{table*}[htbp]
  \centering
  \caption{Statistics of PHM 12}
    \begin{tabular}{c|ccccccccccc}
    \toprule
    \toprule
          & \multicolumn{5}{c|}{Condition 1}      & \multicolumn{5}{c|}{Condition 2}      & Condition 3 \\
    Testing Bearing ID & 3     & 4     & 5     & 6     & \multicolumn{1}{c|}{7} & 3     & 4     & 5     & 6     & \multicolumn{1}{c|}{7} & 3 \\
    \midrule
    Number of Training & \multicolumn{5}{c|}{3,674 }           & \multicolumn{5}{c|}{1,708 }           & 2,152  \\
    \midrule
    Number of Testing & 1,802  & 1,139  & 2,302  & 2,302  & \multicolumn{1}{c|}{1,502 } & 1,202  & 612   & 2,002  & 572   & \multicolumn{1}{c|}{172 } & 352  \\
    \midrule
    Max RUL & 2,375  & 1,173  & 2,463  & 2,448  & \multicolumn{1}{c|}{2,259 } & 1,955  & 751   & 2,311  & 701   & \multicolumn{1}{c|}{230 } & 434  \\
    \midrule
    Time Length & \multicolumn{11}{c}{2,560 } \\
    \midrule
    Number of Channels & \multicolumn{11}{c}{1 } \\
    \bottomrule
    \bottomrule
    \end{tabular}%
  \label{tab:PHM12}%
\end{table*}%

\begin{table}[htbp]
  \centering
  \caption{Statistics of XJTU-SY}
    \begin{tabular}{c|ccccc}
    \toprule
    \toprule
          & \multicolumn{5}{c}{Condition 1} \\
    Testing Bearing ID & 1     & 2     & 3     & 4     & 5 \\
    \midrule
    Number of Training & 493   & 455   & 458   & 494   & 564  \\
    \midrule
    Number of Testing & 123   & 161   & 158   & 122   & 52  \\
    \midrule
    Max RUL & 123   & 161   & 158   & 122   & 52  \\
    \midrule
    \midrule
          & \multicolumn{5}{c}{Condition 2} \\
    Testing Bearing Number & 1     & 2     & 3     & 4     & 5 \\
    \midrule
    Number of Training & 1,075  & 1,405  & 1,033  & 1,524  & 1,227  \\
    \midrule
    Number of Testing & 491   & 161   & 533   & 42    & 339  \\
    \midrule
    Max RUL & 491   & 161   & 533   & 42    & 339  \\
    \midrule
    \midrule
          & \multicolumn{5}{c}{Condition 3} \\
    Testing Bearing Number & 1     & 2     & 3     & 4     & 5 \\
    \midrule
    Number of Training & 4,496  & 4,538  & 6,663  & 5,519  & 6,920  \\
    \midrule
    Number of Testing & 2,538  & 2,496  & 371   & 1,515  & 114  \\
    \midrule
    Max RUL & 2,538  & 2,496  & 371   & 1,515  & 114  \\
    \midrule
    Time Length & \multicolumn{5}{c}{32,768 } \\
    \midrule
    Number of Channels & \multicolumn{5}{c}{1 } \\
    \bottomrule
    \bottomrule
    \end{tabular}%
  \label{tab:XJTU}%
\end{table}%

\paragraph{C-MAPSS} 
The C-MAPSS dataset describes the degradation of aircraft engines. To comprehensively monitor the engine's status, 21 sensors were deployed at different positions to measure various parameters such as temperature, pressure, and fan speed. Following previous studies \citep{chen2020machine, xu2021kdnet}, we remove sensors with constant values, specifically those with indices 1, 5, 6, 10, 16, 18, and 19. This dataset includes four subdatasets from FD001 to FD004, each collected under different operating conditions and fault modes. As these datasets record the entire life cycle of the engines, it is necessary to extract samples from the full duration of the life cycle. For engines with a life cycle of $\mathcal{T}$, a time window of fixed length $L$ slides along the signals by $S$ steps during each sampling process. Consequently, each sample obtained has a length of $L$ with its RUL calculated as $y^i = \mathcal{T}-L-i*S$, where $i$ represents the $i$-th sampling instance. We further adopt a piece-wise linear RUL model to normalize the sample labels due to its effectiveness in representing the RUL of a machine \citep{heimes2008recurrent, jayasinghe2019temporal, zhang2016multiobjective}. Specifically, we predefined a maximum RUL value: if the RUL of a sample exceeds this value, it is set to the maximum value. In our setup, $S$ is set to 1, and the maximum RUL is set to 125 \citep{li2018remaining, wang2023local_15}. The data are then normalized using max-min normalization to restrict the values within the range [0, 1]. Notably, the engines in C-MAPSS have been split as the training and testing datasets. After preprocessing, the statistics are shown as TABLE \ref{tab:aero}.

\paragraph{N-CMAPSS}
The N-CMAPSS dataset also records the run-to-failure trajectories of turbofan engines. Compared with C-MAPSS, which restricted flight conditions to a standard cruise phase, this dataset simulated complete flights, including climb, cruise, and descent flight conditions. Additionally, it increased the fidelity of degradation modeling. These improvements make N-CMAPSS better represent the complex factors present in real systems.
N-CMAPSS includes eight subsets, each consisting of data from five parts: scenario descriptors, sensor measurements, virtual sensors, model health parameters, and auxiliary data. Following the settings in previous works \citep{mo2022multi}, we adopted DS02 and selected 20 channels for training. These channels include all four scenario descriptors, 14 sensor measurements, and two virtual sensors (T40 and P30), as they are important for predicting RULs.
The DS02 dataset consists of nine units recording run-to-failure trajectories, with six units (units 2, 5, 10, 16, 18, and 20) used for training and three units (units 11, 14, and 15) used for testing. As this dataset records the entire life cycles of engines, we applied similar processing as in C-MAPSS. Notably, N-CMAPSS contains many more timestamps compared to C-MAPSS. To reduce training and evaluation costs, we subsampled the dataset with an interval of 100 before processing. The maximum RUL is set to 88 to normalize the RUL labels, and max-min normalization is used to normalize the data. After preprocessing, the statistics are shown in TABLE \ref{tab:aero}.

\paragraph{PHM 12}
The PHM 12 dataset describes the degradation process of rolling bearings accelerated by the PRONOSTIA platform developed by the FEMTO-ST Institute. The raw data were collected under three operating conditions, each with different speed, radial load, and applied torque. The first and second conditions include seven bearings each, while the third condition has three bearings. Each bearing has three channels recording horizontal vibration signals, vertical vibration signals, and temperature data. Following previous studies \citep{wei2023bearing_17, zeng2022remaining_11, wei2023remaining_26}, only the horizontal vibration channel is retained, as the other channels have little influence on RUL prediction.
The data sampling frequency of the vibration channel was 25.6 kHz, and the sampling time was 0.1 seconds, resulting in each data segment having a time length of 2,560 data points. Max-min normalization is used to normalize the data. To normalize the labels, we account for the varying maximal RULs of different bearings by normalizing each bearing by its own maximal RUL.
For training purposes, we adopted the predefined train-test splits. Specifically, the first two bearings in each condition were used for training, and the remaining bearings were used for testing. The maximal RULs of the bearings and their statistics can be found in TABLE \ref{tab:PHM12}.


\paragraph{XJTU-SY}
This dataset was collected by Xi’an Jiaotong University and the Changxing Sumyoung Technology Company, gathered under three operating conditions, each with different rotating speeds and radial forces. Each condition includes five bearings, where vibration signals in two directions were collected using two identical accelerometers positioned at ninety degrees. The vibration signals were recorded every minute with a sampling frequency of 25.6 kHz and a duration of 1.28 seconds, resulting in each data segment having a length of 32,768 data points.
Similar to PHM 12, only the horizontal vibration channel is retained for model training. Max-min normalization is used to normalize the data, and the labels of each bearing's data are normalized by its own maximal RUL. For training and evaluation, we adopted a strategy where one bearing is used for testing and the remaining bearings for training in each operating condition. The statistics of the training and testing datasets are shown in TABLE \ref{tab:XJTU}.

\subsection{Evaluation Metrics}
We adopted multiple metrics to comprehensively evaluate the performance of existing GNN-based methods. Here, Root Mean Square Error (RMSE) (Eq. (\ref{eq:rmse})) and Mean Absolute Error (MAE) (Eq. (\ref{eq:mae})) have been adopted, as they can represent the differences between predicted and real RULs. Notably, RMSE and MAE treat early and late predictions equally; however, in prognostics, late predictions of RUL can be more harmful. To address this, we incorporated score functions that impose heavier penalties for late predictions, thus reflecting the urgency and importance of accurate late-stage prognostics. Currently, two score functions, $S_{\text{v1}}$ \citep{jin2022position,wang2023local_15,wang2023hierarchical_18} and $S_{\text{v2}}$ \citep{yang2022bearing_8,nectoux2012pronostia,cao2023picture_20}, have been widely used, as shown in Eq. (\ref{eq:scorev1}) and (\ref{eq:scorev2}). Particularly, $S_{\text{v1}}$ has primarily been used in aeroengine tasks, while $S_{\text{v2}}$ has mainly been applied to bearing tasks. Thus, $S_{\text{v1}}$ and $S_{\text{v2}}$ have been used respectively for evaluation in aeroengine and bearing tasks.
Notably, for $S_{\text{v1}}$, many existing works \citep{wang2023local_15,wang2023hierarchical_18} have used the sum of the score of each sample, so both $S_{\text{v1}}$ and $S_{\text{v1}, \text{sum}}$ have been reported in the following experiments.

\textbf{RMSE:} 
\begin{equation}
\text{RMSE} = \sqrt{\frac{1}{B}\sum_{i=1}^B (y_i-\hat{y}_i)^2},
\label{eq:rmse}
\end{equation}

\textbf{MAE:} 
\begin{equation}
\text{MAE} = \frac{1}{B}\sum_{i=1}^B |y_i-\hat{y}_i|,
\label{eq:mae}
\end{equation}

\textbf{$S_{\text{v1}}$:} 
\begin{equation}
\label{eq:scorev1}
\begin{aligned}
S_{\text{v1}, i} =
&\begin{cases}
\exp\left(-\frac{\hat{y}_i - y_i}{13}\right) - 1, & \text{if } \hat{y}_i \leq y_i \\
\exp\left(\frac{\hat{y}_i - y_i}{10}\right) - 1, & \text{if } \hat{y}_i > y_i
\end{cases} \\
&S_{\text{v1}, \text{sum}} = \sum_{i=1}^B S_{\text{v1}, i} \\
&S_{\text{v1}} = \frac{\sum_{i=1}^B S_{\text{v1}, i}}{B}
\end{aligned}
\end{equation}

\textbf{$S_{\text{v2}}$:} 
\begin{equation}
\label{eq:scorev2}
\begin{aligned}
S_{\text{v2}, i} &=
\begin{cases}
\exp\left(-\ln(0.5) \cdot \frac{E_{ri}}{5}\right), & \text{if } E_{ri} \leq 0 \\
\exp\left(\ln(0.5) \cdot \frac{E_{ri}}{20}\right), & \text{if } E_{ri} > 0
\end{cases} \\
\quad &E_{ri} = \frac{y_i - \hat{y}_i}{y_i} \times 100\\
&S_{\text{v2}} = \frac{\sum_{i=1}^B S_{\text{v2}, i}}{B}
\end{aligned}
\end{equation}
Here, $\hat{y_i}$ and $y_i$ represent predicted and real labels for sample $i$, respectively, and $B$ denotes the number of samples used for evaluations. For RMSE, MAE, and $S_{\text{v1}}$, lower values indicate better performance, while for $S_{\text{v2}}$, higher values are preferable.

\subsection{Comparisons of GNN-based Methods for RUL prediction}
\begin{table*}[htbp]
  \centering
  \caption{Categorization of benchmarking methods}
  \begin{adjustbox}{width = 1.0\textwidth,center}
    \begin{tabular}{cl|cc|ccc|cccc|ccc|cc}
    \toprule
    \toprule
    \multirow{3}[6]{*}{Years} & \multicolumn{1}{c|}{\multirow{3}[6]{*}{Methods}} & \multicolumn{5}{c|}{Graph Construction} & \multicolumn{4}{c|}{Graph Models} & \multicolumn{3}{c|}{\multirow{2}[4]{*}{Graph Information}} & \multicolumn{2}{c}{\multirow{2}[4]{*}{Graph ReadOut}} \\
\cmidrule{3-11}          &       & \multicolumn{2}{c|}{Node Definition} & \multicolumn{3}{c|}{Edge Definition} & \multicolumn{2}{c}{Spectral Conv.} & \multicolumn{2}{c|}{Spatial Conv.} & \multicolumn{3}{c|}{} & \multicolumn{2}{c}{} \\
\cmidrule{3-16}          &       & Multi & Single & Metric & Att. & Prior & ChebNet & GCN   & MPNN  & GAT   & Seq.  & Parallel & Integ. & Stack. & Pool. \\
    \midrule
    2020  & ASTGCNN \citep{zhang2020adaptive_1} & $\checkmark$     &       & $\checkmark$     &       &       & $\checkmark$     &       &       &       & $\checkmark$     &       &       & $\checkmark$     &  \\
    \midrule
    2020  & GRU-CM \citep{gupta2020handling_2} & $\checkmark$     &       &       & $\checkmark$     &       &       &       & $\checkmark$     &       & $\checkmark$     &       &       & $\checkmark$     &  \\
    \midrule
    2021  & HAGCN \citep{li2021hierarchical_5} & $\checkmark$     &       & $\checkmark$     &       &       &       &       & $\checkmark$     &       & $\checkmark$     &       &       &       & $\checkmark$ \\
    \midrule
    2021  & ST-Conv \citep{wang2021spatio_3} & $\checkmark$     &       & $\checkmark$     &       &       &       &       & $\checkmark$     &       & $\checkmark$     & $\checkmark$     &       & $\checkmark$     &  \\
    \midrule
    2022  & STFA \citep{kong2022spatio_9}  & $\checkmark$     &       &       &       & $\checkmark$     &       &       &       & $\checkmark$     & $\checkmark$     &       &       & $\checkmark$     &  \\
    \midrule
    2023  & RGCNU \citep{zhu2023rgcnu_21} & $\checkmark$     &       & $\checkmark$     &       &       &       & $\checkmark$     &       &       & $\checkmark$     &       &       & $\checkmark$     &  \\
    \midrule
    2023  & STAGNN \citep{huang2024spatio_42} & $\checkmark$     &       & $\checkmark$     &       &       &       & $\checkmark$     &       &       & $\checkmark$     &       &       & $\checkmark$     &  \\
    \midrule
    2023  & HierCorrPool \citep{wang2023hierarchical_18} & $\checkmark$     &       & $\checkmark$     &       &       &       &       & $\checkmark$     &       & $\checkmark$     &       &       &       & $\checkmark$ \\
    \midrule
    2023  & LOGO \citep{wang2023local_15}  & $\checkmark$     &       & $\checkmark$     &       &       &       &       & $\checkmark$     &       & $\checkmark$     &       &       & $\checkmark$     &  \\
    \midrule
    2023  & DVGTformer \citep{wang2024dvgtformer_54} & $\checkmark$     &       &       & $\checkmark$     &       &       &       & $\checkmark$     &       &       &       & $\checkmark$     & $\checkmark$     &  \\
    \midrule
    2023  & STGNN \citep{liu2023condition_31} & $\checkmark$     &       & $\checkmark$     &       &       & $\checkmark$     &       &       &       & $\checkmark$     &       &       & $\checkmark$     &  \\
    \midrule
    2024  & FC-STGNN \citep{wang2024fully_60} & $\checkmark$     &       & $\checkmark$     &       &       &       &       & $\checkmark$     &       &       &       & $\checkmark$     &       & $\checkmark$ \\
    \midrule
    \midrule
    2021  & ST-GCN \citep{li2021remaining_6} &       & $\checkmark$     & $\checkmark$     &       &       &       &       & $\checkmark$     &       & $\checkmark$     &       &       & $\checkmark$     &  \\
    \midrule
    2022  & SAGCN \citep{wei2023bearing_17} &       & $\checkmark$     & $\checkmark$     &       &       &       & $\checkmark$     &       &       &       &       &       & $\checkmark$     &  \\
    \midrule
    2023  & STNet \citep{zeng2022remaining_11} &       & $\checkmark$     &       & $\checkmark$     &       & $\checkmark$     &       &       &       & $\checkmark$     &       &       &       & $\checkmark$ \\
    \midrule
    2023  & GAT-LSTM \citep{Yang2023path_14} &       & $\checkmark$     & $\checkmark$     &       &       &       &       &       & $\checkmark$     & $\checkmark$     &       &       & $\checkmark$     &  \\
    \midrule
    2023  & STMSGCN \citep{yang2023bearing_22} &       & $\checkmark$     & $\checkmark$     &       &       &       & $\checkmark$     &       &       &       &       &       & $\checkmark$     &  \\
    \midrule
    2023  & AGCN-TF \citep{wei2023remaining_26} &       & $\checkmark$     &       & $\checkmark$     &       &       &       & $\checkmark$     &       &       &       & $\checkmark$     &     &$\checkmark$   \\
    \midrule
    2023  & LOGO-S \citep{wang2023local_15} &       & $\checkmark$     & $\checkmark$     &       &       &       &       & $\checkmark$     &       & $\checkmark$     &       &       & $\checkmark$     &  \\
    \midrule
    2023  & HierCorrPool-S \citep{wang2023hierarchical_18} &       & $\checkmark$     & $\checkmark$     &       &       &       &       & $\checkmark$     &       & $\checkmark$     &       &       &       & $\checkmark$ \\
    \midrule
    2024  & GDAGDL \citep{yang2024dynamic_50} &       & $\checkmark$     & $\checkmark$     &       &       &       &       &       & $\checkmark$     & $\checkmark$     &       &       &       & $\checkmark$ \\
    \bottomrule
    \bottomrule
    \end{tabular}%
    \end{adjustbox}
  \label{tab:categorization}%
\end{table*}%
For comprehensive evaluations, the methods in each category have been chosen, as shown in TABLE \ref{tab:categorization}. Where available, we adopted the original settings or codes of these methods. Otherwise, the hyperparameter optimization has been conducted to ensure optimal implementation. Notably, methods designed for multi-channel datasets are evaluated on aeroengine datasets, while those designed for single-channel datasets are evaluated on bearing datasets. Below, we provide detailed descriptions of each method:

\textbf{ASTGCNN}: Utilizes a gating-based TCN to capture temporal information for feature extraction. Generalized Mahalanobis distance is employed to construct graphs, followed by ChebNet for capturing spatial information.

\textbf{GRU-CM}: Leverages attention mechanisms to propagate relevant features across nodes. Then, all node features are aggregated, and GRU captures temporal information from global readout features.

\textbf{HAGCN}: Employs BiLSTM for feature extraction, constructing sequential graphs using PCCs. GIN captures spatial information, and SAGPool reads graph features from nodes. Notably, HAGCN originally utilized multiple samples to construct sequential graphs. However, processing one sample at a time is typically more practical. To adapt, we employed the graph construction method described in \citep{wang2023local_15}, which involves segmenting multiple patches from each sample to construct sequential graphs.

\textbf{ST-Conv}: Constructs graphs using PCCs from the original sample and leverages two parallel branches to process information: TCN for temporal and MPNN/CNN sequentially for spatial and temporal information.

\textbf{STFA}: Defines sequential graphs using prior-knowledge of sensor connections and utilizes multi-patch segmentation to extract node features for sequential graphs. GAT captures spatial information, followed by attention sequence embedding layers to learn attention weights among sequential graphs and LSTM to capture temporal information.

\textbf{RGCNU}: Constructs graphs using a metric-based approach from raw data. A GCN-based module and an LSTM-based module are employed to learn spatial correlations and temporal dependency respectively, followed by a fusion module to integrate outputs with residual information from raw data.

\textbf{STAGNN}: Constructs graphs based on node correlations, with GCN capturing spatial information. Multi-head spatial attention modules enhance spatial information learning. Then, TCN captures temporal information, with multi-head temporal attention for effective temporal learning.

\textbf{HierCorrPool}: Captures temporal features with 1D CNN, based on which sequential graphs are constructed with dot-product distances. To learn hierarchical spatial information, an assignment matrix is introduced by considering both sensor features and correlations.

\textbf{LOGO}: Designs multi-patch segmentation and employs a mapping function to learn sequential micro-graphs. PCCs and dot-product distances are utilized to compute global and local correlations, respectively, for effectively capturing spatial information.

\textbf{DVGTformer}: Processes spatial and temporal information simultaneously with a designed GTformer, which employs a self-attention mechanism to compute edges among nodes.

\textbf{STGNN}: Utilizes a Gaussian kernel weight function to compute similarities and filters weak connections with top-K. Three-order ChebNet and GRU capture spatial and temporal information respectively.

\textbf{FC-STGNN}: Constructs a fully connected graph to link all nodes across all sequential patches. A pooling-based graph convolution layer is designed to capture spatial and temporal information simultaneously.

\textbf{ST-GCN}: Constructs graphs using PCCs and captures spatial and temporal information with MPNN and TCN respectively. Notably, ST-GCN was designed for bearing RUL prediction with single-channel data, so the channel is extended to multi-channel by deriving statistical features. Furthermore, ST-GCN originally constructed sequential graphs with multiple samples, so multi-patch segmentation is utilized for each sample to obtain patches, each deriving statistical features.

\textbf{SAGCN}: Extends single-channel data to multiple channels by extracting temporal and frequency features from patches segmented from each sample. Graphs are constructed with cosine distances and then processed by a GCN with a self-attention module to enhance feature representation.

\textbf{STNet}: Applies Short-Time Fourier Transform (STFT) to extract frequency features from patches segmented from a sample. With each frequency component as a node, the max and mean values of each node are extracted and concatenated to obtain node attention. Edges are connected for nodes with attentions larger than a threshold. Then, ChebNet captures spatial information, and autoencoder-based readout is designed to learn graph features from nodes.

\textbf{GAT-LSTM}: Extends single-channel data to multi-channel by extracting temporal statistical features. PathGraph is introduced to define graphs, with adjacent timestamps as connections. GATs and LSTM are utilized to capture temporal and spatial information respectively.

\textbf{STMSGCN}: Utilizes FFT features to extract SD features for state changes, whose integrated energy is calculated to obtain SED features as nodes. Graphs are constructed with dot-product distances and processed by GCN. The hidden spatial features from multiple GCN layers are concatenated to capture multi-scale spatial information.

\textbf{AGCN-TF}: Extracts features similar to SAGCN. Temporal and spatial attention adjacency matrices are derived based on the features and then processed by MPNN for temporal and spatial information respectively. A self-attention mechanism is designed to integrate the both information effectively.

\textbf{LOGO-S}: Extension of LOGO for single-channel cases using STFT features as nodes.

\textbf{HierCorrPool-S}: Extension of HierCorrPool for single-channel cases using STFT features as nodes.

\textbf{GDAGDL}: Uses STFT to extract frequency features from patches segmented from each sample as nodes. PCC is utilized to compute node relations, based on which node attentions can be computed. Edges are connected for nodes with attentions larger than a threshold, and GAT is utilized for capturing the information. An autoencoder-based graph readout is designed to learn graph features from nodes.

\begin{table*}[htbp]
  \centering
  \caption{Comparisons of SOTA GNN-based methods in C-MAPSS for aeroengine}
        \begin{adjustbox}{width = 1.\textwidth,center}
    \begin{tabular}{l|cccc|cccc|cccc|cccc}
    \toprule
    \toprule
    \multirow{2}[2]{*}{Models} & \multicolumn{4}{c|}{FD001}    & \multicolumn{4}{c|}{FD002}    & \multicolumn{4}{c|}{FD003}    & \multicolumn{4}{c}{FD004} \\
          & RMSE  & MAE   & $S_{\text{v1}, \text{sum}}$ & $S_{\text{v1}}$ & RMSE  & MAE   & $S_{\text{v1}, \text{sum}}$ & $S_{\text{v1}}$ & RMSE  & MAE   & $S_{\text{v1}, \text{sum}}$ & $S_{\text{v1}}$ & RMSE  & MAE   & $S_{\text{v1}, \text{sum}}$ & $S_{\text{v1}}$ \\
    \midrule
    ASTGCNN & 13.91 & 10.36 & 344   & 3.440 & 13.81 & 10.42 & 924   & 3.568 & 12.01 & 8.85  & 259   & 2.590 & 14.03 & 10.63 & \underline{855} & \underline{3.448} \\
    GRU-CM & 21.47 & 17.73 & 1474  & 14.740 & 18.62 & 14.93 & 3280  & 12.664 & 18.66 & 15.33 & 887   & 8.870 & 19.42 & 15.91 & 1966  & 7.927 \\
    HAGCN & 13.63 & 10.10 & 366   & 3.660 & 13.70 & 10.04 & 820   & 3.166 & 13.27 & 10.02 & 311   & 3.110 & 14.63 & 9.89  & 1360  & 5.484 \\
    ST-Conv & 13.72 & 10.48 & 277   & 2.770 & 14.15 & 10.87 & 968   & 3.737 & 13.27 & 9.98  & 315   & 3.150 & 15.70 & 11.51 & 1215  & 4.899 \\
    STFA  & 13.58 & 9.98  & 279   & 2.790 & 13.53 & 9.95  & 806   & 3.112 & 13.35 & 10.23 & 270   & 2.700 & 15.06 & 10.82 & 1184  & 4.774 \\
    RGCNU & 13.91 & 10.72 & 362   & 3.620 & 13.89 & 10.75 & 920   & 3.552 & 13.31 & 10.35 & 296   & 2.960 & 15.34 & 11.93 & 976   & 3.935 \\
    STAGNN & 13.30 & 10.06 & 279   & 2.790 & 16.60 & 12.30 & 1750  & 6.757 & 13.46 & 10.00 & 324   & 3.240 & 16.94 & 12.33 & 1629  & 6.569 \\
    HierCorrPool & 12.61 & 9.47  & 254   & 2.540 & 13.11 & 9.92  & 750   & 2.896 & 12.13 & 9.25  & 218   & 2.180 & 14.65 & 11.08 & 1033  & 4.165 \\
    LOGO  & \underline{12.39} & 9.23  & \underline{246} & \underline{2.460} & \underline{12.49} & \underline{9.23} & \textbf{660} & \textbf{2.548} & 12.33 & 9.12  & 239   & 2.390 & 13.83 & 9.83  & 917   & 3.698 \\
    DVGTformer & 12.48 & \underline{9.17} & 250   & 2.500 & \textbf{12.24} & \textbf{8.78} & \underline{689} & \underline{2.660} & \textbf{11.15} & \underline{8.16} & \underline{206} & \underline{2.060} & \textbf{13.19} & \underline{9.44} & \textbf{827} & \textbf{3.335} \\
    STGNN & 13.97 & 10.84 & 301   & 3.010 & 13.57 & 10.16 & 856   & 3.305 & 11.90 & 8.92  & 236   & 2.360 & 14.79 & 10.75 & 1050  & 4.234 \\
    FC-STGNN & \textbf{12.36} & \textbf{9.13} & \textbf{214} & \textbf{2.140} & 12.91 & 9.35  & 786   & 3.035 & \underline{11.34} & \textbf{8.08} & \textbf{198} & \textbf{1.980} & \underline{13.28} & \textbf{9.29} & 968   & 3.903 \\
    \bottomrule
    \bottomrule
    \end{tabular}%
    \end{adjustbox}
  \label{tab:sotacmpass}%
\end{table*}%

\begin{table*}[htbp]
  \centering
  \caption{Comparisons of SOTA GNN-based methods in N-CMAPSS for aeroengine}
      \begin{adjustbox}{width = .7\textwidth,center}
    \begin{tabular}{l|cccc|cccc|cccc}
    \toprule
    \toprule
    \multirow{2}[2]{*}{Models} & \multicolumn{4}{c|}{Unit 11}   & \multicolumn{4}{c|}{Unit 14}   & \multicolumn{4}{c}{Unit 15} \\
          & RMSE  & MAE   & $S_{\text{v1}, \text{sum}}$ & $S_{\text{v1}}$ & RMSE  & MAE   & $S_{\text{v1}, \text{sum}}$ & $S_{\text{v1}}$ & RMSE  & MAE   & $S_{\text{v1}, \text{sum}}$ & $S_{\text{v1}}$ \\
    \midrule
    ASTGCNN & \textbf{4.23} & \textbf{3.23} & \textbf{2678} & \textbf{0.407} & 7.96  & 6.46  & 1441  & 0.949 & \underline{3.71} & \underline{2.84} & \underline{1433} & \underline{0.334} \\
    GRU-CM & 16.54 & 14.03 & 28078 & 4.263 & 20.90 & 18.18 & 10260 & 6.754 & 18.50 & 15.84 & 22324 & 5.210 \\
    HAGCN & 5.45  & 3.94  & 3593  & 0.546 & 8.95  & 6.79  & 1693  & 1.115 & 4.56  & 3.38  & 1718  & 0.401 \\
    ST-Conv & \underline{4.76} & 3.70  & 3281  & 0.498 & \underline{6.71} & \underline{5.39} & \underline{1094} & \underline{0.720} & 3.89  & 3.03  & 1482  & 0.346 \\
    RGCNU & 9.61  & 8.36  & 9222  & 1.400 & 11.84 & 10.25 & 2714  & 1.787 & 9.70  & 8.32  & 5524  & 1.289 \\
    STAGNN & 17.88 & 15.50 & 34768 & 5.279 & 21.72 & 19.05 & 11517 & 7.582 & 19.14 & 16.57 & 23478 & 5.479 \\
    HierCorrPool & 4.85  & 4.00  & 3444  & 0.523 & 7.42  & 6.01  & 1252  & 0.824 & 3.83  & 2.99  & 1503  & 0.351 \\
    LOGO  & 7.94  & 6.24  & 7403  & 1.124 & 10.28 & 8.20  & 2519  & 1.658 & 7.82  & 6.03  & 4053  & 0.946 \\
    DVGTformer & 4.86  & 3.79  & 3473  & 0.527 & \textbf{6.39} & \textbf{5.00} & \textbf{996} & \textbf{0.656} & \textbf{3.44} & \textbf{2.43} & \textbf{1201} & \textbf{0.280} \\
    STGNN & 8.72  & 7.26  & 8017  & 1.217 & 11.45 & 9.27  & 2673  & 1.760 & 8.26  & 6.79  & 4392  & 1.025 \\
    FC-STGNN & 5.06  & \underline{3.62} & \underline{3211} & \underline{0.488} & 8.48  & 6.83  & 1586  & 1.044 & 4.96  & 3.56  & 1904  & 0.444 \\
    \bottomrule
    \bottomrule
    \end{tabular}%
    \end{adjustbox}
  \label{tab:sotancmpass}%
\end{table*}%

\begin{table*}[htbp]
  \centering
  \caption{Comparisons of SOTA GNN-based methods in condition 1 of PHM 12 for bearing}
      \begin{adjustbox}{width = 1.0\textwidth,center}
    \begin{tabular}{l|ccc|ccc|ccc|ccc|ccc}
    \toprule
    \toprule
    \multirow{3}[2]{*}{Models} & \multicolumn{15}{c}{Condition 1} \\
          & \multicolumn{3}{c|}{Test Bearing 3} & \multicolumn{3}{c|}{Test Bearing 4} & \multicolumn{3}{c|}{Test Bearing 5} & \multicolumn{3}{c|}{Test Bearing 6} & \multicolumn{3}{c}{Test Bearing 7} \\
          & RMSE  & MAE   & $S_{\text{v2}}$ & RMSE  & MAE   & $S_{\text{v2}}$ & RMSE  & MAE   & $S_{\text{v2}}$ & RMSE  & MAE   & $S_{\text{v2}}$ & RMSE  & MAE   & $S_{\text{v2}}$ \\
    \midrule
    ST-GCN & \underline{441.54} & \underline{362.40} & \textbf{0.339} & \underline{322.61} & \underline{277.13} & 0.236 & 675.30 & 583.30 & 0.242 & 676.18 & 581.11 & 0.243 & 467.40 & 390.58 & 0.343 \\
    SAGCN & 528.67 & 455.47 & 0.304 & 328.78 & 284.73 & 0.235 & 664.75 & 575.63 & 0.246 & 664.63 & 575.56 & 0.244 & 472.59 & 399.25 & 0.347 \\
    STNet & 523.43 & 441.31 & 0.320 & 326.67 & 282.59 & 0.235 & 664.97 & 575.76 & 0.246 & 664.86 & 575.69 & 0.244 & 508.48 & 423.28 & 0.343 \\
    GAT-LSTM & 502.53 & 422.94 & 0.317 & 326.17 & 281.42 & \underline{0.237} & 671.26 & 580.23 & 0.243 & 671.03 & 579.77 & 0.241 & 473.76 & 396.22 & 0.346 \\
    STMSGCN & 513.10 & 438.26 & 0.307 & 327.17 & 283.32 & 0.235 & 666.41 & 576.88 & 0.245 & 666.69 & 576.97 & 0.244 & \underline{465.51} & 393.15 & 0.344 \\
    AGCN-TF & 474.32 & 395.02 & 0.329 & 323.51 & 278.44 & 0.232 & 665.48 & 576.09 & 0.245 & 664.99 & 575.81 & 0.243 & 471.01 & \underline{390.25} & \underline{0.351} \\
    LOGO-S & 578.17 & 485.54 & 0.301 & 328.97 & 284.74 & 0.236 & 666.94 & 576.72 & 0.245 & 666.45 & 576.47 & 0.243 & 554.31 & 454.69 & 0.344 \\
    HierCorrPool-S & \textbf{376.30} & \textbf{319.96} & \underline{0.330} & \textbf{211.88} & \textbf{175.59} & \textbf{0.333} & \textbf{646.31} & \textbf{560.64} & \textbf{0.251} & \textbf{643.16} & \textbf{553.41} & \textbf{0.257} & \textbf{219.42} & \textbf{180.70} & \textbf{0.540} \\
    GDAGDL & 552.81 & 469.94 & 0.304 & 329.02 & 284.88 & 0.236 & \underline{664.53} & \underline{575.50} & \underline{0.246} & \underline{664.53} & \underline{575.50} & \underline{0.244} & 518.15 & 429.12 & 0.346 \\
    \bottomrule
    \bottomrule
    \end{tabular}%
    \end{adjustbox}
  \label{tab:sotaphm12cond1}%
\end{table*}%

\begin{table*}[htbp]
  \centering
  \caption{Comparisons of SOTA GNN-based methods in condition 2 and 3 of PHM 12 for bearing}
  \begin{adjustbox}{width = 1.0\textwidth,center}
    \begin{tabular}{l|ccc|ccc|ccc|ccc|ccc|ccc}
    \toprule
    \toprule
    \multirow{3}[2]{*}{Models} & \multicolumn{15}{c|}{Condition 2}                                                                                     & \multicolumn{3}{c}{Condition 3} \\
          & \multicolumn{3}{c|}{Test Bearing 3} & \multicolumn{3}{c|}{Test Bearing 4} & \multicolumn{3}{c|}{Test Bearing 5} & \multicolumn{3}{c|}{Test Bearing 6} & \multicolumn{3}{c|}{Test Bearing 7} & \multicolumn{3}{c}{Test Bearing 3} \\
          & RMSE  & MAE   & $S_{\text{v2}}$ & RMSE  & MAE   & $S_{\text{v2}}$ & RMSE  & MAE   & $S_{\text{v2}}$ & RMSE  & MAE   & $S_{\text{v2}}$ & RMSE  & MAE   & $S_{\text{v2}}$ & RMSE  & MAE   & $S_{\text{v2}}$ \\
    \midrule
    ST-GCN & 498.50 & 397.89 & 0.322 & 191.10 & 163.41 & 0.275 & 618.93 & 523.25 & 0.261 & 174.83 & 149.36 & 0.275 & 53.68 & 44.55 & \underline{0.318} & 112.98 & 93.82 & \underline{0.287} \\
    SAGCN & \underline{354.98} & \underline{305.17} & 0.369 & 182.69 & 156.54 & 0.279 & 600.60 & 513.86 & 0.262 & 170.82 & 146.35 & 0.278 & 50.04 & 43.23 & 0.304 & 105.56 & 90.32 & 0.283 \\
    STNet & 417.90 & 348.69 & 0.369 & 180.59 & 155.35 & 0.283 & \underline{583.65} & 503.86 & 0.266 & 168.68 & 145.12 & 0.282 & 52.54 & 44.77 & 0.308 & 101.96 & 88.20 & 0.285 \\
    GAT-LSTM & 466.48 & 380.88 & 0.369 & 183.37 & 157.21 & 0.282 & 591.30 & 509.00 & 0.263 & 170.39 & 146.59 & 0.279 & 54.28 & 45.90 & 0.304 & 107.09 & 91.42 & 0.280 \\
    STMSGCN & 359.19 & 305.65 & 0.362 & 180.41 & 155.51 & 0.281 & 587.97 & \underline{503.21} & \underline{0.268} & 166.24 & \underline{142.43} & \underline{0.284} & \underline{50.01} & \underline{43.13} & 0.305 & \underline{100.97} & 87.58 & 0.285 \\
    AGCN-TF & 391.34 & 327.93 & \underline{0.375} & \underline{177.17} & \underline{153.29} & \underline{0.283} & 584.43 & 504.29 & 0.266 & \underline{165.60} & 143.27 & 0.283 & 50.16 & 43.30 & 0.309 & 102.43 & 88.47 & 0.284 \\
    LOGO-S & 486.24 & 396.42 & 0.369 & 186.18 & 158.24 & 0.282 & 596.90 & 511.41 & 0.262 & 174.33 & 148.41 & 0.279 & 56.05 & 47.20 & 0.291 & 109.66 & \underline{92.94} & 0.271 \\
    HierCorrPool-S & \textbf{230.58} & \textbf{179.57} & \textbf{0.455} & \textbf{137.98} & \textbf{114.47} & \textbf{0.345} & \textbf{434.57} & \textbf{362.83} & \textbf{0.352} & \textbf{122.46} & \textbf{102.79} & \textbf{0.383} & \textbf{38.77} & \textbf{32.93} & \textbf{0.370} & \textbf{68.95} & \textbf{57.76} & \textbf{0.382} \\
    GDAGDL & 471.25 & 386.73 & 0.372 & 184.31 & 157.57 & 0.281 & 589.05 & 507.05 & 0.264 & 172.16 & 147.21 & 0.281 & 54.87 & 46.22 & 0.307 & 102.64 & 88.59 & 0.285 \\
    \bottomrule
    \bottomrule
    \end{tabular}%
    \end{adjustbox}
  \label{tab:sotaphm12cond23}%
\end{table*}%

\begin{table*}[htbp]
  \centering
  \caption{Comparisons of SOTA GNN-based methods in condition 1 of XJTU-SY for bearing}
  \begin{adjustbox}{width = 1.0\textwidth,center}
    \begin{tabular}{l|ccc|ccc|ccc|ccc|ccc}
    \toprule
    \toprule
    \multirow{3}[2]{*}{Models} & \multicolumn{15}{c}{Condition 1} \\
          & \multicolumn{3}{c|}{Test Bearing 1} & \multicolumn{3}{c|}{Test Bearing 2} & \multicolumn{3}{c|}{Test Bearing 3} & \multicolumn{3}{c|}{Test Bearing 4} & \multicolumn{3}{c}{Test Bearing 5} \\
          & RMSE  & MAE   & $S_{\text{v2}}$ & RMSE  & MAE   & $S_{\text{v2}}$ & RMSE  & MAE   & $S_{\text{v2}}$ & RMSE  & MAE   & $S_{\text{v2}}$ & RMSE  & MAE   & $S_{\text{v2}}$ \\
    \midrule
    ST-GCN & 24.55 & 20.74 & \underline{0.284} & 32.64 & 26.63 & 0.308 & 37.33 & 32.09 & 0.239 & 34.95 & 30.23 & 0.231 & 10.30 & 8.13  & \underline{0.276} \\
    SAGCN & 35.51 & 30.75 & 0.226 & 46.48 & 40.25 & 0.226 & 45.61 & 39.50 & 0.226 & 35.22 & 30.50 & 0.226 & 15.01 & 13.00 & 0.224 \\
    STNet & 32.39 & 27.73 & 0.244 & 43.34 & 37.22 & 0.253 & 43.83 & 37.98 & 0.228 & 35.22 & 30.50 & 0.226 & 13.99 & 12.07 & 0.229 \\
    GAT-LSTM & 27.27 & 22.67 & 0.262 & 36.64 & 31.54 & 0.277 & 38.31 & 33.17 & 0.231 & 35.03 & 30.14 & \underline{0.231} & 12.31 & 10.37 & 0.233 \\
    STMSGCN & 32.41 & 27.98 & 0.227 & 43.57 & 37.47 & 0.241 & 39.56 & 34.46 & 0.236 & 35.13 & 30.42 & 0.226 & 13.82 & 11.93 & 0.231 \\
    AGCN-TF & 35.02 & 30.35 & 0.227 & 45.91 & 39.81 & 0.227 & 44.21 & 38.45 & 0.227 & 35.21 & 30.48 & 0.227 & 12.49 & 10.70 & 0.237 \\
    LOGO-S & 34.34 & 29.61 & 0.227 & 43.81 & 38.12 & 0.239 & 44.35 & 38.12 & 0.233 & \underline{34.53} & \underline{29.87} & 0.226 & 14.28 & 12.02 & 0.217 \\
    HierCorrPool-S & \textbf{17.49} & \textbf{15.15} & 0.273 & \textbf{21.32} & \textbf{17.38} & \textbf{0.348} & \textbf{29.62} & \textbf{25.21} & \textbf{0.246} & \textbf{31.98} & \textbf{27.31} & \textbf{0.243} & \textbf{8.77} & \textbf{7.26} & 0.270 \\
    GDAGDL & \underline{19.90} & \underline{16.29} & \textbf{0.310} & \underline{28.94} & \underline{24.17} & \underline{0.346} & \underline{33.66} & \underline{29.33} & \underline{0.242} & 35.15 & 30.43 & 0.227 & \underline{9.29} & \underline{7.37} & \textbf{0.289} \\
    \bottomrule
    \bottomrule
    \end{tabular}%
    \end{adjustbox}
  \label{tab:sotaxjtucond1}%
\end{table*}%

\begin{table*}[htbp]
  \centering
  \caption{Comparisons of SOTA GNN-based methods in condition 2 of XJTU-SY for bearing}
  \begin{adjustbox}{width = 1.0\textwidth,center}
    \begin{tabular}{l|ccc|ccc|ccc|ccc|ccc}
    \toprule
    \toprule
    \multirow{3}[2]{*}{Models} & \multicolumn{15}{c}{Condition 2} \\
          & \multicolumn{3}{c|}{Test Bearing 1} & \multicolumn{3}{c|}{Test Bearing 2} & \multicolumn{3}{c|}{Test Bearing 3} & \multicolumn{3}{c|}{Test Bearing 4} & \multicolumn{3}{c}{Test Bearing 5} \\
          & RMSE  & MAE   & $S_{\text{v2}}$ & RMSE  & MAE   & $S_{\text{v2}}$ & RMSE  & MAE   & $S_{\text{v2}}$ & RMSE  & MAE   & $S_{\text{v2}}$ & RMSE  & MAE   & $S_{\text{v2}}$ \\
    \midrule
    ST-GCN & 155.33 & 130.11 & 0.223 & 26.88 & 21.67 & \underline{0.372} & 107.34 & 91.65 & 0.253 & 8.04  & 6.85  & 0.262 & \underline{60.07} & \underline{47.11} & 0.367 \\
    SAGCN & 141.74 & 122.75 & 0.227 & 46.48 & 40.25 & 0.226 & 153.87 & 133.25 & 0.227 & 12.12 & 10.50 & 0.224 & 97.86 & 84.75 & 0.227 \\
    STNet & 141.46 & 122.48 & 0.227 & 44.33 & 38.30 & 0.233 & 152.12 & 131.39 & 0.229 & 11.08 & 9.52  & 0.236 & 87.66 & 75.54 & 0.250 \\
    GAT-LSTM & 138.71 & 120.61 & \textbf{0.232} & 36.26 & 30.23 & 0.297 & 110.36 & 91.52 & 0.276 & 9.00  & 7.54  & 0.267 & 90.17 & 74.10 & 0.277 \\
    STMSGCN & 141.06 & 122.24 & 0.226 & 41.49 & 35.71 & 0.222 & 150.51 & 130.20 & 0.229 & 11.62 & 10.06 & 0.224 & 90.23 & 77.39 & 0.242 \\
    AGCN-TF & \underline{138.62} & \underline{120.40} & \underline{0.230} & 33.94 & 27.76 & 0.325 & 106.35 & 85.16 & 0.276 & 8.08  & 6.90  & 0.232 & 78.49 & 61.95 & 0.276 \\
    LOGO-S & 141.18 & 122.34 & 0.228 & 30.86 & 25.32 & 0.332 & 126.16 & 109.54 & 0.229 & 9.16  & 7.73  & 0.242 & 70.47 & 54.74 & \textbf{0.397} \\
    HierCorrPool-S & \textbf{131.36} & \textbf{113.81} & 0.229 & \underline{26.26} & \underline{21.91} & 0.325 & \textbf{82.01} & \textbf{67.04} & \textbf{0.329} & \textbf{6.62} & \textbf{5.44} & \textbf{0.322} & \textbf{44.28} & \textbf{36.57} & 0.321 \\
    GDAGDL & 138.93 & 120.54 & 0.225 & \textbf{24.13} & \textbf{19.01} & \textbf{0.397} & \underline{89.15} & \underline{71.85} & \underline{0.297} & \underline{7.41} & \underline{5.98} & \underline{0.287} & 60.63 & 48.04 & \underline{0.374} \\
    \bottomrule
    \bottomrule
    \end{tabular}%
    \end{adjustbox}
  \label{tab:sotaxjtucond2}%
\end{table*}%

\begin{table*}[htbp]
  \centering
  \caption{Comparisons of SOTA GNN-based methods in condition 3 of XJTU-SY for bearing}
  \begin{adjustbox}{width = 1.0\textwidth,center}
    \begin{tabular}{l|ccc|ccc|ccc|ccc|ccc}
    \toprule
    \toprule
    \multirow{3}[2]{*}{Models} & \multicolumn{15}{c}{Condition 3} \\
          & \multicolumn{3}{c|}{Test Bearing 1} & \multicolumn{3}{c|}{Test Bearing 2} & \multicolumn{3}{c|}{Test Bearing 3} & \multicolumn{3}{c|}{Test Bearing 4} & \multicolumn{3}{c}{Test Bearing 5} \\
          & RMSE  & MAE   & $S_{\text{v2}}$ & RMSE  & MAE   & $S_{\text{v2}}$ & RMSE  & MAE   & $S_{\text{v2}}$ & RMSE  & MAE   & $S_{\text{v2}}$ & RMSE  & MAE   & $S_{\text{v2}}$ \\
    \midrule
    ST-GCN & \underline{717.27} & \underline{619.20} & \underline{0.234} & \underline{546.89} & \underline{457.52} & \underline{0.275} & \underline{91.88} & \underline{77.33} & \underline{0.249} & 421.79 & 361.40 & 0.228 & \underline{23.87} & \underline{20.50} & \underline{0.281} \\
    SAGCN & 733.41 & 634.94 & 0.226 & 720.67 & 624.08 & 0.228 & 107.15 & 92.78 & 0.226 & 437.52 & 378.85 & 0.226 & 32.91 & 28.50 & 0.226 \\
    STNet & 725.99 & 629.45 & 0.227 & 719.52 & 623.35 & 0.228 & 105.18 & 91.10 & 0.227 & 434.08 & 375.59 & 0.228 & 31.36 & 27.04 & 0.231 \\
    GAT-LSTM & 723.96 & 628.41 & 0.230 & 697.36 & 600.40 & 0.233 & 102.98 & 89.57 & 0.230 & 421.22 & 366.36 & 0.229 & 32.03 & 27.80 & 0.219 \\
    STMSGCN & 726.47 & 629.34 & 0.227 & 724.00 & 625.82 & 0.226 & 105.75 & 91.60 & 0.226 & 437.18 & 378.45 & 0.227 & 30.81 & 26.31 & 0.241 \\
    AGCN-TF & 720.70 & 623.77 & 0.231 & 605.43 & 523.97 & 0.244 & 97.55 & 83.45 & 0.233 & 418.60 & 363.79 & 0.228 & 30.82 & 27.38 & 0.195 \\
    LOGO-S & 721.86 & 627.49 & 0.229 & 597.25 & 514.44 & 0.250 & 98.30 & 83.50 & 0.229 & \underline{414.27} & \underline{359.36} & \underline{0.229} & 31.23 & 26.85 & 0.228 \\
    HierCorrPool-S & \textbf{375.94} & \textbf{303.02} & \textbf{0.349} & \textbf{504.83} & \textbf{403.50} & \textbf{0.306} & \textbf{89.07} & \textbf{73.74} & \textbf{0.250} & \textbf{345.76} & \textbf{288.56} & \textbf{0.279} & \textbf{22.39} & \textbf{18.80} & \textbf{0.287} \\
    GDAGDL & 720.22 & 621.16 & 0.232 & 663.74 & 566.98 & 0.246 & 101.28 & 86.60 & 0.237 & 423.62 & 365.13 & 0.229 & 32.46 & 28.12 & 0.224 \\
    \bottomrule
    \bottomrule
    \end{tabular}%
    \end{adjustbox}
  \label{tab:sotaxjtucond3}%
\end{table*}%

The comparisons are shown from TABLE \ref{tab:sotacmpass} to \ref{tab:sotaxjtucond3}, covering C-MAPSS (TABLE \ref{tab:sotacmpass}), N-CMAPSS (TABLE \ref{tab:sotancmpass}), PHM 12 (TABLEs \ref{tab:sotaphm12cond1} and \ref{tab:sotaphm12cond23}), and XJTU-SY (TABLEs \ref{tab:sotaxjtucond1}, \ref{tab:sotaxjtucond2}, and \ref{tab:sotaxjtucond3}). The results show that most GNN-based methods perform well. Specifically, LOGO, DVGTformer, and FC-STGNN excel in C-MAPSS for aeroengine tasks, while ASTGCNN, ST-Conv, DVGTformer, and FC-STGNN excel in N-CMAPSS. For PHM 12 and XJTU-SY in bearing tasks, ST-GCN, SAGCN, AGCN-TF, HierCorrPool-S, and GDAGDL demonstrate superior performance, with especially HierCorrPool-S achieving the best results in almost all cases.

\subsection{Model Complexity Analysis}
\begin{table}[htbp]
  \centering
  \caption{Model complexity analysis on FD001 of C-MAPSS}
    \begin{tabular}{l|ccc}
    \toprule
    \toprule
    Models & FLOPs & \# Weights & Backbone \\
    \midrule
    ASTGCNN & \textbf{421,808} & \underline{7,523} & TCN+ChebNet \\
    GRU-CM & 3,520,600 & 17,441 & GRU+MPNN \\
    HAGCN & 46,278,144 & 371,143 & BiLSTM+MPNN+SAGPool \\
    ST-Conv & 951,160 & \textbf{6,877} & TCN+MPNN \\
    STFA  & 3,140,028 & 41,612 & LSTM+GAT \\
    RGCNU & 2,560,800 & 15,511 & LSTM+GCN \\
    STAGNN & 3,002,656 & 47,397 & TCN+GCN \\
    HierCorrPool & 12,117,180 & 2,025,273 & CNN+MPNN+DiffPool \\
    LOGO  & 8,018,352 & 82,527 & BiLSTM+MPNN \\
    DVGTformer & 30,904,000 & 850,557 & Self-Attention \\
    STGNN & \underline{713,216} & 25,857 & GRU+ChebNet \\
    FC-STGNN & 1,877,008 & 19,409 & CNN+MPNN \\
    \bottomrule
    \bottomrule
    \end{tabular}%
  \label{tab:complexitycmapss}%
\end{table}%

\begin{table}[htbp]
  \centering
  \caption{Model complexity analysis on Condition 1 of PHM 12}
    \begin{tabular}{l|ccc}
    \toprule
    \toprule
    Models & FLOPs & \# Weights & Backbone \\
    \midrule
    ST-GCN & \textbf{146,480} & \textbf{5,841} & TCN+MPNN \\
    SAGCN & 24,352,000 & 124,081 & Self-Attention+GCN \\
    STNet & 4,303,520 & 108,934 & CNN+ChebNet \\
    GAT-LSTM & 12,137,600 & 105,904 & LSTM+GAT \\
    STMSGCN & \underline{3,028,480} & \underline{6,050} & GRU+GCN \\
    AGCN-TF & 6,736,000 & 62,781 & MPNN only \\
    LOGO-S & 33,439,472 & 177,434 & BiLSTM+MPNN \\
    HierCorrPool-S & 21,285,120 & 1,165,559 & CNN+MPNN+DiffPool \\
    GDAGDL & 87,802,368 & 230,347 & LSTM+GAT \\
    \bottomrule
    \bottomrule
    \end{tabular}%
  \label{tab:complexityphm12}%
\end{table}%
The complexity of a model is crucial in determining its applicability to real-world systems. Even if a highly complex model performs well, it may be impractical. Thus, we assess the complexity of all GNN-based methods. Specifically, FLoating-point Operations Per second (FLOPs) and the number of model weights have been utilized to measure time and space complexity. To ensure fairness, all methods are evaluated on the same computational platform. TABLEs \ref{tab:complexitycmapss} and \ref{tab:complexityphm12} show the complexity analysis on FD001 of C-MAPSS and condition 1 of PHM 12. Two observations can be made from these results. First, compared with temporal encoders, GNNs generally have less impact on model complexity. For example, with a similar temporal encoder TCN, ASTGCNN and ST-Conv have comparable model complexities. Similarly, STFA and RGCNU show similar magnitudes of complexity based on a similar encoder LSTM. Second, in temporal encoders, TCN, CNN, and GRU-based frameworks usually exhibit lower model complexity. For instance, among TCN-based models, ASTGCNN and ST-Conv require the least FLOPs and number of weights, respectively. Similarly, GRU and CNN-based frameworks like STGNN and FC-STGNN also demonstrate lower complexity. This conclusion holds true for bearing tasks, as seen in ST-GCN and STMSGCN, which are based on TCN and GRU, respectively. It is noteworthy that due to the use of segmentation and DiffPool, HierCorrPool-S requires large weights and FLOPs, but it also achieves the best performance in most bearing cases.

\section{Limitations and Future Directions}

\subsection{\textcolor{black}{Graph Modelling with Missing Channels}}
Missing channels arise when certain data streams become unavailable due to sensor failures, software bugs, or data transmission errors, posing significant challenges for constructing accurate graphs. Some initial efforts have been made to address this problem. For example, Gupta et al. \citep{gupta2020handling_2} constructed a fully connected graph using only active sensors, which was then processed by a GNN. While this represents a promising step forward, it remains an early exploration.

To advance graph construction in the presence of missing channels, future research should focus on two key areas. First, there is a need for robust methodologies to build graphs when certain channels are missing due to unavailable data. This could involve developing adaptive graph structures that dynamically reconfigure based on the data available. Second, effective graph models must be devised to propagate information from nodes with available values (active nodes) to those without (inactive nodes), thereby reconstructing the graph structure.
Advancements in these areas will be crucial for enhancing the robustness and applicability of graph-based models in scenarios with missing data channels, ultimately improving the reliability and performance of GNNs in real-world applications.

\subsection{\textcolor{black}{Graph Generalizability}}
\textcolor{black}{Graph generalizability refers to the ability of graph construction and GNNs to perform effectively across diverse RUL prediction tasks and datasets, without the need for extensive retraining or redesigning. In the context of RUL prediction, this is particularly challenging due to the variability in sensor configurations, operating conditions, and failure modes across different applications. Existing GNNs often struggle to generalize well when applied to new tasks or datasets, as they are typically tailored to specific scenarios.}

\textcolor{black}{To enhance graph generalizability, future research could explore adaptive graph construction techniques that can dynamically adjust to varying numbers of sensor channels and different operational environments. This might involve developing more flexible graph representations that can automatically scale based on the available data, ensuring that the model remains robust even when the input structure changes. Another promising direction could be the integration of advanced DL techniques such as foundation models \citep{liu2023towards,liang2024foundation}, unsupervised domain adaptation \citep{yao2023survey,wang2024sea++}, and self-supervised learning \citep{wang2024graph}, where GNNs are trained on a variety of tasks to learn a more generalized model that can be quickly adapted to new, unseen RUL prediction scenarios with minimal fine-tuning. These approaches would significantly reduce the need for task-specific redesigns, making GNNs more practical and versatile in real-world applications.
}

\subsection{\textcolor{black}{Graph Interpretability}}
\textcolor{black}{Graph interpretability is crucial for ensuring that the predictions made by GNNs are transparent and understandable, especially in critical applications like RUL prediction, where decisions directly impact maintenance and operational safety. However, most current methods for constructing graphs rely heavily on data-driven approaches, which can obscure the rationale behind the model's predictions. Although incorporating prior knowledge into graph construction can improve interpretability, it is often too domain-specific, limiting the broader applicability of the model.}

\textcolor{black}{To address these challenges, future research could focus on developing interpretability-focused graph construction methods. One potential approach is to leverage graph signal processing techniques \citep{ortega2018graph}, such as frequency analysis, to identify and retain only the most informative features or relationships within the data. This could help in constructing graphs that are not only effective but also easier to interpret. Another direction could be the design of hybrid models that combine data-driven insights with interpretable heuristics or rules, e.g., integrating large models \citep{wang2024k}, ensuring that the resulting graphs align with domain expertise while maintaining generalizability. By prioritizing interpretability in graph construction, these models could offer more transparent and trustworthy predictions, ultimately facilitating their adoption in industrial settings where understanding the `why' behind a decision is as important as the decision itself.
}
\section{Conclusion}

This survey provides a comprehensive review of Graph Neural Networks (GNNs) for Remaining Useful Life (RUL) prediction. By examining SOTA works and introducing a novel taxonomy based on the stages of adapting GNNs for RUL prediction, we offer an in-depth understanding of the unique challenges and considerations present in each stage of the GNN workflow. Our survey also includes a detailed experimental evaluation of various SOTA GNN methods for RUL prediction, conducted under consistent experimental settings to ensure fair comparisons across techniques. This rigorous analysis serves as a valuable guide for both researchers and practitioners, offering insights into the strengths and limitations of current approaches. Additionally, we identify and discuss several promising research directions to address existing challenges in this field. Ultimately, this survey contributes to the ongoing development of more robust, efficient, and accurate GNN-based methods for predicting the remaining useful life of complex systems.













\bibliographystyle{cas-model2-names}

\bibliography{cas-refs}

\begin{thebibliography}{110}
\expandafter\ifx\csname natexlab\endcsname\relax\def\natexlab#1{#1}\fi
\providecommand{\url}[1]{\texttt{#1}}
\providecommand{\href}[2]{#2}
\providecommand{\path}[1]{#1}
\providecommand{\DOIprefix}{doi:}
\providecommand{\ArXivprefix}{arXiv:}
\providecommand{\URLprefix}{URL: }
\providecommand{\Pubmedprefix}{pmid:}
\providecommand{\doi}[1]{\href{http://dx.doi.org/#1}{\path{#1}}}
\providecommand{\Pubmed}[1]{\href{pmid:#1}{\path{#1}}}
\providecommand{\bibinfo}[2]{#2}
\ifx\xfnm\relax \def\xfnm[#1]{\unskip,\space#1}\fi
\bibitem[{Arias~Chao et~al.(2021)Arias~Chao, Kulkarni, Goebel and Fink}]{arias2021aircraft}
\bibinfo{author}{Arias~Chao, M.}, \bibinfo{author}{Kulkarni, C.}, \bibinfo{author}{Goebel, K.}, \bibinfo{author}{Fink, O.}, \bibinfo{year}{2021}.
\newblock \bibinfo{title}{Aircraft engine run-to-failure dataset under real flight conditions for prognostics and diagnostics}.
\newblock \bibinfo{journal}{Data} \bibinfo{volume}{6}, \bibinfo{pages}{5}.
\bibitem[{Cao et~al.(2020)Cao, Wang, Duan, Zhang, Zhu, Huang, Tong, Xu, Bai, Tong et~al.}]{cao2020spectral}
\bibinfo{author}{Cao, D.}, \bibinfo{author}{Wang, Y.}, \bibinfo{author}{Duan, J.}, \bibinfo{author}{Zhang, C.}, \bibinfo{author}{Zhu, X.}, \bibinfo{author}{Huang, C.}, \bibinfo{author}{Tong, Y.}, \bibinfo{author}{Xu, B.}, \bibinfo{author}{Bai, J.}, \bibinfo{author}{Tong, J.}, et~al., \bibinfo{year}{2020}.
\newblock \bibinfo{title}{Spectral temporal graph neural network for multivariate time-series forecasting}.
\newblock \bibinfo{journal}{Advances in neural information processing systems} \bibinfo{volume}{33}, \bibinfo{pages}{17766--17778}.
\bibitem[{Cao et~al.(2023)Cao, Zhuang, Jia, Zhao, Yan and Liu}]{cao2023picture_20}
\bibinfo{author}{Cao, Y.}, \bibinfo{author}{Zhuang, J.}, \bibinfo{author}{Jia, M.}, \bibinfo{author}{Zhao, X.}, \bibinfo{author}{Yan, X.}, \bibinfo{author}{Liu, Z.}, \bibinfo{year}{2023}.
\newblock \bibinfo{title}{Picture-in-picture strategy based complex graph neural network for remaining useful life prediction of rotating machinery}.
\newblock \bibinfo{journal}{IEEE Transactions on Instrumentation and Measurement} .
\bibitem[{Chaoying et~al.()Chaoying, Jie and Kaibo}]{Yang2023path_14}
\bibinfo{author}{Chaoying, Y.}, \bibinfo{author}{Jie, L.}, \bibinfo{author}{Kaibo, Z.}, .
\newblock \bibinfo{title}{Path graph attention network-based bearing remaining useful life prediction method}.
\newblock \bibinfo{journal}{Journal of Mechanical Engineering} \bibinfo{volume}{59}, \bibinfo{pages}{195--201}.
\bibitem[{Chen et~al.(2023)Chen, Huang, Chen, Mao and Li}]{chen2023transfer}
\bibinfo{author}{Chen, J.}, \bibinfo{author}{Huang, R.}, \bibinfo{author}{Chen, Z.}, \bibinfo{author}{Mao, W.}, \bibinfo{author}{Li, W.}, \bibinfo{year}{2023}.
\newblock \bibinfo{title}{Transfer learning algorithms for bearing remaining useful life prediction: A comprehensive review from an industrial application perspective}.
\newblock \bibinfo{journal}{Mechanical Systems and Signal Processing} \bibinfo{volume}{193}, \bibinfo{pages}{110239}.
\bibitem[{Chen and Zeng(2023)}]{chen2023convolution_24}
\bibinfo{author}{Chen, X.}, \bibinfo{author}{Zeng, M.}, \bibinfo{year}{2023}.
\newblock \bibinfo{title}{Convolution-graph attention network with sensor embeddings for remaining useful life prediction of turbofan engines}.
\newblock \bibinfo{journal}{IEEE Sensors Journal} .
\bibitem[{Chen et~al.(2020)Chen, Wu, Zhao, Guretno, Yan and Li}]{chen2020machine}
\bibinfo{author}{Chen, Z.}, \bibinfo{author}{Wu, M.}, \bibinfo{author}{Zhao, R.}, \bibinfo{author}{Guretno, F.}, \bibinfo{author}{Yan, R.}, \bibinfo{author}{Li, X.}, \bibinfo{year}{2020}.
\newblock \bibinfo{title}{Machine remaining useful life prediction via an attention-based deep learning approach}.
\newblock \bibinfo{journal}{IEEE Transactions on Industrial Electronics} \bibinfo{volume}{68}, \bibinfo{pages}{2521--2531}.
\bibitem[{Chen et~al.(2021)Chen, Xu, Alippi, Ding, Shardt, Peng and Yang}]{chen2021graph}
\bibinfo{author}{Chen, Z.}, \bibinfo{author}{Xu, J.}, \bibinfo{author}{Alippi, C.}, \bibinfo{author}{Ding, S.X.}, \bibinfo{author}{Shardt, Y.}, \bibinfo{author}{Peng, T.}, \bibinfo{author}{Yang, C.}, \bibinfo{year}{2021}.
\newblock \bibinfo{title}{Graph neural network-based fault diagnosis: a review}.
\newblock \bibinfo{journal}{arXiv preprint arXiv:2111.08185} .
\bibitem[{Cheng et~al.(2024)Cheng, Zhang, Wang, Yang, Li and Wang}]{cheng2024research_57}
\bibinfo{author}{Cheng, K.}, \bibinfo{author}{Zhang, K.}, \bibinfo{author}{Wang, Y.}, \bibinfo{author}{Yang, C.}, \bibinfo{author}{Li, J.}, \bibinfo{author}{Wang, Y.}, \bibinfo{year}{2024}.
\newblock \bibinfo{title}{Research on gas turbine health assessment method based on physical prior knowledge and spatial-temporal graph neural network}.
\newblock \bibinfo{journal}{Applied Energy} \bibinfo{volume}{367}, \bibinfo{pages}{123419}.
\bibitem[{Cui et~al.(2024a)Cui, Wang, Liu and Wang}]{cui2024adaptive_55}
\bibinfo{author}{Cui, L.}, \bibinfo{author}{Wang, X.}, \bibinfo{author}{Liu, D.}, \bibinfo{author}{Wang, H.}, \bibinfo{year}{2024}a.
\newblock \bibinfo{title}{An adaptive sparse graph learning method based on digital twin dictionary for remaining useful life prediction of rolling element bearings}.
\newblock \bibinfo{journal}{IEEE Transactions on Industrial Informatics} .
\bibitem[{Cui et~al.(2024b)Cui, Xiao, Liu and Han}]{cui2024digital_43}
\bibinfo{author}{Cui, L.}, \bibinfo{author}{Xiao, Y.}, \bibinfo{author}{Liu, D.}, \bibinfo{author}{Han, H.}, \bibinfo{year}{2024}b.
\newblock \bibinfo{title}{Digital twin-driven graph domain adaptation neural network for remaining useful life prediction of rolling bearing}.
\newblock \bibinfo{journal}{Reliability Engineering \& System Safety} , \bibinfo{pages}{109991}.
\bibitem[{Defferrard et~al.(2016)Defferrard, Bresson and Vandergheynst}]{defferrard2016convolutional}
\bibinfo{author}{Defferrard, M.}, \bibinfo{author}{Bresson, X.}, \bibinfo{author}{Vandergheynst, P.}, \bibinfo{year}{2016}.
\newblock \bibinfo{title}{Convolutional neural networks on graphs with fast localized spectral filtering}.
\newblock \bibinfo{journal}{Advances in neural information processing systems} \bibinfo{volume}{29}.
\bibitem[{Diehl(2019)}]{diehl2019edge}
\bibinfo{author}{Diehl, F.}, \bibinfo{year}{2019}.
\newblock \bibinfo{title}{Edge contraction pooling for graph neural networks}.
\newblock \bibinfo{journal}{arXiv preprint arXiv:1905.10990} .
\bibitem[{Ding et~al.(2024)Ding, Xia, Zhao and Jia}]{ding2024graph_48}
\bibinfo{author}{Ding, P.}, \bibinfo{author}{Xia, J.}, \bibinfo{author}{Zhao, X.}, \bibinfo{author}{Jia, M.}, \bibinfo{year}{2024}.
\newblock \bibinfo{title}{Graph structure few-shot prognostics for machinery remaining useful life prediction under variable operating conditions}.
\newblock \bibinfo{journal}{Advanced Engineering Informatics} \bibinfo{volume}{60}, \bibinfo{pages}{102360}.
\bibitem[{Ferreira and Gon{\c{c}}alves(2022)}]{ferreira2022remaining}
\bibinfo{author}{Ferreira, C.}, \bibinfo{author}{Gon{\c{c}}alves, G.}, \bibinfo{year}{2022}.
\newblock \bibinfo{title}{Remaining useful life prediction and challenges: A literature review on the use of machine learning methods}.
\newblock \bibinfo{journal}{Journal of Manufacturing Systems} \bibinfo{volume}{63}, \bibinfo{pages}{550--562}.
\bibitem[{Gao et~al.(2024)Gao, Jiang, Wu, Dai and Zhu}]{gao2024nonlinear_52}
\bibinfo{author}{Gao, Z.}, \bibinfo{author}{Jiang, W.}, \bibinfo{author}{Wu, J.}, \bibinfo{author}{Dai, T.}, \bibinfo{author}{Zhu, H.}, \bibinfo{year}{2024}.
\newblock \bibinfo{title}{Nonlinear slow-varying dynamics-assisted temporal graph transformer network for remaining useful life prediction}.
\newblock \bibinfo{journal}{Reliability Engineering \& System Safety} \bibinfo{volume}{248}, \bibinfo{pages}{110162}.
\bibitem[{Gilmer et~al.(2017)Gilmer, Schoenholz, Riley, Vinyals and Dahl}]{gilmer2017neural}
\bibinfo{author}{Gilmer, J.}, \bibinfo{author}{Schoenholz, S.S.}, \bibinfo{author}{Riley, P.F.}, \bibinfo{author}{Vinyals, O.}, \bibinfo{author}{Dahl, G.E.}, \bibinfo{year}{2017}.
\newblock \bibinfo{title}{Neural message passing for quantum chemistry}, in: \bibinfo{booktitle}{International conference on machine learning}, \bibinfo{organization}{PMLR}. pp. \bibinfo{pages}{1263--1272}.
\bibitem[{Gupta et~al.(2020)Gupta, Narwariya, Malhotra, Vig and Shroff}]{gupta2020handling_2}
\bibinfo{author}{Gupta, V.}, \bibinfo{author}{Narwariya, J.}, \bibinfo{author}{Malhotra, P.}, \bibinfo{author}{Vig, L.}, \bibinfo{author}{Shroff, G.}, \bibinfo{year}{2020}.
\newblock \bibinfo{title}{Handling variable-dimensional time series with graph neural networks}.
\newblock \bibinfo{journal}{arXiv preprint arXiv:2007.00411} .
\bibitem[{He et~al.(2023)He, Su, Zio, Peng, Fan, Yang, Yang and Zhang}]{he2023systematic_16}
\bibinfo{author}{He, Y.}, \bibinfo{author}{Su, H.}, \bibinfo{author}{Zio, E.}, \bibinfo{author}{Peng, S.}, \bibinfo{author}{Fan, L.}, \bibinfo{author}{Yang, Z.}, \bibinfo{author}{Yang, Z.}, \bibinfo{author}{Zhang, J.}, \bibinfo{year}{2023}.
\newblock \bibinfo{title}{A systematic method of remaining useful life estimation based on physics-informed graph neural networks with multisensor data}.
\newblock \bibinfo{journal}{Reliability Engineering \& System Safety} \bibinfo{volume}{237}, \bibinfo{pages}{109333}.
\bibitem[{Heimes(2008)}]{heimes2008recurrent}
\bibinfo{author}{Heimes, F.O.}, \bibinfo{year}{2008}.
\newblock \bibinfo{title}{Recurrent neural networks for remaining useful life estimation}, in: \bibinfo{booktitle}{2008 international conference on prognostics and health management}, \bibinfo{organization}{IEEE}. pp. \bibinfo{pages}{1--6}.
\bibitem[{Huang et~al.(2024)Huang, He and Sick}]{huang2024spatio_42}
\bibinfo{author}{Huang, Z.}, \bibinfo{author}{He, Y.}, \bibinfo{author}{Sick, B.}, \bibinfo{year}{2024}.
\newblock \bibinfo{title}{Spatio-temporal attention graph neural network for remaining useful life prediction}.
\newblock \bibinfo{journal}{arXiv preprint arXiv:2401.15964} .
\bibitem[{Jayasinghe et~al.(2019)Jayasinghe, Samarasinghe, Yuenv, Low and Ge}]{jayasinghe2019temporal}
\bibinfo{author}{Jayasinghe, L.}, \bibinfo{author}{Samarasinghe, T.}, \bibinfo{author}{Yuenv, C.}, \bibinfo{author}{Low, J.C.N.}, \bibinfo{author}{Ge, S.S.}, \bibinfo{year}{2019}.
\newblock \bibinfo{title}{Temporal convolutional memory networks for remaining useful life estimation of industrial machinery}, in: \bibinfo{booktitle}{2019 ieee international conference on industrial technology (icit)}, \bibinfo{organization}{IEEE}. pp. \bibinfo{pages}{915--920}.
\bibitem[{Jiang and Luo(2022)}]{jiang2022graph}
\bibinfo{author}{Jiang, W.}, \bibinfo{author}{Luo, J.}, \bibinfo{year}{2022}.
\newblock \bibinfo{title}{Graph neural network for traffic forecasting: A survey}.
\newblock \bibinfo{journal}{Expert Systems with Applications} \bibinfo{volume}{207}, \bibinfo{pages}{117921}.
\bibitem[{Jiang et~al.(2022)Jiang, Dai, Fang, Zhong and Cao}]{jiang2022electrical_13}
\bibinfo{author}{Jiang, Y.}, \bibinfo{author}{Dai, P.}, \bibinfo{author}{Fang, P.}, \bibinfo{author}{Zhong, R.Y.}, \bibinfo{author}{Cao, X.}, \bibinfo{year}{2022}.
\newblock \bibinfo{title}{Electrical-stgcn: An electrical spatio-temporal graph convolutional network for intelligent predictive maintenance}.
\newblock \bibinfo{journal}{IEEE Transactions on Industrial Informatics} \bibinfo{volume}{18}, \bibinfo{pages}{8509--8518}.
\bibitem[{Jin et~al.(2022)Jin, Wu, Wu, Gao, Chen and Li}]{jin2022position}
\bibinfo{author}{Jin, R.}, \bibinfo{author}{Wu, M.}, \bibinfo{author}{Wu, K.}, \bibinfo{author}{Gao, K.}, \bibinfo{author}{Chen, Z.}, \bibinfo{author}{Li, X.}, \bibinfo{year}{2022}.
\newblock \bibinfo{title}{Position encoding based convolutional neural networks for machine remaining useful life prediction}.
\newblock \bibinfo{journal}{IEEE/CAA Journal of Automatica Sinica} \bibinfo{volume}{9}, \bibinfo{pages}{1427--1439}.
\bibitem[{Karatzinis et~al.(2024)Karatzinis, Boutalis and Van~Vaerenbergh}]{karatzinis2024aircraft}
\bibinfo{author}{Karatzinis, G.D.}, \bibinfo{author}{Boutalis, Y.S.}, \bibinfo{author}{Van~Vaerenbergh, S.}, \bibinfo{year}{2024}.
\newblock \bibinfo{title}{Aircraft engine remaining useful life prediction: A comparison study of kernel adaptive filtering architectures}.
\newblock \bibinfo{journal}{Mechanical Systems and Signal Processing} \bibinfo{volume}{218}, \bibinfo{pages}{111551}.
\bibitem[{Kipf and Welling(2016)}]{kipf2016semi}
\bibinfo{author}{Kipf, T.N.}, \bibinfo{author}{Welling, M.}, \bibinfo{year}{2016}.
\newblock \bibinfo{title}{Semi-supervised classification with graph convolutional networks}.
\newblock \bibinfo{journal}{arXiv preprint arXiv:1609.02907} .
\bibitem[{Klepl et~al.(2024)Klepl, Wu and He}]{klepl2024graph}
\bibinfo{author}{Klepl, D.}, \bibinfo{author}{Wu, M.}, \bibinfo{author}{He, F.}, \bibinfo{year}{2024}.
\newblock \bibinfo{title}{Graph neural network-based eeg classification: A survey}.
\newblock \bibinfo{journal}{IEEE Transactions on Neural Systems and Rehabilitation Engineering} .
\bibitem[{Kong et~al.(2024)Kong, Jin, Wang and Xu}]{kong2024spatio_41}
\bibinfo{author}{Kong, Z.}, \bibinfo{author}{Jin, X.}, \bibinfo{author}{Wang, F.}, \bibinfo{author}{Xu, Z.}, \bibinfo{year}{2024}.
\newblock \bibinfo{title}{Spatio-temporal propagation: An extended message passing graph neural network for remaining useful life prediction}.
\newblock \bibinfo{journal}{IEEE Sensors Journal} .
\bibitem[{Kong et~al.(2022)Kong, Jin, Xu and Zhang}]{kong2022spatio_9}
\bibinfo{author}{Kong, Z.}, \bibinfo{author}{Jin, X.}, \bibinfo{author}{Xu, Z.}, \bibinfo{author}{Zhang, B.}, \bibinfo{year}{2022}.
\newblock \bibinfo{title}{Spatio-temporal fusion attention: A novel approach for remaining useful life prediction based on graph neural network}.
\newblock \bibinfo{journal}{IEEE Transactions on Instrumentation and Measurement} \bibinfo{volume}{71}, \bibinfo{pages}{1--12}.
\bibitem[{Lee et~al.(2019)Lee, Lee and Kang}]{lee2019self}
\bibinfo{author}{Lee, J.}, \bibinfo{author}{Lee, I.}, \bibinfo{author}{Kang, J.}, \bibinfo{year}{2019}.
\newblock \bibinfo{title}{Self-attention graph pooling}, in: \bibinfo{booktitle}{International conference on machine learning}, \bibinfo{organization}{PMLR}. pp. \bibinfo{pages}{3734--3743}.
\bibitem[{Lei et~al.(2016)Lei, Li, Gontarz, Lin, Radkowski and Dybala}]{lei2016model}
\bibinfo{author}{Lei, Y.}, \bibinfo{author}{Li, N.}, \bibinfo{author}{Gontarz, S.}, \bibinfo{author}{Lin, J.}, \bibinfo{author}{Radkowski, S.}, \bibinfo{author}{Dybala, J.}, \bibinfo{year}{2016}.
\newblock \bibinfo{title}{A model-based method for remaining useful life prediction of machinery}.
\newblock \bibinfo{journal}{IEEE Transactions on reliability} \bibinfo{volume}{65}, \bibinfo{pages}{1314--1326}.
\bibitem[{Li et~al.(2024)Li, Zhang, Li and Si}]{li2024review}
\bibinfo{author}{Li, H.}, \bibinfo{author}{Zhang, Z.}, \bibinfo{author}{Li, T.}, \bibinfo{author}{Si, X.}, \bibinfo{year}{2024}.
\newblock \bibinfo{title}{A review on physics-informed data-driven remaining useful life prediction: Challenges and opportunities}.
\newblock \bibinfo{journal}{Mechanical Systems and Signal Processing} \bibinfo{volume}{209}, \bibinfo{pages}{111120}.
\bibitem[{Li et~al.(2021a)Li, Liu and Yang}]{li2021remaining_6}
\bibinfo{author}{Li, P.}, \bibinfo{author}{Liu, X.}, \bibinfo{author}{Yang, Y.}, \bibinfo{year}{2021}a.
\newblock \bibinfo{title}{Remaining useful life prognostics of bearings based on a novel spatial graph-temporal convolution network}.
\newblock \bibinfo{journal}{Sensors} \bibinfo{volume}{21}, \bibinfo{pages}{4217}.
\bibitem[{Li et~al.(2021b)Li, Zhao, Sun, Yan and Chen}]{li2021hierarchical_5}
\bibinfo{author}{Li, T.}, \bibinfo{author}{Zhao, Z.}, \bibinfo{author}{Sun, C.}, \bibinfo{author}{Yan, R.}, \bibinfo{author}{Chen, X.}, \bibinfo{year}{2021}b.
\newblock \bibinfo{title}{Hierarchical attention graph convolutional network to fuse multi-sensor signals for remaining useful life prediction}.
\newblock \bibinfo{journal}{Reliability Engineering \& System Safety} \bibinfo{volume}{215}, \bibinfo{pages}{107878}.
\bibitem[{Li et~al.(2022)Li, Zhou, Li, Sun, Yan and Chen}]{li2022emerging}
\bibinfo{author}{Li, T.}, \bibinfo{author}{Zhou, Z.}, \bibinfo{author}{Li, S.}, \bibinfo{author}{Sun, C.}, \bibinfo{author}{Yan, R.}, \bibinfo{author}{Chen, X.}, \bibinfo{year}{2022}.
\newblock \bibinfo{title}{The emerging graph neural networks for intelligent fault diagnostics and prognostics: A guideline and a benchmark study}.
\newblock \bibinfo{journal}{Mechanical Systems and Signal Processing} \bibinfo{volume}{168}, \bibinfo{pages}{108653}.
\bibitem[{Li et~al.(2018)Li, Ding and Sun}]{li2018remaining}
\bibinfo{author}{Li, X.}, \bibinfo{author}{Ding, Q.}, \bibinfo{author}{Sun, J.Q.}, \bibinfo{year}{2018}.
\newblock \bibinfo{title}{Remaining useful life estimation in prognostics using deep convolution neural networks}.
\newblock \bibinfo{journal}{Reliability Engineering \& System Safety} \bibinfo{volume}{172}, \bibinfo{pages}{1--11}.
\bibitem[{Li et~al.(2023)Li, Chen, Hu and Zhang}]{li2023remaining_36}
\bibinfo{author}{Li, Y.}, \bibinfo{author}{Chen, Y.}, \bibinfo{author}{Hu, Z.}, \bibinfo{author}{Zhang, H.}, \bibinfo{year}{2023}.
\newblock \bibinfo{title}{Remaining useful life prediction of aero-engine enabled by fusing knowledge and deep learning models}.
\newblock \bibinfo{journal}{Reliability Engineering \& System Safety} \bibinfo{volume}{229}, \bibinfo{pages}{108869}.
\bibitem[{Liang et~al.(2022)Liang, Qian, Yu, Griffith and Golmie}]{liang2022survey}
\bibinfo{author}{Liang, F.}, \bibinfo{author}{Qian, C.}, \bibinfo{author}{Yu, W.}, \bibinfo{author}{Griffith, D.}, \bibinfo{author}{Golmie, N.}, \bibinfo{year}{2022}.
\newblock \bibinfo{title}{Survey of graph neural networks and applications}.
\newblock \bibinfo{journal}{Wireless Communications and Mobile Computing} \bibinfo{volume}{2022}, \bibinfo{pages}{9261537}.
\bibitem[{Liang et~al.(2023)Liang, Li, Wang, Yuan and Zhang}]{liang2023remaining_23}
\bibinfo{author}{Liang, P.}, \bibinfo{author}{Li, Y.}, \bibinfo{author}{Wang, B.}, \bibinfo{author}{Yuan, X.}, \bibinfo{author}{Zhang, L.}, \bibinfo{year}{2023}.
\newblock \bibinfo{title}{Remaining useful life prediction via a deep adaptive transformer framework enhanced by graph attention network}.
\newblock \bibinfo{journal}{International Journal of Fatigue} \bibinfo{volume}{174}, \bibinfo{pages}{107722}.
\bibitem[{Liang et~al.(2024)Liang, Wen, Nie, Jiang, Jin, Song, Pan and Wen}]{liang2024foundation}
\bibinfo{author}{Liang, Y.}, \bibinfo{author}{Wen, H.}, \bibinfo{author}{Nie, Y.}, \bibinfo{author}{Jiang, Y.}, \bibinfo{author}{Jin, M.}, \bibinfo{author}{Song, D.}, \bibinfo{author}{Pan, S.}, \bibinfo{author}{Wen, Q.}, \bibinfo{year}{2024}.
\newblock \bibinfo{title}{Foundation models for time series analysis: A tutorial and survey}, in: \bibinfo{booktitle}{Proceedings of the 30th ACM SIGKDD Conference on Knowledge Discovery and Data Mining}, pp. \bibinfo{pages}{6555--6565}.
\bibitem[{Liu et~al.(2023a)Liu, Wang, Xie, Wu and Li}]{liu2023condition_31}
\bibinfo{author}{Liu, J.}, \bibinfo{author}{Wang, X.}, \bibinfo{author}{Xie, F.}, \bibinfo{author}{Wu, S.}, \bibinfo{author}{Li, D.}, \bibinfo{year}{2023}a.
\newblock \bibinfo{title}{Condition monitoring of wind turbines with the implementation of spatio-temporal graph neural network}.
\newblock \bibinfo{journal}{Engineering Applications of Artificial Intelligence} \bibinfo{volume}{121}, \bibinfo{pages}{106000}.
\bibitem[{Liu et~al.(2023b)Liu, Yang, Lu, Chen, Li, Zhang, Bai, Fang, Sun, Yu et~al.}]{liu2023towards}
\bibinfo{author}{Liu, J.}, \bibinfo{author}{Yang, C.}, \bibinfo{author}{Lu, Z.}, \bibinfo{author}{Chen, J.}, \bibinfo{author}{Li, Y.}, \bibinfo{author}{Zhang, M.}, \bibinfo{author}{Bai, T.}, \bibinfo{author}{Fang, Y.}, \bibinfo{author}{Sun, L.}, \bibinfo{author}{Yu, P.S.}, et~al., \bibinfo{year}{2023}b.
\newblock \bibinfo{title}{Towards graph foundation models: A survey and beyond}.
\newblock \bibinfo{journal}{arXiv preprint arXiv:2310.11829} .
\bibitem[{Liu et~al.(2023c)Liu, Song, Sun, Gong and Li}]{liu2023mmoe_32}
\bibinfo{author}{Liu, L.}, \bibinfo{author}{Song, X.}, \bibinfo{author}{Sun, B.}, \bibinfo{author}{Gong, G.}, \bibinfo{author}{Li, W.}, \bibinfo{year}{2023}c.
\newblock \bibinfo{title}{Mmoe-gat: A multi-gate mixture-of-experts boosted graph attention network for aircraft engine remaining useful life prediction}, in: \bibinfo{booktitle}{Asia Simulation Conference}, \bibinfo{organization}{Springer}. pp. \bibinfo{pages}{451--465}.
\bibitem[{Liu et~al.(2023d)Liu, Guo, Chen and Zhang}]{liu2023aero_37}
\bibinfo{author}{Liu, N.}, \bibinfo{author}{Guo, J.}, \bibinfo{author}{Chen, S.}, \bibinfo{author}{Zhang, X.}, \bibinfo{year}{2023}d.
\newblock \bibinfo{title}{Aero-engines remaining useful life prognostics based on multi-hierarchical gated recurrent graph convolutional network}, in: \bibinfo{booktitle}{2023 International Conference on Cyber-Physical Social Intelligence (ICCSI)}, \bibinfo{organization}{IEEE}. pp. \bibinfo{pages}{642--647}.
\bibitem[{Long et~al.(2023)Long, Zhang, Long, He, Liu and Li}]{long2023gnnrotating_33}
\bibinfo{author}{Long, K.}, \bibinfo{author}{Zhang, R.}, \bibinfo{author}{Long, J.}, \bibinfo{author}{He, N.}, \bibinfo{author}{Liu, Y.}, \bibinfo{author}{Li, C.}, \bibinfo{year}{2023}.
\newblock \bibinfo{title}{A graph neural network-based method for predicting remaining useful life of rotating machinery}, in: \bibinfo{booktitle}{2023 Prognostics and Health Management Conference (PHM)}, pp. \bibinfo{pages}{287--292}.
\newblock \DOIprefix\doi{10.1109/PHM58589.2023.00060}.
\bibitem[{Lv and Liu(2023)}]{lv2023new_27}
\bibinfo{author}{Lv, S.}, \bibinfo{author}{Liu, S.}, \bibinfo{year}{2023}.
\newblock \bibinfo{title}{A new method for remaining useful life prediction by implementing joint learning of sensor dynamic graph and spatio-temporal features}.
\newblock \bibinfo{journal}{Measurement Science and Technology} \bibinfo{volume}{34}, \bibinfo{pages}{095123}.
\bibitem[{Ma et~al.(2024)Ma, Wang and Zhong}]{ma2024transformer_49}
\bibinfo{author}{Ma, M.}, \bibinfo{author}{Wang, Z.}, \bibinfo{author}{Zhong, Z.}, \bibinfo{year}{2024}.
\newblock \bibinfo{title}{Transformer encoder enhanced by an adaptive graph convolutional neural network for prediction of aero-engines’ remaining useful life}.
\newblock \bibinfo{journal}{Aerospace} \bibinfo{volume}{11}, \bibinfo{pages}{289}.
\bibitem[{Mo and Iacca(2022)}]{mo2022multi}
\bibinfo{author}{Mo, H.}, \bibinfo{author}{Iacca, G.}, \bibinfo{year}{2022}.
\newblock \bibinfo{title}{Multi-objective optimization of extreme learning machine for remaining useful life prediction}, in: \bibinfo{booktitle}{International Conference on the Applications of Evolutionary Computation (Part of EvoStar)}, \bibinfo{organization}{Springer}. pp. \bibinfo{pages}{191--206}.
\bibitem[{Nectoux et~al.(2012)Nectoux, Gouriveau, Medjaher, Ramasso, Chebel-Morello, Zerhouni and Varnier}]{nectoux2012pronostia}
\bibinfo{author}{Nectoux, P.}, \bibinfo{author}{Gouriveau, R.}, \bibinfo{author}{Medjaher, K.}, \bibinfo{author}{Ramasso, E.}, \bibinfo{author}{Chebel-Morello, B.}, \bibinfo{author}{Zerhouni, N.}, \bibinfo{author}{Varnier, C.}, \bibinfo{year}{2012}.
\newblock \bibinfo{title}{Pronostia: An experimental platform for bearings accelerated degradation tests.}, in: \bibinfo{booktitle}{IEEE International Conference on Prognostics and Health Management, PHM'12.}, \bibinfo{organization}{IEEE Catalog Number: CPF12PHM-CDR}. pp. \bibinfo{pages}{1--8}.
\bibitem[{Ortega et~al.(2018)Ortega, Frossard, Kova{\v{c}}evi{\'c}, Moura and Vandergheynst}]{ortega2018graph}
\bibinfo{author}{Ortega, A.}, \bibinfo{author}{Frossard, P.}, \bibinfo{author}{Kova{\v{c}}evi{\'c}, J.}, \bibinfo{author}{Moura, J.M.}, \bibinfo{author}{Vandergheynst, P.}, \bibinfo{year}{2018}.
\newblock \bibinfo{title}{Graph signal processing: Overview, challenges, and applications}.
\newblock \bibinfo{journal}{Proceedings of the IEEE} \bibinfo{volume}{106}, \bibinfo{pages}{808--828}.
\bibitem[{Qiu et~al.(2023)Qiu, Niu, Shang, Gao and Xu}]{qiu2023piecewise}
\bibinfo{author}{Qiu, H.}, \bibinfo{author}{Niu, Y.}, \bibinfo{author}{Shang, J.}, \bibinfo{author}{Gao, L.}, \bibinfo{author}{Xu, D.}, \bibinfo{year}{2023}.
\newblock \bibinfo{title}{A piecewise method for bearing remaining useful life estimation using temporal convolutional networks}.
\newblock \bibinfo{journal}{Journal of Manufacturing Systems} \bibinfo{volume}{68}, \bibinfo{pages}{227--241}.
\bibitem[{Saxena et~al.(2008)Saxena, Goebel, Simon and Eklund}]{saxena2008damage}
\bibinfo{author}{Saxena, A.}, \bibinfo{author}{Goebel, K.}, \bibinfo{author}{Simon, D.}, \bibinfo{author}{Eklund, N.}, \bibinfo{year}{2008}.
\newblock \bibinfo{title}{Damage propagation modeling for aircraft engine run-to-failure simulation}, in: \bibinfo{booktitle}{2008 international conference on prognostics and health management}, \bibinfo{organization}{IEEE}. pp. \bibinfo{pages}{1--9}.
\bibitem[{Shang et~al.(2024)Shang, Xu, Qiu, Gao, Jiang and Yi}]{shang2024novel}
\bibinfo{author}{Shang, J.}, \bibinfo{author}{Xu, D.}, \bibinfo{author}{Qiu, H.}, \bibinfo{author}{Gao, L.}, \bibinfo{author}{Jiang, C.}, \bibinfo{author}{Yi, P.}, \bibinfo{year}{2024}.
\newblock \bibinfo{title}{A novel data augmentation framework for remaining useful life estimation with dense convolutional regression network}.
\newblock \bibinfo{journal}{Journal of Manufacturing Systems} \bibinfo{volume}{74}, \bibinfo{pages}{30--40}.
\bibitem[{Si et~al.(2011)Si, Wang, Hu and Zhou}]{si2011remaining}
\bibinfo{author}{Si, X.S.}, \bibinfo{author}{Wang, W.}, \bibinfo{author}{Hu, C.H.}, \bibinfo{author}{Zhou, D.H.}, \bibinfo{year}{2011}.
\newblock \bibinfo{title}{Remaining useful life estimation--a review on the statistical data driven approaches}.
\newblock \bibinfo{journal}{European journal of operational research} \bibinfo{volume}{213}, \bibinfo{pages}{1--14}.
\bibitem[{Sikorska et~al.(2011)Sikorska, Hodkiewicz and Ma}]{sikorska2011prognostic}
\bibinfo{author}{Sikorska, J.Z.}, \bibinfo{author}{Hodkiewicz, M.}, \bibinfo{author}{Ma, L.}, \bibinfo{year}{2011}.
\newblock \bibinfo{title}{Prognostic modelling options for remaining useful life estimation by industry}.
\newblock \bibinfo{journal}{Mechanical systems and signal processing} \bibinfo{volume}{25}, \bibinfo{pages}{1803--1836}.
\bibitem[{Song et~al.(2024)Song, Jin, Lin, Zhao, Wei and Wang}]{song2024remaining_45}
\bibinfo{author}{Song, L.}, \bibinfo{author}{Jin, Y.}, \bibinfo{author}{Lin, T.}, \bibinfo{author}{Zhao, S.}, \bibinfo{author}{Wei, Z.}, \bibinfo{author}{Wang, H.}, \bibinfo{year}{2024}.
\newblock \bibinfo{title}{Remaining useful life prediction method based on the spatiotemporal graph and gcn nested parallel route model}.
\newblock \bibinfo{journal}{IEEE Transactions on Instrumentation and Measurement} .
\bibitem[{Tianhao et~al.(2023)Tianhao, Kang, Jie, Ruyi and Weihua}]{shen2023graph_34}
\bibinfo{author}{Tianhao, S.}, \bibinfo{author}{Kang, D.}, \bibinfo{author}{Jie, L.}, \bibinfo{author}{Ruyi, H.}, \bibinfo{author}{Weihua, L.}, \bibinfo{year}{2023}.
\newblock \bibinfo{title}{Graph structure and temporal data driven remaining useful life prediction method for machinery}.
\newblock \bibinfo{journal}{Journal of Mechanical Engineering} \bibinfo{volume}{59}, \bibinfo{pages}{183--194}.
\bibitem[{Veli{\v{c}}kovi{\'c} et~al.(2017)Veli{\v{c}}kovi{\'c}, Cucurull, Casanova, Romero, Lio and Bengio}]{velivckovic2017graph}
\bibinfo{author}{Veli{\v{c}}kovi{\'c}, P.}, \bibinfo{author}{Cucurull, G.}, \bibinfo{author}{Casanova, A.}, \bibinfo{author}{Romero, A.}, \bibinfo{author}{Lio, P.}, \bibinfo{author}{Bengio, Y.}, \bibinfo{year}{2017}.
\newblock \bibinfo{title}{Graph attention networks}.
\newblock \bibinfo{journal}{arXiv preprint arXiv:1710.10903} .
\bibitem[{Wang et~al.(2018)Wang, Lei, Li and Li}]{wang2018hybrid}
\bibinfo{author}{Wang, B.}, \bibinfo{author}{Lei, Y.}, \bibinfo{author}{Li, N.}, \bibinfo{author}{Li, N.}, \bibinfo{year}{2018}.
\newblock \bibinfo{title}{A hybrid prognostics approach for estimating remaining useful life of rolling element bearings}.
\newblock \bibinfo{journal}{IEEE Transactions on Reliability} \bibinfo{volume}{69}, \bibinfo{pages}{401--412}.
\bibitem[{Wang et~al.(2023a)Wang, Zhang, Lu and Wu}]{wang2023hierarchical_18}
\bibinfo{author}{Wang, G.}, \bibinfo{author}{Zhang, Y.}, \bibinfo{author}{Lu, M.}, \bibinfo{author}{Wu, Z.}, \bibinfo{year}{2023}a.
\newblock \bibinfo{title}{Hierarchical graph neural network with adaptive cross-graph fusion for remaining useful life prediction}.
\newblock \bibinfo{journal}{Measurement Science and Technology} \bibinfo{volume}{34}, \bibinfo{pages}{055112}.
\bibitem[{Wang et~al.(2023b)Wang, Zhang, Li, Deng and Jiang}]{wang2023comprehensive_19}
\bibinfo{author}{Wang, H.}, \bibinfo{author}{Zhang, Z.}, \bibinfo{author}{Li, X.}, \bibinfo{author}{Deng, X.}, \bibinfo{author}{Jiang, W.}, \bibinfo{year}{2023}b.
\newblock \bibinfo{title}{Comprehensive dynamic structure graph neural network for aero-engine remaining useful life prediction}.
\newblock \bibinfo{journal}{IEEE Transactions on Instrumentation and Measurement} .
\bibitem[{Wang et~al.(2022)Wang, Cao, Xu and Liu}]{wang2022gated_10}
\bibinfo{author}{Wang, L.}, \bibinfo{author}{Cao, H.}, \bibinfo{author}{Xu, H.}, \bibinfo{author}{Liu, H.}, \bibinfo{year}{2022}.
\newblock \bibinfo{title}{A gated graph convolutional network with multi-sensor signals for remaining useful life prediction}.
\newblock \bibinfo{journal}{Knowledge-Based Systems} \bibinfo{volume}{252}, \bibinfo{pages}{109340}.
\bibitem[{Wang et~al.(2024a)Wang, Cao, Ye, Xu and Yan}]{wang2024dvgtformer_54}
\bibinfo{author}{Wang, L.}, \bibinfo{author}{Cao, H.}, \bibinfo{author}{Ye, Z.}, \bibinfo{author}{Xu, H.}, \bibinfo{author}{Yan, J.}, \bibinfo{year}{2024}a.
\newblock \bibinfo{title}{Dvgtformer: A dual-view graph transformer to fuse multi-sensor signals for remaining useful life prediction}.
\newblock \bibinfo{journal}{Mechanical Systems and Signal Processing} \bibinfo{volume}{207}, \bibinfo{pages}{110935}.
\bibitem[{Wang et~al.(2021)Wang, Li, Zhang and Jia}]{wang2021spatio_3}
\bibinfo{author}{Wang, M.}, \bibinfo{author}{Li, Y.}, \bibinfo{author}{Zhang, Y.}, \bibinfo{author}{Jia, L.}, \bibinfo{year}{2021}.
\newblock \bibinfo{title}{Spatio-temporal graph convolutional neural network for remaining useful life estimation of aircraft engines}.
\newblock \bibinfo{journal}{Aerospace Systems} \bibinfo{volume}{4}, \bibinfo{pages}{29--36}.
\bibitem[{Wang et~al.(2024b)Wang, Jin, Wu, Li, Xie and Chen}]{wang2024k}
\bibinfo{author}{Wang, Y.}, \bibinfo{author}{Jin, R.}, \bibinfo{author}{Wu, M.}, \bibinfo{author}{Li, X.}, \bibinfo{author}{Xie, L.}, \bibinfo{author}{Chen, Z.}, \bibinfo{year}{2024}b.
\newblock \bibinfo{title}{K-link: Knowledge-link graph from llms for enhanced representation learning in multivariate time-series data}.
\newblock \bibinfo{journal}{arXiv preprint arXiv:2403.03645} .
\bibitem[{Wang et~al.(2023c)Wang, Wu, Jin, Li, Xie and Chen}]{wang2023local_15}
\bibinfo{author}{Wang, Y.}, \bibinfo{author}{Wu, M.}, \bibinfo{author}{Jin, R.}, \bibinfo{author}{Li, X.}, \bibinfo{author}{Xie, L.}, \bibinfo{author}{Chen, Z.}, \bibinfo{year}{2023}c.
\newblock \bibinfo{title}{Local--global correlation fusion-based graph neural network for remaining useful life prediction}.
\newblock \bibinfo{journal}{IEEE Transactions on Neural Networks and Learning Systems} .
\bibitem[{Wang et~al.(2023d)Wang, Wu, Li, Xie and Chen}]{wang2023multivariate_59}
\bibinfo{author}{Wang, Y.}, \bibinfo{author}{Wu, M.}, \bibinfo{author}{Li, X.}, \bibinfo{author}{Xie, L.}, \bibinfo{author}{Chen, Z.}, \bibinfo{year}{2023}d.
\newblock \bibinfo{title}{Multivariate time series representation learning via hierarchical correlation pooling boosted graph neural network}.
\newblock \bibinfo{journal}{IEEE Transactions on Artificial Intelligence} .
\bibitem[{Wang et~al.(2023e)Wang, Xu, Yang, Chen, Wu, Li and Xie}]{wang2023sensor_39}
\bibinfo{author}{Wang, Y.}, \bibinfo{author}{Xu, Y.}, \bibinfo{author}{Yang, J.}, \bibinfo{author}{Chen, Z.}, \bibinfo{author}{Wu, M.}, \bibinfo{author}{Li, X.}, \bibinfo{author}{Xie, L.}, \bibinfo{year}{2023}e.
\newblock \bibinfo{title}{Sensor alignment for multivariate time-series unsupervised domain adaptation}, in: \bibinfo{booktitle}{Proceedings of the AAAI Conference on Artificial Intelligence}, pp. \bibinfo{pages}{10253--10261}.
\bibitem[{Wang et~al.(2024c)Wang, Xu, Yang, Wu, Li, Xie and Chen}]{wang2024fully_60}
\bibinfo{author}{Wang, Y.}, \bibinfo{author}{Xu, Y.}, \bibinfo{author}{Yang, J.}, \bibinfo{author}{Wu, M.}, \bibinfo{author}{Li, X.}, \bibinfo{author}{Xie, L.}, \bibinfo{author}{Chen, Z.}, \bibinfo{year}{2024}c.
\newblock \bibinfo{title}{Fully-connected spatial-temporal graph for multivariate time-series data}, in: \bibinfo{booktitle}{Proceedings of the AAAI Conference on Artificial Intelligence}, pp. \bibinfo{pages}{15715--15724}.
\bibitem[{Wang et~al.(2024d)Wang, Xu, Yang, Wu, Li, Xie and Chen}]{wang2024graph}
\bibinfo{author}{Wang, Y.}, \bibinfo{author}{Xu, Y.}, \bibinfo{author}{Yang, J.}, \bibinfo{author}{Wu, M.}, \bibinfo{author}{Li, X.}, \bibinfo{author}{Xie, L.}, \bibinfo{author}{Chen, Z.}, \bibinfo{year}{2024}d.
\newblock \bibinfo{title}{Graph-aware contrasting for multivariate time-series classification}, in: \bibinfo{booktitle}{Proceedings of the AAAI Conference on Artificial Intelligence}, pp. \bibinfo{pages}{15725--15734}.
\bibitem[{Wang et~al.(2024e)Wang, Xu, Yang, Wu, Li, Xie and Chen}]{wang2024sea++}
\bibinfo{author}{Wang, Y.}, \bibinfo{author}{Xu, Y.}, \bibinfo{author}{Yang, J.}, \bibinfo{author}{Wu, M.}, \bibinfo{author}{Li, X.}, \bibinfo{author}{Xie, L.}, \bibinfo{author}{Chen, Z.}, \bibinfo{year}{2024}e.
\newblock \bibinfo{title}{Sea++: Multi-graph-based higher-order sensor alignment for multivariate time-series unsupervised domain adaptation}.
\newblock \bibinfo{journal}{IEEE transactions on pattern analysis and machine intelligence} .
\bibitem[{Wang et~al.(2020)Wang, Zhao and Addepalli}]{wang2020remaining}
\bibinfo{author}{Wang, Y.}, \bibinfo{author}{Zhao, Y.}, \bibinfo{author}{Addepalli, S.}, \bibinfo{year}{2020}.
\newblock \bibinfo{title}{Remaining useful life prediction using deep learning approaches: A review}.
\newblock \bibinfo{journal}{Procedia manufacturing} \bibinfo{volume}{49}, \bibinfo{pages}{81--88}.
\bibitem[{Wang et~al.(2024f)Wang, Xu, Li, Ren, Dong, Chen, Du, Wang, Shi and Zhang}]{wang2024remaining}
\bibinfo{author}{Wang, Z.}, \bibinfo{author}{Xu, Z.}, \bibinfo{author}{Li, Y.}, \bibinfo{author}{Ren, W.}, \bibinfo{author}{Dong, L.}, \bibinfo{author}{Chen, Z.}, \bibinfo{author}{Du, W.}, \bibinfo{author}{Wang, J.}, \bibinfo{author}{Shi, H.}, \bibinfo{author}{Zhang, X.}, \bibinfo{year}{2024}f.
\newblock \bibinfo{title}{A remaining useful life prediction framework with adaptive dynamic feedback}.
\newblock \bibinfo{journal}{Mechanical Systems and Signal Processing} \bibinfo{volume}{218}, \bibinfo{pages}{111595}.
\bibitem[{Wei and Wu(2023a)}]{wei2023prediction_25}
\bibinfo{author}{Wei, Y.}, \bibinfo{author}{Wu, D.}, \bibinfo{year}{2023}a.
\newblock \bibinfo{title}{Prediction of state of health and remaining useful life of lithium-ion battery using graph convolutional network with dual attention mechanisms}.
\newblock \bibinfo{journal}{Reliability Engineering \& System Safety} \bibinfo{volume}{230}, \bibinfo{pages}{108947}.
\bibitem[{Wei and Wu(2023b)}]{wei2023remaining_26}
\bibinfo{author}{Wei, Y.}, \bibinfo{author}{Wu, D.}, \bibinfo{year}{2023}b.
\newblock \bibinfo{title}{Remaining useful life prediction of bearings with attention-awared graph convolutional network}.
\newblock \bibinfo{journal}{Advanced Engineering Informatics} \bibinfo{volume}{58}, \bibinfo{pages}{102143}.
\bibitem[{Wei and Wu(2024)}]{wei2024state_53}
\bibinfo{author}{Wei, Y.}, \bibinfo{author}{Wu, D.}, \bibinfo{year}{2024}.
\newblock \bibinfo{title}{State of health and remaining useful life prediction of lithium-ion batteries with conditional graph convolutional network}.
\newblock \bibinfo{journal}{Expert Systems with Applications} \bibinfo{volume}{238}, \bibinfo{pages}{122041}.
\bibitem[{Wei et~al.(2023)Wei, Wu and Terpenny}]{wei2023bearing_17}
\bibinfo{author}{Wei, Y.}, \bibinfo{author}{Wu, D.}, \bibinfo{author}{Terpenny, J.}, \bibinfo{year}{2023}.
\newblock \bibinfo{title}{Bearing remaining useful life prediction using self-adaptive graph convolutional networks with self-attention mechanism}.
\newblock \bibinfo{journal}{Mechanical Systems and Signal Processing} \bibinfo{volume}{188}, \bibinfo{pages}{110010}.
\bibitem[{Wei et~al.(2024)Wei, Wu and Terpenny}]{wei2024remaining_44}
\bibinfo{author}{Wei, Y.}, \bibinfo{author}{Wu, D.}, \bibinfo{author}{Terpenny, J.}, \bibinfo{year}{2024}.
\newblock \bibinfo{title}{Remaining useful life prediction using graph convolutional attention networks with temporal convolution-aware nested residual connections}.
\newblock \bibinfo{journal}{Reliability Engineering \& System Safety} \bibinfo{volume}{242}, \bibinfo{pages}{109776}.
\bibitem[{Wen et~al.(2024)Wen, Fang, Wei, Liu, Chen and Wu}]{wen2024temporal_40}
\bibinfo{author}{Wen, Z.}, \bibinfo{author}{Fang, Y.}, \bibinfo{author}{Wei, P.}, \bibinfo{author}{Liu, F.}, \bibinfo{author}{Chen, Z.}, \bibinfo{author}{Wu, M.}, \bibinfo{year}{2024}.
\newblock \bibinfo{title}{Temporal and heterogeneous graph neural network for remaining useful life prediction}.
\newblock \bibinfo{journal}{arXiv preprint arXiv:2405.04336} .
\bibitem[{Wu et~al.(2024)Wu, He, Li, Miao, Li, Li and Shan}]{wu2024temporal_46}
\bibinfo{author}{Wu, J.}, \bibinfo{author}{He, D.}, \bibinfo{author}{Li, J.}, \bibinfo{author}{Miao, J.}, \bibinfo{author}{Li, X.}, \bibinfo{author}{Li, H.}, \bibinfo{author}{Shan, S.}, \bibinfo{year}{2024}.
\newblock \bibinfo{title}{Temporal multi-resolution hypergraph attention network for remaining useful life prediction of rolling bearings}.
\newblock \bibinfo{journal}{Reliability Engineering \& System Safety} \bibinfo{volume}{247}, \bibinfo{pages}{110143}.
\bibitem[{Wu et~al.(2020)Wu, Pan, Chen, Long, Zhang and Philip}]{wu2020comprehensive}
\bibinfo{author}{Wu, Z.}, \bibinfo{author}{Pan, S.}, \bibinfo{author}{Chen, F.}, \bibinfo{author}{Long, G.}, \bibinfo{author}{Zhang, C.}, \bibinfo{author}{Philip, S.Y.}, \bibinfo{year}{2020}.
\newblock \bibinfo{title}{A comprehensive survey on graph neural networks}.
\newblock \bibinfo{journal}{IEEE transactions on neural networks and learning systems} \bibinfo{volume}{32}, \bibinfo{pages}{4--24}.
\bibitem[{Xia et~al.(2023)Xia, Liang, Leng and Zheng}]{xia2023maintenance_38}
\bibinfo{author}{Xia, L.}, \bibinfo{author}{Liang, Y.}, \bibinfo{author}{Leng, J.}, \bibinfo{author}{Zheng, P.}, \bibinfo{year}{2023}.
\newblock \bibinfo{title}{Maintenance planning recommendation of complex industrial equipment based on knowledge graph and graph neural network}.
\newblock \bibinfo{journal}{Reliability Engineering \& System Safety} \bibinfo{volume}{232}, \bibinfo{pages}{109068}.
\bibitem[{Xiang et~al.(2021)Xiang, Qin, Luo and Pu}]{xiang2021spatiotemporally_7}
\bibinfo{author}{Xiang, S.}, \bibinfo{author}{Qin, Y.}, \bibinfo{author}{Luo, J.}, \bibinfo{author}{Pu, H.}, \bibinfo{year}{2021}.
\newblock \bibinfo{title}{Spatiotemporally multidifferential processing deep neural network and its application to equipment remaining useful life prediction}.
\newblock \bibinfo{journal}{IEEE Transactions on Industrial Informatics} \bibinfo{volume}{18}, \bibinfo{pages}{7230--7239}.
\bibitem[{Xiang et~al.(2023)Xiang, Qin, Luo, Wu and Gryllias}]{xiang2023concise}
\bibinfo{author}{Xiang, S.}, \bibinfo{author}{Qin, Y.}, \bibinfo{author}{Luo, J.}, \bibinfo{author}{Wu, F.}, \bibinfo{author}{Gryllias, K.}, \bibinfo{year}{2023}.
\newblock \bibinfo{title}{A concise self-adapting deep learning network for machine remaining useful life prediction}.
\newblock \bibinfo{journal}{Mechanical Systems and Signal Processing} \bibinfo{volume}{191}, \bibinfo{pages}{110187}.
\bibitem[{Xiao et~al.(2024)Xiao, Cui and Liu}]{xiao2024multi}
\bibinfo{author}{Xiao, Y.}, \bibinfo{author}{Cui, l.}, \bibinfo{author}{Liu, D.}, \bibinfo{year}{2024}.
\newblock \bibinfo{title}{Multi-graph attention fusion graph neural network for remaining useful life prediction of rolling bearings}.
\newblock \bibinfo{journal}{Measurement Science and Technology} .
\bibitem[{Xing et~al.(2023)Xing, Ding, Zhang and Li}]{xing2023stcgcn_29}
\bibinfo{author}{Xing, M.}, \bibinfo{author}{Ding, W.}, \bibinfo{author}{Zhang, T.}, \bibinfo{author}{Li, H.}, \bibinfo{year}{2023}.
\newblock \bibinfo{title}{Stcgcn: a spatio-temporal complete graph convolutional network for remaining useful life prediction of power transformer}.
\newblock \bibinfo{journal}{International Journal of Web Information Systems} \bibinfo{volume}{19}, \bibinfo{pages}{102--117}.
\bibitem[{Xu et~al.(2023a)Xu, Yang, Wang, Du and Gao}]{xu2023novel_35}
\bibinfo{author}{Xu, H.}, \bibinfo{author}{Yang, X.}, \bibinfo{author}{Wang, W.}, \bibinfo{author}{Du, J.}, \bibinfo{author}{Gao, J.}, \bibinfo{year}{2023}a.
\newblock \bibinfo{title}{A novel pre-trained model based on graph-labeling graph neural networks for tool wear prediction under variable working conditions}.
\newblock \bibinfo{journal}{Measurement Science and Technology} \bibinfo{volume}{34}, \bibinfo{pages}{125026}.
\bibitem[{Xu et~al.(2021)Xu, Chen, Wu, Wang, Wu and Li}]{xu2021kdnet}
\bibinfo{author}{Xu, Q.}, \bibinfo{author}{Chen, Z.}, \bibinfo{author}{Wu, K.}, \bibinfo{author}{Wang, C.}, \bibinfo{author}{Wu, M.}, \bibinfo{author}{Li, X.}, \bibinfo{year}{2021}.
\newblock \bibinfo{title}{Kdnet-rul: A knowledge distillation framework to compress deep neural networks for machine remaining useful life prediction}.
\newblock \bibinfo{journal}{IEEE Transactions on Industrial Electronics} \bibinfo{volume}{69}, \bibinfo{pages}{2022--2032}.
\bibitem[{Xu et~al.(2023b)Xu, Wu, Khoo, Chen and Li}]{xu2023hybrid}
\bibinfo{author}{Xu, Q.}, \bibinfo{author}{Wu, M.}, \bibinfo{author}{Khoo, E.}, \bibinfo{author}{Chen, Z.}, \bibinfo{author}{Li, X.}, \bibinfo{year}{2023}b.
\newblock \bibinfo{title}{A hybrid ensemble deep learning approach for early prediction of battery remaining useful life}.
\newblock \bibinfo{journal}{IEEE/CAA Journal of Automatica Sinica} \bibinfo{volume}{10}, \bibinfo{pages}{177--187}.
\bibitem[{Yang et~al.(2024)Yang, Liu, Zhou and Li}]{yang2024dynamic_50}
\bibinfo{author}{Yang, C.}, \bibinfo{author}{Liu, J.}, \bibinfo{author}{Zhou, K.}, \bibinfo{author}{Li, X.}, \bibinfo{year}{2024}.
\newblock \bibinfo{title}{Dynamic spatial--temporal graph-driven machine remaining useful life prediction method using graph data augmentation}.
\newblock \bibinfo{journal}{Journal of Intelligent Manufacturing} \bibinfo{volume}{35}, \bibinfo{pages}{355--366}.
\bibitem[{Yang et~al.(2023)Yang, Li, Zheng, Zhang and Wong}]{yang2023bearing_22}
\bibinfo{author}{Yang, X.}, \bibinfo{author}{Li, X.}, \bibinfo{author}{Zheng, Y.}, \bibinfo{author}{Zhang, Y.}, \bibinfo{author}{Wong, D.S.H.}, \bibinfo{year}{2023}.
\newblock \bibinfo{title}{Bearing remaining useful life prediction using spatial-temporal multiscale graph convolutional neural network}.
\newblock \bibinfo{journal}{Measurement Science and Technology} \bibinfo{volume}{34}, \bibinfo{pages}{085009}.
\bibitem[{Yang et~al.(2022)Yang, Zheng, Zhang, Wong and Yang}]{yang2022bearing_8}
\bibinfo{author}{Yang, X.}, \bibinfo{author}{Zheng, Y.}, \bibinfo{author}{Zhang, Y.}, \bibinfo{author}{Wong, D.S.H.}, \bibinfo{author}{Yang, W.}, \bibinfo{year}{2022}.
\newblock \bibinfo{title}{Bearing remaining useful life prediction based on regression shapalet and graph neural network}.
\newblock \bibinfo{journal}{IEEE Transactions on Instrumentation and Measurement} \bibinfo{volume}{71}, \bibinfo{pages}{1--12}.
\bibitem[{Yao et~al.(2023)Yao, Kang, Zhou, Rawa and Abusorrah}]{yao2023survey}
\bibinfo{author}{Yao, S.}, \bibinfo{author}{Kang, Q.}, \bibinfo{author}{Zhou, M.}, \bibinfo{author}{Rawa, M.J.}, \bibinfo{author}{Abusorrah, A.}, \bibinfo{year}{2023}.
\newblock \bibinfo{title}{A survey of transfer learning for machinery diagnostics and prognostics}.
\newblock \bibinfo{journal}{Artificial Intelligence Review} \bibinfo{volume}{56}, \bibinfo{pages}{2871--2922}.
\bibitem[{Ying et~al.(2018)Ying, You, Morris, Ren, Hamilton and Leskovec}]{ying2018hierarchical}
\bibinfo{author}{Ying, Z.}, \bibinfo{author}{You, J.}, \bibinfo{author}{Morris, C.}, \bibinfo{author}{Ren, X.}, \bibinfo{author}{Hamilton, W.}, \bibinfo{author}{Leskovec, J.}, \bibinfo{year}{2018}.
\newblock \bibinfo{title}{Hierarchical graph representation learning with differentiable pooling}.
\newblock \bibinfo{journal}{Advances in neural information processing systems} \bibinfo{volume}{31}.
\bibitem[{Yuan et~al.(2022)Yuan, Yu, Gui and Ji}]{yuan2022explainability}
\bibinfo{author}{Yuan, H.}, \bibinfo{author}{Yu, H.}, \bibinfo{author}{Gui, S.}, \bibinfo{author}{Ji, S.}, \bibinfo{year}{2022}.
\newblock \bibinfo{title}{Explainability in graph neural networks: A taxonomic survey}.
\newblock \bibinfo{journal}{IEEE transactions on pattern analysis and machine intelligence} \bibinfo{volume}{45}, \bibinfo{pages}{5782--5799}.
\bibitem[{Zeng et~al.(2022)Zeng, Yang, Liu, Zhou, Li, Wei and Liu}]{zeng2022remaining_11}
\bibinfo{author}{Zeng, X.}, \bibinfo{author}{Yang, C.}, \bibinfo{author}{Liu, J.}, \bibinfo{author}{Zhou, K.}, \bibinfo{author}{Li, D.}, \bibinfo{author}{Wei, S.}, \bibinfo{author}{Liu, Y.}, \bibinfo{year}{2022}.
\newblock \bibinfo{title}{Remaining useful life prediction for rotating machinery based on dynamic graph and spatial--temporal network}.
\newblock \bibinfo{journal}{Measurement Science and Technology} \bibinfo{volume}{34}, \bibinfo{pages}{035102}.
\bibitem[{Zhang et~al.(2016)Zhang, Lim, Qin and Tan}]{zhang2016multiobjective}
\bibinfo{author}{Zhang, C.}, \bibinfo{author}{Lim, P.}, \bibinfo{author}{Qin, A.K.}, \bibinfo{author}{Tan, K.C.}, \bibinfo{year}{2016}.
\newblock \bibinfo{title}{Multiobjective deep belief networks ensemble for remaining useful life estimation in prognostics}.
\newblock \bibinfo{journal}{IEEE transactions on neural networks and learning systems} \bibinfo{volume}{28}, \bibinfo{pages}{2306--2318}.
\bibitem[{Zhang et~al.(2022)Zhang, Tian, Li, Leon, Franquelo, Luo and Yin}]{zhang2022parallel_12}
\bibinfo{author}{Zhang, J.}, \bibinfo{author}{Tian, J.}, \bibinfo{author}{Li, M.}, \bibinfo{author}{Leon, J.I.}, \bibinfo{author}{Franquelo, L.G.}, \bibinfo{author}{Luo, H.}, \bibinfo{author}{Yin, S.}, \bibinfo{year}{2022}.
\newblock \bibinfo{title}{A parallel hybrid neural network with integration of spatial and temporal features for remaining useful life prediction in prognostics}.
\newblock \bibinfo{journal}{IEEE Transactions on Instrumentation and Measurement} \bibinfo{volume}{72}, \bibinfo{pages}{1--12}.
\bibitem[{Zhang et~al.(2024a)Zhang, Tian, Yan, Wu, Luo and Yin}]{zhang2024multi_51}
\bibinfo{author}{Zhang, J.}, \bibinfo{author}{Tian, J.}, \bibinfo{author}{Yan, P.}, \bibinfo{author}{Wu, S.}, \bibinfo{author}{Luo, H.}, \bibinfo{author}{Yin, S.}, \bibinfo{year}{2024}a.
\newblock \bibinfo{title}{Multi-hop graph pooling adversarial network for cross-domain remaining useful life prediction: A distributed federated learning perspective}.
\newblock \bibinfo{journal}{Reliability Engineering \& System Safety} \bibinfo{volume}{244}, \bibinfo{pages}{109950}.
\bibitem[{Zhang et~al.(2024b)Zhang, Yuan, Jiang and Zhao}]{zhang2024novel}
\bibinfo{author}{Zhang, K.}, \bibinfo{author}{Yuan, J.}, \bibinfo{author}{Jiang, H.}, \bibinfo{author}{Zhao, Q.}, \bibinfo{year}{2024}b.
\newblock \bibinfo{title}{A novel weighted sparsity index based on multichannel fused graph spectra for machine health monitoring}.
\newblock \bibinfo{journal}{Mechanical Systems and Signal Processing} \bibinfo{volume}{215}, \bibinfo{pages}{111417}.
\bibitem[{Zhang et~al.(2023a)Zhang, Leng, Zhao, Li, Yu and Chen}]{zhang2023spatial_28}
\bibinfo{author}{Zhang, X.}, \bibinfo{author}{Leng, Z.}, \bibinfo{author}{Zhao, Z.}, \bibinfo{author}{Li, M.}, \bibinfo{author}{Yu, D.}, \bibinfo{author}{Chen, X.}, \bibinfo{year}{2023}a.
\newblock \bibinfo{title}{Spatial-temporal dual-channel adaptive graph convolutional network for remaining useful life prediction with multi-sensor information fusion}.
\newblock \bibinfo{journal}{Advanced Engineering Informatics} \bibinfo{volume}{57}, \bibinfo{pages}{102120}.
\bibitem[{Zhang et~al.(2021)Zhang, Li, Wang, Yang and Wei}]{zhang2021adaptive_4}
\bibinfo{author}{Zhang, Y.}, \bibinfo{author}{Li, Y.}, \bibinfo{author}{Wang, Y.}, \bibinfo{author}{Yang, Y.}, \bibinfo{author}{Wei, X.}, \bibinfo{year}{2021}.
\newblock \bibinfo{title}{Adaptive spatio-temporal graph information fusion for remaining useful life prediction}.
\newblock \bibinfo{journal}{IEEE Sensors Journal} \bibinfo{volume}{22}, \bibinfo{pages}{3334--3347}.
\bibitem[{Zhang et~al.(2020)Zhang, Li, Wei and Jia}]{zhang2020adaptive_1}
\bibinfo{author}{Zhang, Y.}, \bibinfo{author}{Li, Y.}, \bibinfo{author}{Wei, X.}, \bibinfo{author}{Jia, L.}, \bibinfo{year}{2020}.
\newblock \bibinfo{title}{Adaptive spatio-temporal graph convolutional neural network for remaining useful life estimation}, in: \bibinfo{booktitle}{2020 International joint conference on neural networks (IJCNN)}, \bibinfo{organization}{IEEE}. pp. \bibinfo{pages}{1--7}.
\bibitem[{Zhang et~al.(2023b)Zhang, Zhou, Huang, Jin and Xiao}]{zhang2023temporal_30}
\bibinfo{author}{Zhang, Y.}, \bibinfo{author}{Zhou, W.}, \bibinfo{author}{Huang, J.}, \bibinfo{author}{Jin, X.}, \bibinfo{author}{Xiao, G.}, \bibinfo{year}{2023}b.
\newblock \bibinfo{title}{Temporal knowledge graph informer network for remaining useful life prediction}.
\newblock \bibinfo{journal}{IEEE Transactions on Instrumentation and Measurement} .
\bibitem[{Zheng et~al.(2024)Zheng, Jia and Yang}]{zheng2024improved_58}
\bibinfo{author}{Zheng, L.}, \bibinfo{author}{Jia, W.}, \bibinfo{author}{Yang, R.}, \bibinfo{year}{2024}.
\newblock \bibinfo{title}{Improved adaptive war strategy optimization algorithm assisted-adaptive multi-head graph attention mechanism network for remaining useful life of complex equipment}.
\newblock \bibinfo{journal}{AIP Advances} \bibinfo{volume}{14}.
\bibitem[{Zhou et~al.(2021)Zhou, Zhang, Peng, Zhang, Li, Xiong and Zhang}]{zhou2021informer}
\bibinfo{author}{Zhou, H.}, \bibinfo{author}{Zhang, S.}, \bibinfo{author}{Peng, J.}, \bibinfo{author}{Zhang, S.}, \bibinfo{author}{Li, J.}, \bibinfo{author}{Xiong, H.}, \bibinfo{author}{Zhang, W.}, \bibinfo{year}{2021}.
\newblock \bibinfo{title}{Informer: Beyond efficient transformer for long sequence time-series forecasting}, in: \bibinfo{booktitle}{Proceedings of the AAAI conference on artificial intelligence}, pp. \bibinfo{pages}{11106--11115}.
\bibitem[{Zhou and Wang(2024)}]{zhou2024mst_47}
\bibinfo{author}{Zhou, L.}, \bibinfo{author}{Wang, H.}, \bibinfo{year}{2024}.
\newblock \bibinfo{title}{Mst-gat: A multi-perspective spatial-temporal graph attention network for multi-sensor equipment remaining useful life prediction}.
\newblock \bibinfo{journal}{Information Fusion} \bibinfo{volume}{110}, \bibinfo{pages}{102462}.
\bibitem[{Zhu et~al.(2023)Zhu, Xiong, Yang and Yu}]{zhu2023rgcnu_21}
\bibinfo{author}{Zhu, Q.}, \bibinfo{author}{Xiong, Q.}, \bibinfo{author}{Yang, Z.}, \bibinfo{author}{Yu, Y.}, \bibinfo{year}{2023}.
\newblock \bibinfo{title}{Rgcnu: Recurrent graph convolutional network with uncertainty estimation for remaining useful life prediction}.
\newblock \bibinfo{journal}{IEEE/CAA Journal of Automatica Sinica} \bibinfo{volume}{10}, \bibinfo{pages}{1640--1642}.
\bibitem[{Zhuang et~al.(2024)Zhuang, Chen, Zhao, Jia and Feng}]{zhuang2024graph_56}
\bibinfo{author}{Zhuang, J.}, \bibinfo{author}{Chen, Y.}, \bibinfo{author}{Zhao, X.}, \bibinfo{author}{Jia, M.}, \bibinfo{author}{Feng, K.}, \bibinfo{year}{2024}.
\newblock \bibinfo{title}{A graph-embedded subdomain adaptation approach for remaining useful life prediction of industrial iot systems}.
\newblock \bibinfo{journal}{IEEE Internet of Things Journal} .

\end{thebibliography}



\end{document}